\documentclass[11pt,a4paper,oneside,dvipsnames]{article}

\usepackage{graphicx}
\usepackage{rotating}
\usepackage{wrapfig}
\usepackage{placeins}

\usepackage{sidecap}

\usepackage[top=1in]{geometry}

\usepackage{latexsym}
\usepackage{graphicx}
\usepackage{multicol,multirow}
\usepackage{amsmath,amssymb,amsfonts}
\usepackage{mathrsfs}
\usepackage{amsthm}
\usepackage{rotating}
\usepackage{appendix}
\usepackage{ifpdf}
\usepackage[T1]{fontenc}
\usepackage{times}
\usepackage{sourcesanspro}
\usepackage{newtxmath}
\usepackage{textcomp}%
\usepackage{xcolor}%
\usepackage{csvsimple}
\usepackage{adjustbox}
\usepackage{xr}
\usepackage{soul}

\usepackage{xr-hyper}
\usepackage{hyperref}
\usepackage{siunitx}
\usepackage{authblk}

\usepackage{subcaption}
\usepackage[utf8]{inputenc}

\usepackage{array}

\DeclareUnicodeCharacter{2080}{\textsubscript{0}}
\DeclareUnicodeCharacter{2081}{\textsubscript{1}}
\DeclareUnicodeCharacter{2082}{\textsubscript{2}}
\DeclareUnicodeCharacter{2083}{\textsubscript{3}}
\DeclareUnicodeCharacter{2084}{\textsubscript{4}}
\DeclareUnicodeCharacter{2085}{\textsubscript{5}}
\DeclareUnicodeCharacter{2093}{\textsubscript{x}}

\usepackage{geometry}

\newgeometry{vmargin={15mm}, hmargin={12mm,17mm}}   

\hypersetup{
    colorlinks = true,
    linkcolor = blue,
    filecolor = magenta,      
    urlcolor = cyan,
}

\usepackage{booktabs}
\usepackage{pgfplotstable} 

\definecolor{reddish}{HTML}{FBB4AE}
\definecolor{blueish}{HTML}{B3CDE3}
\definecolor{magentish}{HTML}{FF00AA}
\definecolor{greenish}{HTML}{a1d99b}

\makeatletter
\newcommand*{\addFileDependency}[1]{
  \typeout{(#1)}
  \@addtofilelist{#1}
  \IfFileExists{#1}{}{\typeout{No file #1.}}
}
\makeatother

\newcommand*{\myexternaldocument}[2]{%
    \externaldocument[#2]{#1}%
    \addFileDependency{#2.tex}%
    \addFileDependency{#2.aux}%
}

\myexternaldocument{supplementary_data}{S-}

\title{A Data-Driven Supervised Machine Learning Approach to Estimating Global Ambient Air Pollution Concentrations With Associated Prediction Intervals}
\author{
  Liam J. Berrisford$^{1,2,3,*}$\\
  \texttt{l.berrisford@exeter.ac.uk}
 \and
 Hugo Barbosa$^{1}$\\
 \texttt{h.barbosa@exeter.ac.uk}
 \and
  Ronaldo Menezes$^{1,4}$\\
  \texttt{r.menezes@exeter.ac.uk}
}

\date{%
    $^1$ BioComplex Laboratory, Department of Computer Science, University of Exeter, England\\%
    $^2$ Department of Mathematics, University of Exeter, England\\
    $^3$ UKRI Centre for Doctoral Training in Environmental Intelligence, University of Exeter, England\\%
    $^4$ Department of Computer Science, Federal University of Ceará, Fortaleza, Brazil\\%
    $^{*}$ Corresponding Author\\[2ex]%
    \today
}

\begin{document}

\maketitle

\begin{abstract}
Global ambient air pollution, a transboundary challenge, is typically addressed through interventions relying on data from spatially sparse and heterogeneously placed monitoring stations. These stations often encounter temporal data gaps due to issues such as power outages. In response, we have developed a scalable, data-driven, supervised machine learning framework. This model is designed to impute missing temporal and spatial measurements, thereby generating a comprehensive dataset for pollutants including NO$_2$, O$_3$, PM$_{10}$, PM$_{2.5}$, and SO$_2$. The dataset, with a fine granularity of 0.25$^{\circ}$ at hourly intervals and accompanied by prediction intervals for each estimate, caters to a wide range of stakeholders relying on outdoor air pollution data for downstream assessments. This enables more detailed studies. Additionally, the model's performance across various geographical locations is examined, providing insights and recommendations for strategic placement of future monitoring stations to further enhance the model's accuracy.
\end{abstract}

\noindent\textbf{Keywords:} Machine Learning, Global Air Pollution, Temporal and Spatial Modelling, Prediction Intervals

\clearpage
\section{Introduction}
\label{sec:introduction}

Air pollution represents a significant global challenge. Astonishingly, 99\% of the global population is exposed to air pollution that exceeds the air quality limits established by the World Health Organization (WHO) \cite{WHO:2021:GlobalAirPollutionStatistics}. In response, numerous countries have implemented air pollution monitoring systems\footnote{For a global view of air pollution monitoring, visit the OpenAQ Global Air Pollution Monitoring Map: \url{https://explore.openaq.org/}}. While direct monitoring through stations is an essential initial step in comprehending air pollution, it is impractical to deploy a monitoring station at every location. This impracticality stems from both logistical challenges, such as the necessity for infrastructure like power lines and data connections, and financial considerations, with a standard monitoring station in the UK costing up to £198,000 \cite{AEATechnology:2006:PurchasingAURNCost}. Consequently, models are crucial for bridging spatial and temporal gaps in air pollution concentration measurements. Air pollution models, like those used by the WHO, typically provide data with an annual temporal resolution. However, there is an evident gap in global models that focus on hourly air pollution concentrations. Considering that certain air pollution guidelines, including WHO's, specifically demand hourly resolution \cite{hoffmann:2021:WHO2021AQGs}, the need for comprehensive, hourly resolved air pollution concentration estimates is paramount for effective decision-making. This paper introduces a data-driven, supervised machine-learning model designed to predict air pollution concentrations at an hourly resolution on a global scale. There are four key outputs from this work:

\begin{enumerate}
    \item Development of a comprehensive air pollution concentration map, covering the entire globe for the year 2022, at an hourly resolution. This map includes concentrations of NO$_2$, O$_3$, PM$_{10}$, PM$_{2.5}$, and SO$_2$, the quintet of pollutants that constitute the Daily Air Quality Index (DAQI) in the UK \cite{UKAIR:2023:DAQI}.
    \item Analysis to gauge the feasibility of extrapolating air pollution concentrations from one region to another. This involves addressing the critical question: \textbf{To what extent can the air pollution data of one country accurately predict the air pollution levels in another?}
    \item Provision of strategic recommendations for the placement of future air pollution monitoring stations, informed by the uncertainty metrics derived from our model.
    \item A comprehensive evaluation of global air quality, identifying regions with the most severe air pollution and pinpointing which of the five DAQI pollutants predominantly contributes to poor air quality in these areas.
\end{enumerate}

The complete spatial map of air pollution concentrations can be used downstream in other scientific studies, such as the health implication of particulate matter (PM) \cite{boldo:2006:PMHealthImpact}, or the impact of ozone on vegetation \cite{ashmore:2005:OzoneVegetationImpact} at a resolution not previously possible. The unprecedented resolution of this map enables more detailed and accurate analyses than were previously feasible. Furthermore, our approach to estimating air pollution concentrations from one country based on data from another represents a step towards open data initiatives. This fosters international data sharing, contributing to a more equitable global understanding of air pollution and research \cite{hajat:2015:GlobalAirPollutionResearch}. Our findings illustrate that even data-rich countries like the United Kingdom can benefit from exchanging data with countries that have less comprehensive data. The manuscript explores the concept that air pollution data from France can offer insights into air pollution in the UK, suggesting that international cooperation can be a cost-effective strategy for achieving a complete and detailed spatial air pollution map.

\section{Related Work}

There are numerous advantages to estimating air pollution concentrations on a global rather than regional scale. Air pollution is inherently transboundary, disregarding national boundaries. Pollutants released in one country can traverse long distances, impacting areas far from their original emission points, such as Saharan dust adversely affecting air quality in the UK \cite{vieno:2016:UKAirQualitySaharanDust}. Additionally, climate change is anticipated to broadly affect global processes that influence air pollution. For example, ozone formation is expected to increase with rising temperatures \cite{bell:2007:OzoneClimateChange}, wildfires are predicted to become more frequent \cite{goss:2020:WildfireClimateChange}, and their contribution to air pollution is likely to grow \cite{knorr:2017:WildfireAirPollution}. Beyond the direct sources of air pollution, climate change is forecasted to alter several influencing factors, such as an increase in stagnation events—periods of minimal wind leading to localized pollution accumulation \cite{horton:2014:StangnationEventsClimateChange}. Expected changes in precipitation patterns \cite{trenberth:2011:PrecipitationChangesClimateChange} will further complicate matters, affecting the wet deposition process crucial for removing atmospheric pollutants. A comprehensive, spatially complete view of air pollution concentrations globally, at a high temporal resolution like hourly, is essential for understanding the effects of such events. This understanding enables the development of mitigation strategies aimed at improving human and environmental health amidst the global air pollution crisis.

\subsection{Measuring Air Pollution}
\label{sec:measuringAirPollution}

The primary method of determining a location's air pollution concentration is to conduct measurements with specialised equipment. In-situ measuring equipment can be split into two categories: high-quality stationary monitoring stations, such as those used in the UK AURN network\footnote{\href{https://uk-air.defra.gov.uk/interactive-map}{UK Automatic Urban and Rural Network (AURN) Interactive Map.}}, and mobile, low-cost air quality sensors \cite{kang:2022:LowCostAirPollutionSensor}, such as Purple Air sensors\footnote{\href{https://map.purpleair.com/}{Purple Air Interactive Map.}}. While monitoring stations used in the AURN are desirable, at an individual cost of £198,000 \cite{AEATechnology:2006:PurchasingAURNCost}, they are impossible to use for complete spatial coverage over a geographic region such as the UK, with the UK only currently having 171 monitoring stations online. This fundamental shortcoming of stationary monitoring stations can be overcome using low-cost sensors, providing measurements over a larger geographic region. However, issues are present with the current state of the technology used in low-cost sensors. Changes in atmospheric composition and meteorological conditions can influence the measurements being made, alongside other air pollutants impacting the measurement of a target pollutant \cite{DEFRA:2023:LowCostSensors}. One of the main costs associated with stationary monitoring stations is the quality control conducted on the sensor, ensuring measurements are made under the same conditions concerning elements, such as the height at which the measurement is taken \cite{concas:2021:LowCostSensorCalibration}. Therefore, while low-cost sensors provide a benefit, they should be used cautiously, with their primary purpose being raising awareness rather than applications requiring higher accuracy, such as epidemiological studies or compliance with air quality legislation \cite{Castell:2017:LowCostSensorEUComparison}.

Ex-situ measurements of air pollution concentrations are also possible with remote sensing. Sentinel 5P \cite{veefkind:2012:Sentinel5PDescription} is an ESA satellite providing air pollution concentration measurements over large geographic regions. Sentinel 5P has a near-polar, sun-synchronous orbit \cite{ESA:2023:Sentinel5POrbit}, meaning that the platform will always pass over a region at a similar time each day. While this makes it great for comparing locations between each other, it makes it challenging to have a neat, complete air pollution concentration picture across a whole day, with questions such as the difference between rush hour and midnight air pollution concentrations impossible to answer. Furthermore, environmental conditions can make measurements at given locations and times impossible \cite{ESA:2017:ESAQualityAssurance}. These issues result in the datasets produced by Sentinel 5P being incomplete, with considerable missing data. The recently operational TEMPO \cite{zoogman:2017:NASATEMPORemoteSensing} remote sensing platform has similar properties and limitations to Sentinel 5P, with the critical distinction between the two being TEMPOs mission of hourly observations over just North America.  

While various methods exist for measuring air pollution concentrations, each approach has its limitations in certain aspects. To compensate for these shortcomings, model outputs are employed to augment observational data. This ensures the availability of air pollution concentration maps that are both temporally and spatially complete, facilitating their use in downstream applications.

\subsection{Modelling Air Pollution}

There are two primary frameworks for modelling air pollution: Lagrangian and Eulerian. Lagrangian models focus on tracking individual air parcels (or particles), aiming to trace their movement through the atmosphere over time and space \cite{eliassen:1984:LagrangianAirModels}. In contrast, Eulerian models do not track air parcels individually; instead, they divide the atmosphere into a grid of regions to monitor air pollution concentrations at fixed points over time \cite{Byun:1984:EulerianDispersionModels}. Eulerian models are particularly useful for examining the spatial distribution of air pollution across large areas, such as providing a comprehensive view of air pollution throughout Europe \cite{zlatev:1992:EulerianModelUse}. Conversely, Lagrangian models are more appropriate for investigating specific pollution sources, like the dispersion of ash from a volcanic eruption \cite{Vitturi:2010:lagrangianVolcano}. The research presented in this manuscript utilises the Eulerian modelling approach, with the goal of employing machine learning techniques to estimate air pollution concentrations on a global scale. 

The options for employing an Eulerian air pollution model framework capable of providing global measurements are notably more restricted than those for regional or local models, yet they do exist. GEOS-Chem \cite{Henze:2007:GeosChem}, a mechanistic model, represents a global 3D atmospheric chemistry model that incorporates input data including meteorological factors. However, GEOS-Chem demands significant expertise in the domain to effectively navigate its complexities, alongside substantial infrastructure support. For instance, a standard simulation with GEOS-Chem at \ang{4}x\ang{5} resolution necessitates 15GB of RAM\footnote{\href{https://geos-chem.readthedocs.io/en/stable/gcc-guide/01-startup/memory.html}{GEOS-Chem Hardware Requirements.}}. Should the computational demands of models like GEOS-Chem prove too onerous for certain applications, statistical models, such as Land Use Regression (LUR), may offer a more feasible alternative. LUR provides a method to develop stochastic air pollution models using input data variables like meteorological conditions, terrain, land use, and road network data \cite{hoek:2008:review}.

A rapidly emerging area is the use of deterministic models to address the current gap within the existing suite of models that can provide high-resolution air pollution concentration data, both temporally and spatially, to empower stakeholders to make informed decisions concerning air pollution. A range of these models are based on a data-driven supervised machine learning model where a target vector, generally air pollution concentrations, is estimated from a feature vector, such as meteorological variables. The model aims to learn the relationship between the target and feature vectors in situations where both are available, enabling subsequent predictions of target vectors in situations where only the feature vector is available. In the scientific literature, several studies use machine learning techniques to forecast air pollution concentrations \cite{freeman:2018:ForecastingAirPollutionConcentration, tao:2019:ForecastingAirPollutionConcentration2, harishkumar:2020:ForecastingAirPollutionConcentration3}. However, this has the limitation of needing existing air pollution data where the forecast is being conducted. Historical air pollution concentrations limit the model's use to locations where an air pollution monitoring station exists.

Existing studies have tackled the problem of estimating air pollution concentrations in locations without monitoring stations. However, the studies focus either on small geographical areas, such as the Bay of Algeciras (Spain) with hourly temporal resolution \cite{van:2019:MissingLocationAirPollutionEstimationSmallArea} or a large geographical area with low temporal resolution, such as monthly \cite{chen:2021:MissingLocationAirPollutionMonthly}. Some work exists in the middle of existing studies, such as daily temporal resolution over a larger spatial area \cite{he:2023:AirPollutionLeaveOneOutValidationDaily, li:2020:AirPollutionLeaveOneOutValidationDaily2}, or larger area still with annual temporal resolution \cite{berrisford:2022:MLAnnualEnglandWales}. 

In our previous research, we introduced a model that integrates various aspects to predict air pollution concentrations with an hourly temporal resolution across England's extensive geographical area \cite{berrisford:2024:MLHourlyEngland}. This task presented a significant challenge due to the variability of air pollution concentrations across the locations covered. We successfully developed an air pollution model for England, providing data at a 1km$^2$ hourly resolution, which is invaluable for numerous downstream applications. This model employed a modified Eulerian framework, utilising a machine-learning model as a synthetic monitoring station. The model's process involves simulating the air pollution concentration reading of a monitoring station under the environmental conditions specified by the input data. By training on data that illustrate the relationship between environmental conditions and air pollution concentrations, the model can utilise known environmental conditions across the study area to predict air pollution concentrations in locations where data are less readily available, thus offering a comprehensive overview of air pollution levels. A key advantage of this approach over other deterministic methods is the significant improvement in computational speed, especially critical for predictions at a global spatial resolution. This efficiency gain is achieved by eliminating spatial dependencies between grid cells—treating each synthetic monitoring station as independent, which facilitates computational speed through the parallelisation of predictions. This also enables easier data exploration by allowing predictions for individual locations. In this work, we aim to apply this approach globally, exploring the challenges associated with employing a transboundary data-driven supervised machine learning air pollution model and underscoring the advantages of an ambient air pollution model that scales linearly with computational complexity.

\section{Data}
\subsection{Target Vector: Air Pollution Concentrations}
\label{sec:targetVector}

\begin{figure}
    \includegraphics[width=.9\linewidth]{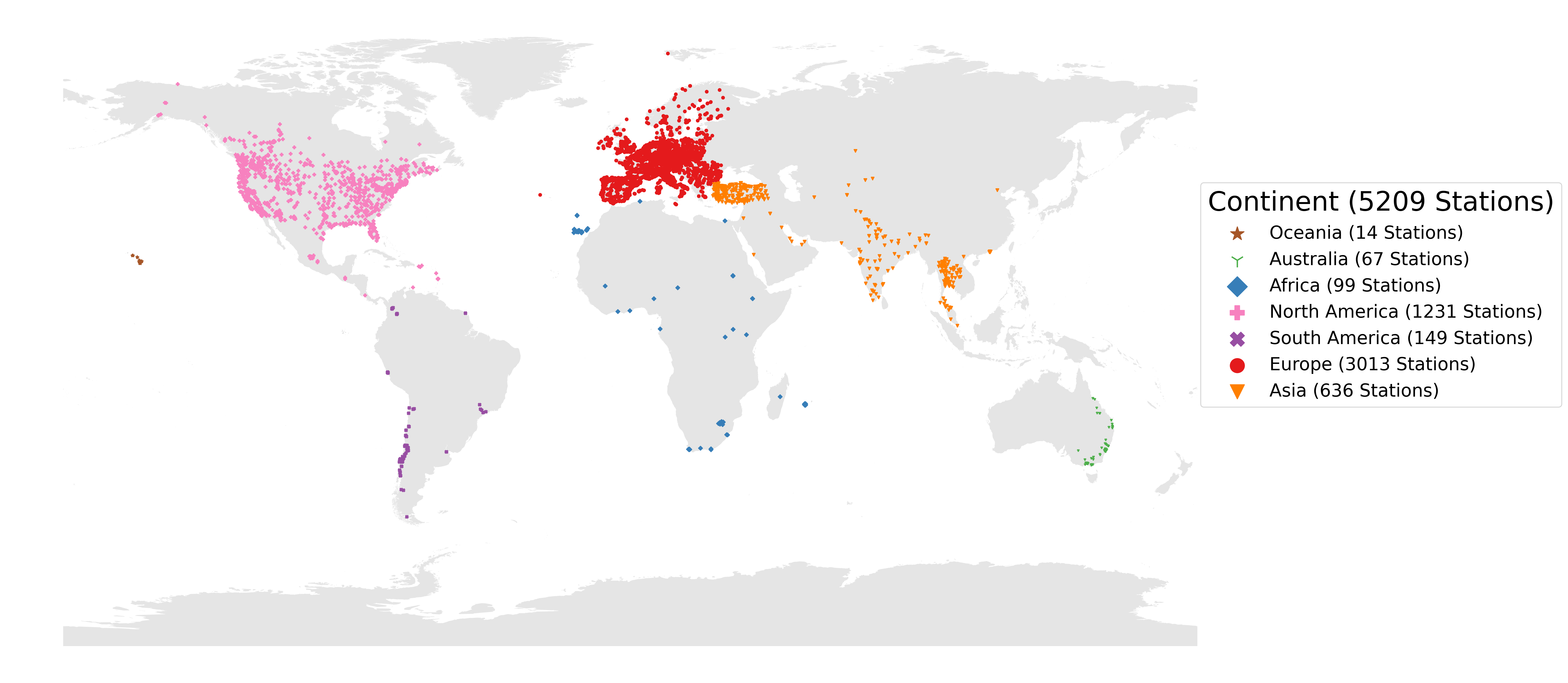}
\caption{{\bfseries Spatial Distribution of Monitoring Station Locations Within the OpenAQ Dataset for All Air Pollutants in 2022.} The map reveals a high density of stations within Europe and the US relative to the geographical areas these regions cover. This underscores the disparity between countries regarding the extent of their monitoring station networks and the availability of data to address air pollution challenges.}
\label{fig:openAQMonitoringStationLocations}
\end{figure}

The air pollution data utilised in this study were sourced from OpenAQ \cite{OpenAQ:2023:AboutUs}. OpenAQ serves as a platform aggregating data from diverse sources into a unified standard format. Given the wide variety of monitoring station networks globally, each operated by different governments and each with its unique data format, amalgamating these data into a consistent format necessary for this research would entail significant effort and data pre-processing from their original datasets. The OpenAQ air pollution dataset is global, offering data from monitoring stations located in various regions, as illustrated in Figure \ref{fig:openAQMonitoringStationLocations}.

One of the issues with the data acquired was that some of the monitoring stations, rather than measuring in \si{\micro\gram/\meter^3}, would measure in parts per billion (ppb).  \si{\micro\gram/\meter^3} measurements can be calculated from ppb measurements when the molecular weight, temperature and pressure are known \cite{vallero:2014fundamentalsConvertPPBtomicrogram}. However, this approach introduces another dimension to the study and potentially systematic errors, arising from the fundamental differences in how concentrations are measured across various monitoring networks. As such, we decided that for this study, OpenAQ monitoring stations that measured an air pollutant in ppb would be discarded. The case of ppb measurements only occurred in monitoring stations for NO$_2$, O$_3$ and SO$_2$, with the impact of the spatial distribution of monitoring stations visible in Figure \ref{S-fig:allOpenAQMonitoringStationLocations}.

We excluded further monitoring stations from the dataset based on several criteria indicative of poor data quality. These criteria included stations with two or fewer data points, occurrences of repeated timestamps with differing air pollution concentration measurements, and stations recording all measurements as zero or consistently reporting a single value across all timestamps. To ensure the high quality of data for our model, we removed stations lacking an hourly reading for each hour of the day across their dataset (e.g., at least one measurement for hours 0000-2300) and at least one measurement for each day of the week. Although the exclusion of data based on these criteria may appear extensive, they constituted a minor portion of the total dataset. After cleaning, the dataset still comprised considerable numbers: 2,894 stations for NO$_2$, 2,076 for O$_3$, 2,924 for PM$_{10}$, 2,939 for PM$_{2.5}$, and 1,446 for SO$_2$. As a proof of concept in this study, we focused on 2022.

\subsection{Feature Vector}
\label{sec:featureVectors}

While data concerning a range of phenomena that act as sinks or sources of air pollution could be included in this study, we decided to limit the input data to data sources with global coverage with a consistent schema. The goal was to simplify the model to be created and remove the need to worry about integrating potentially 100s of datasets from national datasets to achieve global coverage, which we saw as outside this work's scope in achieving the goals outlined in Section \ref{sec:introduction}. As such, we focused on four prominent datasets concerning broad phenomena related to air pollution, each containing individual datasets describing particular phenomena. The study includes temporal, meteorological, remote sensing and emissions datasets, with a total of 26 feature vector elements. 

\paragraph{Temporal, 5 features. }
Air pollution concentrations exhibit various cyclical patterns, including diurnal cycles influenced by rush hour traffic for NO$_2$ \cite{goldberg:2021:NO2DailyCycles} and sunlight hours for O$_3$ \cite{garland:1979:OzoneDailyCycle}. Additionally, weekly trends emerge due to the working week's schedule \cite{beirle:2003:NOxWeekly, gietl:2009:PMWeeklyCycle}. Seasonal cycles, driven by residential heating, elevate air pollutants during winter months \cite{feng:2014:WinterPMResidentialHeating, meng:2018:SO2Increases}. Moreover, winter conditions stabilise atmospheric movements, diminishing air pollution dispersion \cite{Qianhui:2022:StableWinterAirPollutionIncrease}, with specific effects on O$_3$ levels due to colder temperatures and reduced sunlight \cite{cichowicz:2017:OzoneWinterReduction}. Consequently, the hour, day of the week, week number, and month were incorporated as elements of the feature vector, based on UTC to maintain consistency across different time zones globally. However, utilising UTC presents challenges, such as misalignment with local times for specific activities, like rush hour. To address this, another feature vector element, the UTC offset, was introduced to account for the difference between the local time and UTC time.

\paragraph{Meteorological, 11 features.} A range of meteorological phenomena plays a crucial role in air pollution concentrations. The direction and speed of air pollution dispersion are heavily influenced by wind \cite{Jurado:2021:WindSpeedDirectionAirPollutionConcentrations}. Temperature can influence air pollution through a range of processes such as temperature inversions \cite{wallace:2010:temperatureInversionAirPollution}, O$_3$ production \cite{bloomer:2009:observedOzoneAndTemperatureRelationship}, alongside direct solar radiation \cite{finlayson:1986atmosphericChemistryOzoneProduction}. Temperature can also have impacts indirectly by changing air pollution removal by plants \cite{nowak:1998:VegetationTemperatureAirPollution}. Rainfall drives key air pollution removal processes, such as wet disposition \cite{jolliet:2005:WetDeposition} and wash off from surfaces \cite{yuan:2017:RainfallRoadWashoff, xu:2019:RainfallLeafWashOff}. Pressure can influence air pollution concentrations, with vertical mixing occurring at higher levels in low-pressure systems dispersing air pollutants \cite{ning:2018:LowPressureSystemAirPollution}, alongside the reverse occurring in high-pressure systems \cite{vukovich:1979:HighPressureSystemAirPollution}. O$_3$ production also occurs at higher rates at higher pressures \cite{hippler:1990:OzoneFormationPressure}. The boundary layer further influences vertical mixing \cite{xiang:2019:BoundaryLayerHeightAirPollution}, alongside its height providing the overall atmospheric volume for air pollution to concentrate \cite{davies:2007:BoundaryLayerHeightAirPollution, xiang:2019:BoundaryLayerHeightAirPollution2}. 

Meteorological data were sourced from the ECMWF Re-Analysis Version 5 (ERA5) dataset \cite{hersbach:2016:era5}. ERA5 provides historical estimates for hundreds of environmental phenomena at locations worldwide, encompassing the entire duration of this study. The dataset was spatially interpolated to approximate values for specific locations required in the study, including the sites of actual monitoring stations used in the model's training phase or the positions of synthetic air pollution monitoring stations when generating a comprehensive air pollution map across the globe. The Meteorological dataset family comprised of the features 100m U Component of Wind, 100m V Component of Wind, 10m U Component of Wind, 10m V Component of Wind, 2m Dewpoint Temperature, 2m Temperature, Boundary Layer Height, Downward UV Radiation at Surface, Instantaneous 10m Wind Gust, Surface Pressure and Total Column Rain Water.

\begin{figure}[ht]
    \includegraphics[width=.9\linewidth]{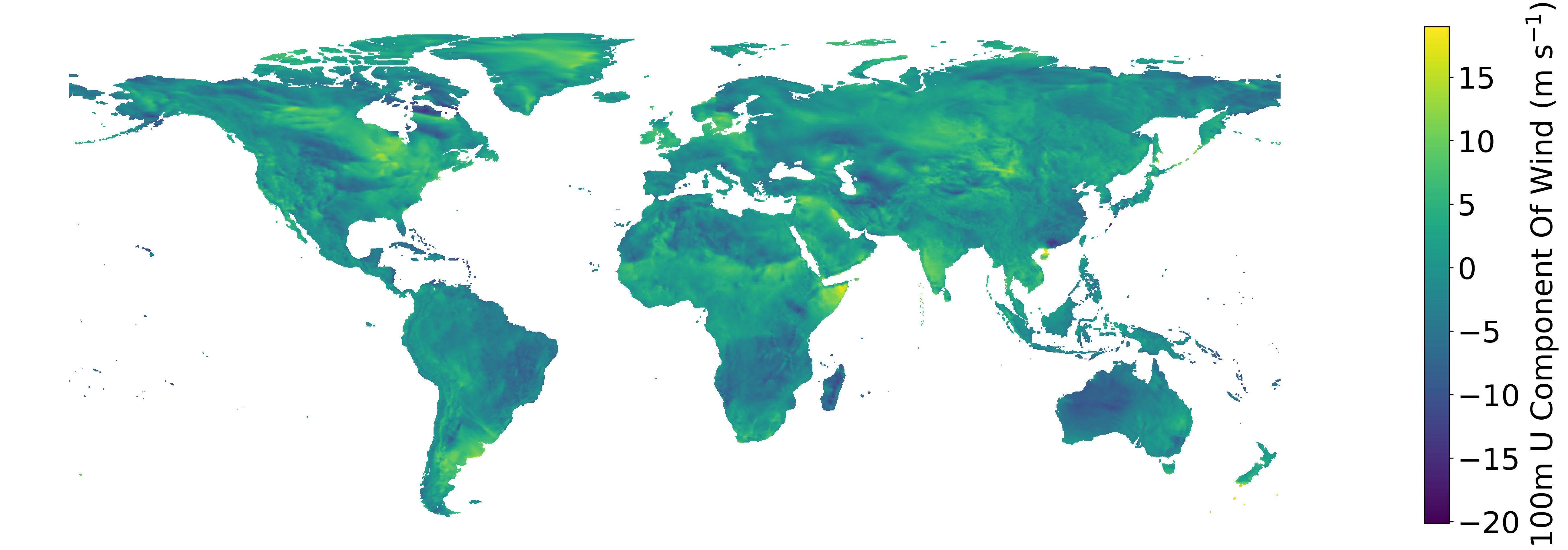}
\caption{{\bfseries 100m U Component of Wind From the Meteorological Dataset Family.}}
\label{fig:meteoroligcalSingleExample}
\end{figure}

\paragraph{Remote Sensing, 4 features.} The Remote Sensing data utilised in this study were obtained from the Sentinel 5P platform for the year 2022, where a monthly average for each location was computed to address the data quality issues outlined in Section \ref{sec:measuringAirPollution}. The selected variables for analysis included NO$_2$, O$_3$, SO$_2$, and the Absorbing Aerosol Index.

\begin{figure}[ht]
    \includegraphics[width=.9\linewidth]{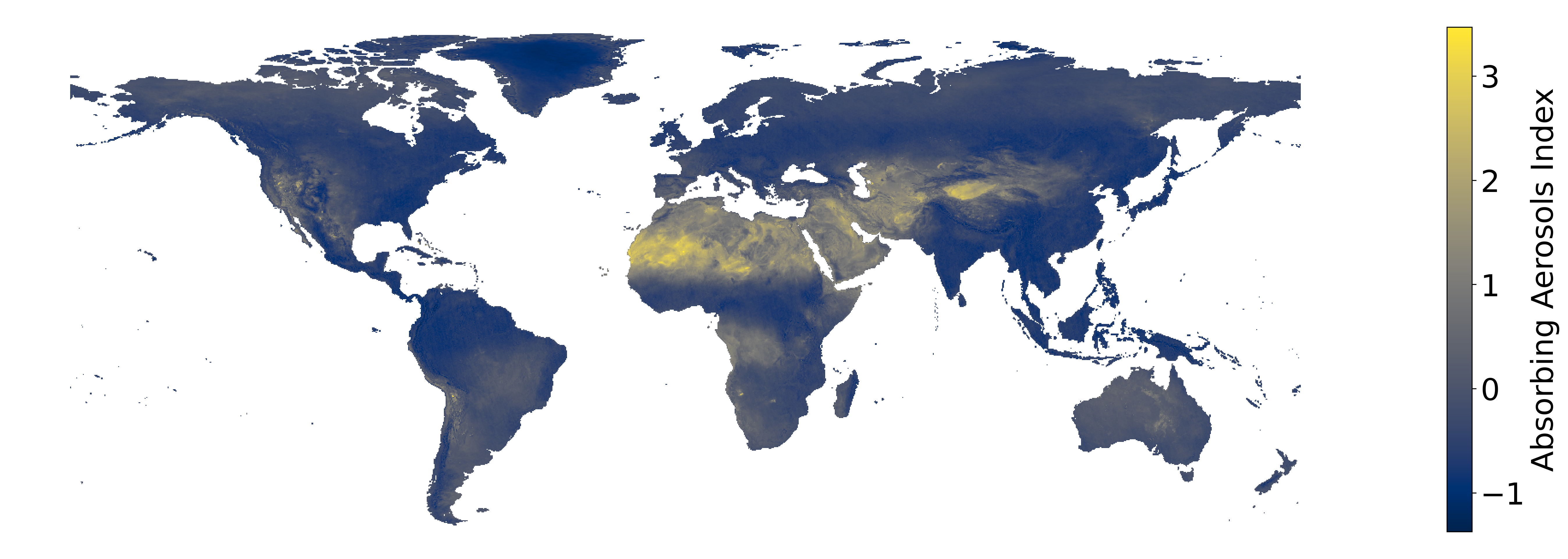}
\caption{{\bfseries Absorbing Aerosol Index From Sentinel 5P for the Remote Sensing Dataset Family.} }
\label{fig:remoteSensingSingleExample}
\end{figure}

\paragraph{Emissions, 6 features. } Emissions from various processes are critical in determining the air pollution concentrations within a specific area. The CAMS global anthropogenic emissions datasets \cite{soulie:2023:EmissionsDatasetAnthrogenic,granier:2019:EmissionsDatasetAnthrogenicCopernicus} from ECCAD \cite{ECCAD:2023:AboutUs} were incorporated to characterise emissions of CO, NO$_x$, NMVOCs, Other-VOCs, and SO$_2$ from the following sectors:

\begin{enumerate}
    \item \textbf{Refineries.} Refineries refer to the industrial facilities for processing crude oil into products such as gasoline, diesel fuel, etc., producing several different air pollutants \cite{adebiyi:2022:EmissionsPetroleumRefining}. 
    
    \item \textbf{Ships.} Ships emit NO$_x$, PM, CO$_2$ and VOCs \cite{corbett:1997:ShipEmissions}, and particularly SO$_x$ from the marine fuels with high sulfur content \cite{tao:2013:ShipHighSulfur}. However, there has been a marked reduction in SO$_x$ as new regulations for sulfur content in marine fuels have emerged recently, substantially reducing emissions \cite{zisi:2021:EUEmissionsImpact}.
    
    \item \textbf{Fugitives.} Fugitive emissions are unintentional and undesirable emissions, leakage or discharges \cite{sotoodeh:2022:FugitiveEmissionsDefinition}. While fugitive emissions are smaller than other sources, they represent a measurable factor in air pollution concentrations \cite{santacatalina:2010:FugitiveEmissionsImpactPM}.
    
    \item \textbf{Power Generation.} The source of energy greatly influences the emissions from the power sector. If the fuel contains sulfur, then when burned, SO$_2$ is produced. Coal, particularly bituminous coal and lignite, contains significant amounts of sulfur \cite{qian:2020:ChinaCoalSO2Emissions}, with the particular amount depending on the origin of the coal \cite{querol:1995:CoalSO2Content}. SO$_2$ can also be emitted from crude oil and oil-derived fuels like diesel, heating and bunker oil containing sulfur \cite{shi:2021:OilSulfurContent}. Likewise, natural gas contains sulfur, albeit at a much lower content than other fuels \cite{chaaban:2004:NaturalGasSulfurEmissions}. Biomass (wood and crop residues etc.) can also produce harmful air pollution such as SO$_2$ \cite{zhang:2007:BiomassHouseholdAirPollutionSulfur}.
    
    \item \textbf{Off Road Transportation.} This refers to any transportation device not used on public roads. Some of the phenomena described by this sector include railway activities emitting PM, NO$_x$, CO$_2$ \cite{fuller:2014:TrainEmissions}, aeroplanes emitting CO$_2$, NO$_x$, SO$_2$, PM, unburned hydrocarbons and black carbon \cite{barrett:2010:AeroplanesEmissions}, alongside agricultural \cite{gonzalez:2016:AgriculturalEquipmentEmissions}, construction \cite{wang:2023:constructionEmissions} and mining equipment \cite{barati:2015:MiningEquipmentEmissions}. 
    
    \item \textbf{Road Transportation.} Road vehicles exhaust gas air pollutants such as CO, CO$_2$, NO$_x$, SO$_2$ \cite{watkins1991:RoadPollutionAirPollution} and PM air pollution \cite{yan:2011:RoadVechilesPMEmissions}.
    
    \item \textbf{Residential.} The residential emissions sector captures residential activities and household sources, including heating \cite{feng:2014:WinterPMResidentialHeating} and cooking \cite{archer:2016:ResidentialCookingImpact}.
    
    \item \textbf{Industrial Process.} They refer to a range of different activities that release emissions, such as chemical manufacturing \cite{LIANG:2020:VOCEmissionsInventoryIndustry, ehrlich:2007:PMChemicalIndustryEmissions, guo:2022:SO2NOXChemical}, cement production \cite{meo:2004:DustCementEmisions}, production of plastics and emissions of VOCs \cite{cabanes:2020:PlasticProductionVOC}.
    
    \item \textbf{Solvent.} Solvents can be used for various processes encompassing manufacturing, printing, automobile, and pharmaceutical industries with associated emissions 
    \cite{yuan:2010:SolventSourceUse}.
    
    \item \textbf{Agricultural Waste Burning.} Waste burning in agriculture can occur for a range of reasons, such as disease control, pest control, crop propagation or crop rotation \cite{lal:2008:AgriculturalWasteBurning}, with implications on air quality \cite{kumar:2013:AgriculturalPollution}. 
    
    \item \textbf{Solid Waste and Waste Water.} Solid waste refers to emissions from processes such as landfills \cite{huber:2008:LandfillBiogenicEmissions, nair:2019:LandfillVOCEmissions}, waste treatment, such as incineration \cite{rani:2008:WasteIncineration}. Wastewater treatment plants also release emissions \cite{law:2012:WasteWaterTreatmentEmissions}.
\end{enumerate}
 
In addition to anthropogenic sources of emissions, biogenic emissions of CO are also accounted for using the CAMS Global Biogenic Emissions dataset \cite{sindelarova:2022:BIogenicEmissionsOne, sindelarova:2014:BIogenicEmissionsTwo, granier:2019:BiogenicEmissionsThree}, which captures emissions from organic matter \cite{hudman:2008:biogenicCOEmissions}.

\begin{figure}[ht]
    \includegraphics[width=.9\linewidth]{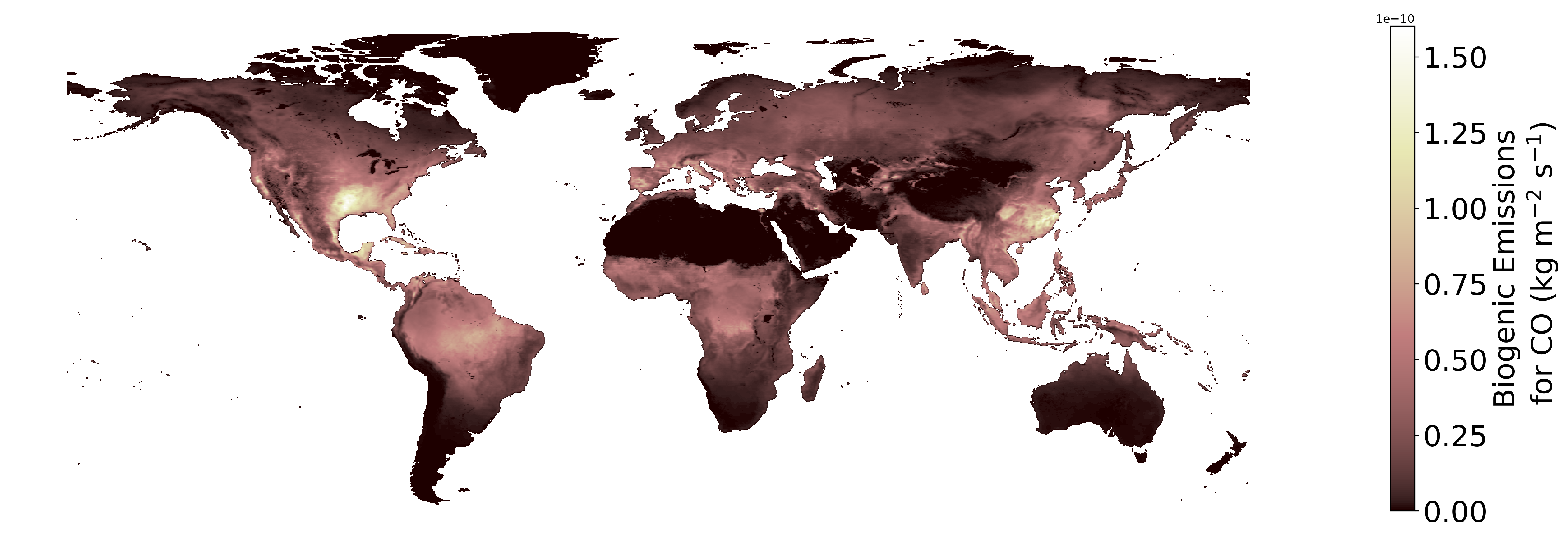}
\caption{{\bfseries Biogenic Emissions Example for the Emissions Dataset Family.}} \label{fig:emissionsSingleExample}
\end{figure}

\section{Models}
\label{sec:models}

\subsection{Model Design and Thinking}

LightGBM \cite{ke:2017:lightgbmDefinition} was chosen as the machine learning approach for the problem, having been used successfully in our previous work in estimating air pollution concentrations across England. LightGBM is a gradient-boosting decision tree (GBDT) algorithm where an ensemble of decision trees are trained in a sequence, with the $n+1$ decision tree fitting the residuals of the first n decision trees, learning the difference between the actual target vector and the weighted sum of predictions of the first $n$ decision trees.

Air pollution concentration datasets frequently contain various anomalies and outliers, a challenge that persists across both individual monitoring stations and wider networks \cite{van:2018:AirPollutionOutlier,rollo:2023:AirPollutionAnaomaly}. This issue is exacerbated in global datasets encompassing multiple monitoring station networks. Consequently, the selection of a model robust against both outliers and anomalies was crucial. LightGBM addresses this challenge effectively through its use of decision trees as the foundational learning mechanism. These decision trees aim to cluster homogeneous data instances, facilitated by the ``minimum data in leaf'' parameter. This parameter sets a threshold for the smallest number of data instances that constitute a valid leaf, thereby ensuring that the LightGBM algorithm only considers a sufficiently large and homogenous set of data points for learning and prediction. For example, in the context of air pollution prediction, an anomalous observation—such as an unusually high pollution reading coinciding with high wind speed and remote sensing data, which occurs only once in the dataset—would not lead to the formation of a dedicated leaf. This scenario could arise from a transient pollution source, like a vehicle passing near a monitoring station, resulting in a measurement that inaccurately reflects the broader area's air quality.

A further benefit of LightGBM over other models is the approach taken to building the model. The approach that LightGBM takes when building the decision trees is to split observations based on the feature vector values, looking for the best possible split regarding information gain and reducing the uncertainty regarding the target vector, grouping homogenous instances of data points, such as the instances where there is high wind speed at monitoring station that is measuring low concentration readings for NO$_2$. One of the core issues with our air pollution training data is that many data points within the datasets repeat the same information due to the cyclical nature of air pollution measurements, causing a considerable amount of bloat in the datasets. The standard approach to identifying split points within a GBDT is the pre-sorted algorithm where all possible split points are explored, an approach which, in this use case, would be highly costly regarding computation and memory. LightGBM helps tackle this issue by using histograms when performing the splits, where continuous variables are put into discrete bins, changing the computational cost from dependent on the number of data points to the number of discrete bins created.

A key consideration during the model design was the choice of the loss function. The loss function represents the error of a given prediction, in this case, quantifying the difference between the actual air pollution concentration and prediction of the model, thereby allowing for comparisons between model variations and subsequent choice of the optimal model configuration. The choice of the loss function in this situation was between the mean absolute error (MAE) and the mean squared error (MSE) \cite{hodson:2022:RMSEMAE}. The MAE would help reduce the influence of higher air pollution measurements on the model present due to the known presence of outliers and anomalies within the dataset. However, these high air pollution measurements are of vital interest within the context of air pollution, even if they are potentially erroneous. So, a tradeoff of potentially overfitting on these higher values was seen as a worthwhile tradeoff, and as such, the MSE was chosen as the loss function. The underlying premise is that 10\si{\micro\gram/\meter^3} is more than twice as bad on human health than 5\si{\micro\gram/\meter^3}, so using the MSE is more appropriate given the domain in which the model would be used. Supporting the existing literature understanding that there is a non-linear relationship between the detrimental effect of air pollution concentrations increases and damage to human health and well being \cite{yang:2022:nonlinearAirPollution, zhao:2019:nonlinearAirPollution2}.

As an air pollution concentration can never measure less than zero, we trained the model on the log transformation of the target vector. We added a small constant of 1 × 10$^{-7}$ to the target vector and then performed the log transformation due to 0 concentration measurements within the dataset. The log transformation and addition ensured that the model would never predict a negative value as the model output, as the reverse transformation of calculating the exponential and subtracting 1 × 10$^{-7}$ was performed on the output. A further hyperparameter explored during model training was L2 regularization \cite{hoerl:1970:RidgeRegression}. Including L2 regularization helps distribute the weights within the decision tree, encouraging the weights to be closer to 0 but keeping all feature vectors, ensuring that no single feature vector drives predictions, key with the considerable number of feature vectors used.

The framework for choosing the model's hyperparameters was a randomized grid search of 5 parameter sets. The parameters we optimized during the randomized search were the L2 regularisation and the min data in each leaf already discussed, alongside the number of leaves, the number of trees and the max depth \cite{LightGBM:2023:Parameters}. The number of leaves search space was given the range of 1000 to 4095 \cite{Github:2017:ParameterGridSuggestions}, with the optimal values being chosen near the centre of this range, validating its choice. The number of trees was controlled via early stopping, where no additional tree would be added after 10 trees had been added without any improvement in the loss function performance. Similarly, the max depth was not limited and left to grow as needed until performance did not improve during training.

Some model parameters were kept constant throughout the search, such as the max bin, kept constant at 63. The max bin refers to the number of discrete bins created for a continuous feature vector. 63 was chosen to ensure that a range of different splits during model training could be created while also helping to reduce training time by allowing data to be stored optimally as an int8 data type, with minimal reduction in model accuracy \cite{LightGBM:2023:Parameters}. The boosting type used during training was Gradient-based One-Side Sampling (GOSS) \cite{ke:2017:lightgbmDefinition}. GOSS is a method of boosting that allows the n+1 decision tree discussed at the start of this section to be trained on a subsample of the data. The subsample of data chosen is the data that has a large gradient, i.e. the data the model has yet to learn well from and a random sample of the small gradient data, helping to reduce the amount of data used drastically, and therefore training time. The tradeoff with GOSS is the potential for overfitting when the datasets are small; however, this was not a concern in the context of air pollution dataset used for training in this study. 

The data was split randomly into 70\% training, 20\% validation and 10\% test sets. The data was split into sets by the magnitude of the concentrations, based on the UK Daily Air Quality Index (DAQI) \cite{UKAIR:2023:DAQI}. Each index of the DAQI data points was put equally into each set to ensure that potential outliers or anomalous points existed in each of the three datasets; if less than 3 data points for a particular DAQI band were in the monitoring station data, then all the data points would be put into the training set. Hold-out validation was acceptable due to the considerable dataset size and the minimal gain in overall performance at the cost of considerable additional computational cost \cite{Ng:2020:CrossValidation}. We chose the best parameter set based on the model's MSE on the validation set across the parameter sets. 

As we wanted to allow the possibility of extending the model as new data becomes available, we did not include any feature vector element detailing monitoring station identifiers, such as the name or location where the observation was measured. A further consideration was the lack of inclusion of lags of the air pollution concentration, such as using the concentration at T-1 to estimate the concentration at time T. Thereby allowing us to make estimations where there has never been an air pollution measurement. Further, this allows us to mix observations where subsequent observations were treated as independent. This allows the data to be put into a tabular format, where LightGBM has state-of-the-art performance. Together, these elements of temporal and spatial independence of observations allow the creation of a lightweight model where parallel computation is possible of different locations and time points.

\subsection{Predictive Capability}
\label{sec:predicitiveCapability}
The first experiment conducted during the study aimed to look at how well a data-driven supervised machine-learning model could learn the relationship between air pollution concentrations and the feature vectors described in Section \ref{sec:featureVectors}. Table \ref{tab:temporalR2Scores} shows the number of monitoring stations with a positive R$^2$ score, representing a prediction by the model of the time series that provides additional information overestimating the mean of the time series \cite{scikit:2023:R2Score}. It can be seen that the model works well for nearly all of the air pollution monitoring stations included in the study. The reduced performance for SO$_2$ can be explained by the limited amount of data, with only 1446 monitoring stations.

\begin{figure}[!htb]
    \hspace*{\fill}   
  \begin{subfigure}{0.99\textwidth}
    \includegraphics[width=\linewidth]{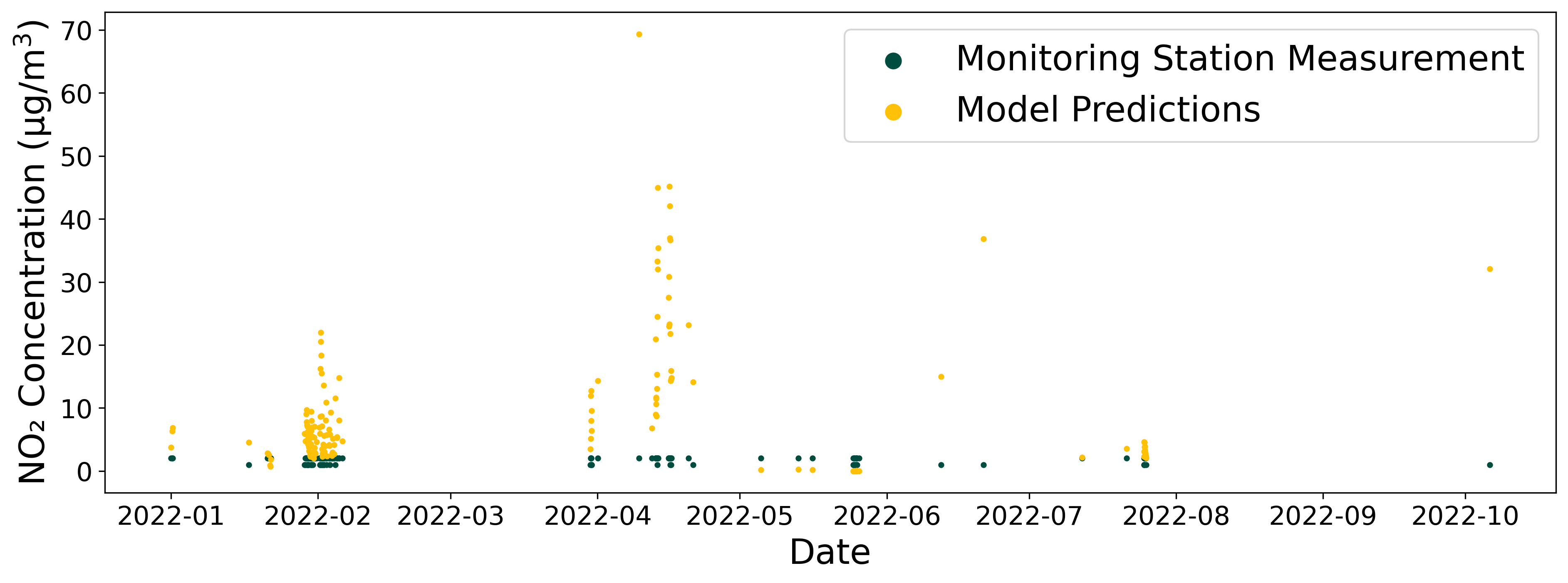}
    \caption{Monitoring Station CAAQM 8171. R$^2$ -741.} \label{fig:temporalIndividualPredicitionsCAAQM8171}
  \end{subfigure}%
  \hspace*{\fill}   
  \\
   \hspace*{\fill}   
  \begin{subfigure}{0.99\textwidth}
    \includegraphics[width=\linewidth]{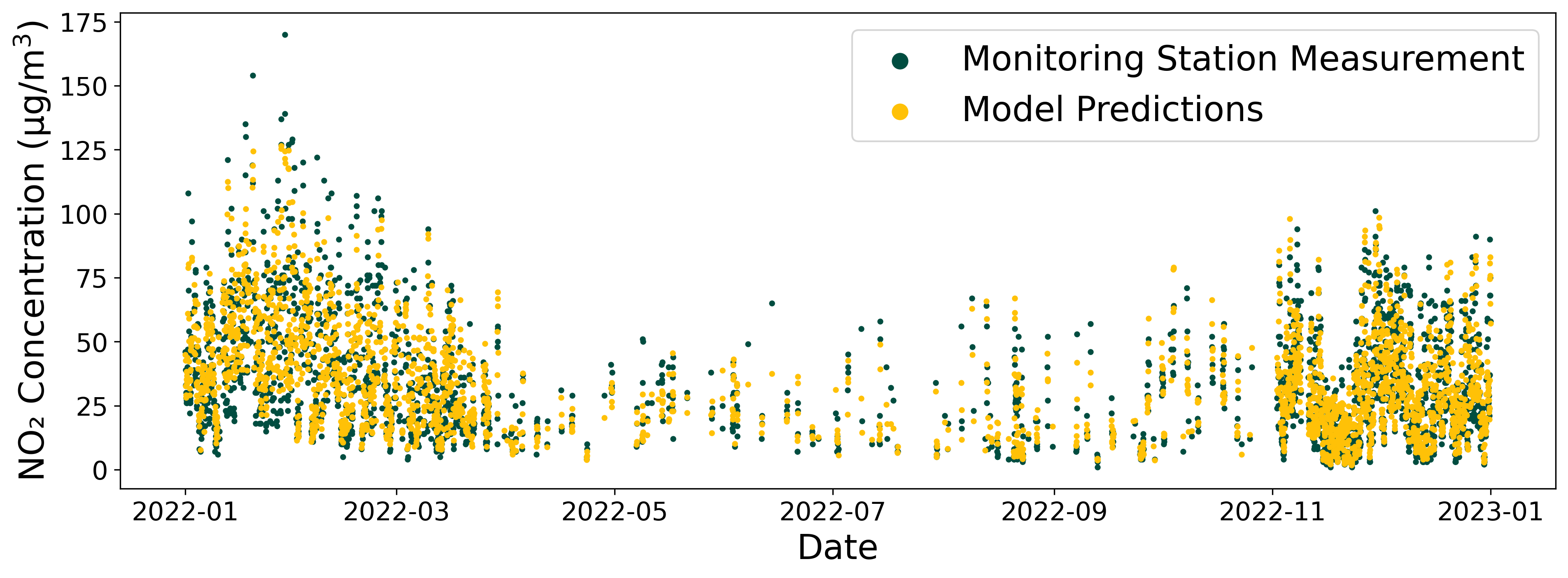}
    \caption{Monitoring Station EEA Spain 4327. R$^2$ 0.81.} \label{fig:temporalIndividualPredicitionsEEASpain4327}
  \end{subfigure}%
  \hspace*{\fill}   
  \\
\caption{{\bfseries Model Predictions for a Well-Performing (EEA Spain 4327) and a Poor-Performing (CAAQM 8171) Monitoring Station During the Baseline Experiment.} While the model does not perform well for CAAQM 8171, looking at the data does raise questions about its validity, with the NO$_2$ concentrations having poor precision with only integers recorded and never exceeding 2\si{\micro\gram/\meter^3}, highly unlikely given the industrial location of Nandesari, India. In contrast, EEA Spain 4327 performs very well, with data that appears accurate. Highlighting the contrast in the quality of data in the dataset used and the importance of performing a baseline experiment rather than simply assuming all monitoring stations represent accurate data. } \label{fig:temporalIndividualPredicitions}
\end{figure}

Further, even with the criteria used to subset the dataset outlined in Section \ref{sec:targetVector}, some monitoring stations with questionable data are included in the study, such as an Indian air pollution monitoring station, denoted CAAQM 8171 in the OpenAQ dataset maintained by the Central Pollution Control Board of India \cite{CPCB:2023:AirPollutionData}, for which the model achieved an R$_2$ score of -741. The measurements from the monitoring station can be seen in Figure \ref{fig:temporalIndividualPredicitionsCAAQM8171}, where the measurements are simply 129 2\si{\micro\gram/\meter^3}, 40 1\si{\micro\gram/\meter^3}, and 2 0\si{\micro\gram/\meter^3} measurements. The first cause for concern is the resolution of the measurements; however, further issues are raised when looking at the location of the monitoring station, at longitude 73.1 and latitude 22.4, denoting Nandesari, India, an urban area with considerable industrial activities. The measurements may be correct; however, they appear unlikely, given the local environment around the monitoring station. As such, the broader context of potentially problematic data must be considered when analysing the results. In contrast to CAAQM 8171, EEA Spain 4327 shown in Figure \ref{fig:temporalIndividualPredicitionsEEASpain4327} provides an example of a monitoring station with considerably more complex measurements, with the model accurately capturing the time series, resulting in an R$_2$ score of 0.81. Therefore, in this manuscript's context, these initial experiments act as a baseline for later experiments.

\begin{table}[!htb]
\resizebox{\linewidth}{!}{
\begin{filecontents*}{CSVs/temporal.csv}
,Pollutant,Number of Stations,Asia,Australia,South America,Africa,Europe,North America,Oceania,Antarctica
0,NO₂,2613 / 2894 (90\%),201 / 295 (68\%),-,12 / 12 (100\%),53 / 55 (96\%),2345 / 2527 (93\%),2 / 5 (40\%),-,-
1,O₃,1904 / 2076 (92\%),187 / 273 (68\%),-,7 / 10 (70\%),48 / 50 (96\%),1658 / 1739 (95\%),4 / 4 (100\%),-,-
2,PM₁₀,2780 / 2924 (95\%),439 / 520 (84\%),49 / 53 (92\%),94 / 101 (93\%),59 / 75 (79\%),1779 / 1804 (99\%),357 / 368 (97\%),3 / 3 (100\%),-
3,PM₂₅,2573 / 2939 (88\%),337 / 433 (78\%),13 / 63 (21\%),97 / 114 (85\%),62 / 77 (81\%),1071 / 1127 (95\%),993 / 1111 (89\%),0 / 14 (0\%),-
4,SO₂,882 / 1446 (61\%),193 / 280 (69\%),-,47 / 66 (71\%),30 / 51 (59\%),611 / 1048 (58\%),1 / 1 (100\%),-,-
\end{filecontents*}
\pgfplotstabletypeset[
    multicolumn names=l, 
    col sep=comma, 
    string type, 
    header = has colnames,
    columns={Pollutant,Number of Stations, Asia, Australia, South America, Africa, Europe, North America, Oceania},
    columns/Pollutant/.style={column type=l, column name=Pollutant Name},
    columns/Number of Stations/.style={column type=l, column name=Number of Stations},
    columns/Asia/.style={column type=l, column name=Asia},
    columns/Australia/.style={column type=l, column name=Australia},
    columns/South America/.style={column type=l, column name=South America},
    columns/Africa/.style={column type=l, column name=Africa},
    columns/Europe/.style={column type=l, column name=Europe},
    columns/North America/.style={column type=l, column name=North America},
    columns/Oceania/.style={column type=l, column name=Oceania},
    every head row/.style={before row=\toprule, after row=\midrule},
    every last row/.style={after row=\bottomrule}
    ]{CSVs/temporal.csv}}
    \smallskip
\caption{{ \bfseries Number of Monitoring Stations With Positive R$^2$ Scores With a Model Trained on All Data.} The table shows a baseline set of results for the model's predictive capability to learn the relationship between the feature and target vector as a comparison for future experiments. The baseline aims to accurately represent the data quality used, highlighting the number of stations with questionable data, such as CAAQM 8171 shown in Figure \ref{fig:temporalIndividualPredicitionsCAAQM8171} that likely represents incorrect data. The results show that the model can capture the relationship across the monitoring stations while not being limited to performing well in a single continent. } \label{tab:temporalR2Scores}
\end{table}

\subsection{Estimating Missing Stations}

With it clear that the machine learning model framework proposed can learn the relationship between the outlined feature vectors described in Section \ref{sec:featureVectors} and air pollution concentrations, the following set of experiments explore the ability of the model to estimate missing locations: monitoring stations data that have been held back and an assessment done on the ability to accurately predict air pollution concentrations in a location never before seen, evaluating the performance of the model in a use case of filling in missing locations air pollution concentrations. 

\subsubsection{Within Network}

\begin{figure}
    \includegraphics[width=\linewidth]{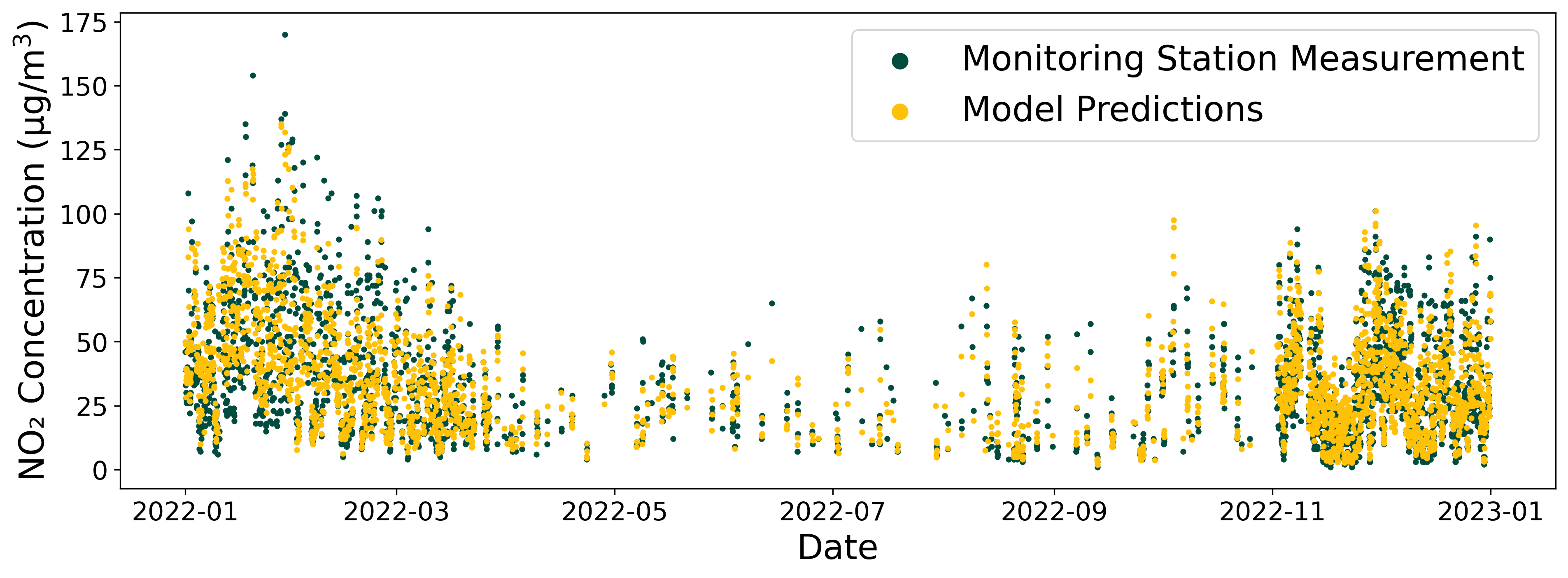}
\caption{{\bfseries Monitoring Station EEA Spain 4327. R$^2$ Score 0.78, for the Within Network Experiment.} For EEA Spain 4327, the model still performs well when it has not seen any of the data from this location, with the R$^2$ reducing by 0.03. The slight drop in performance highlights that the model can estimate missing locations well in the case of the EEA Spain monitoring network when it has seen some monitoring stations within the network. } \label{fig:monitoringStationEEASpain4327SpatialKFold}
\end{figure}

Within the air pollution concentration datasets are several different monitoring station networks. The first missing station experiment aims to understand the model's performance when estimating monitoring stations that it has never seen before, but it has seen data from other monitoring stations within the network. A 10-fold leave-one-out validation was conducted. Each monitoring network had its monitoring stations split into ten groups, with a model trained on nine groups and the final group having their measurements estimated and then assessed. As seen in Table \ref{tab:spatialKFold}, there is a drop in the model's performance when estimating monitoring stations it hasn't seen before. Still, there is a signal that the model works as intended and can estimate the air pollution concentrations. For example, Figure \ref{fig:monitoringStationEEASpain4327SpatialKFold} shows monitoring stations EEA Spain 4327 for this experiment, which experienced a performance drop of 0.03 in R$^2$ score compared with the baseline. 

\begin{table}[!htb]
\resizebox{\linewidth}{!}{
\begin{filecontents*}{CSVs/spatial_kfold.csv}
,Pollutant,Number of Stations,Asia,Australia,South America,Africa,Europe,North America,Oceania,Antarctica
0,NO₂,860 / 2894 (30\%),37 / 295 (13\%),-,1 / 12 (8\%),6 / 55 (11\%),815 / 2527 (32\%),1 / 5 (20\%),-,-
1,O₃,1495 / 2076 (72\%),50 / 273 (18\%),-,3 / 10 (30\%),24 / 50 (48\%),1418 / 1739 (82\%),0 / 4 (0\%),-,-
2,PM₁₀,388 / 2924 (13\%),85 / 520 (16\%),0 / 53 (0\%),3 / 101 (3\%),1 / 75 (1\%),286 / 1804 (16\%),13 / 368 (4\%),0 / 3 (0\%),-
3,PM₂₅,873 / 2939 (30\%),177 / 433 (41\%),3 / 63 (5\%),29 / 114 (25\%),5 / 77 (6\%),457 / 1127 (41\%),202 / 1111 (18\%),0 / 14 (0\%),-
4,SO₂,6 / 1446 (0\%),2 / 280 (1\%),-,0 / 66 (0\%),0 / 51 (0\%),4 / 1048 (0\%),0 / 1 (0\%),-,-
\end{filecontents*}
\pgfplotstabletypeset[
    multicolumn names=l, 
    col sep=comma, 
    string type, 
    header = has colnames,
    columns={Pollutant,Number of Stations, Asia, Australia, South America, Africa, Europe, North America, Oceania},
    columns/Pollutant/.style={column type=l, column name=Pollutant Name},
    columns/Number of Stations/.style={column type=l, column name=Number of Stations},
    columns/Asia/.style={column type=l, column name=Asia},
    columns/Australia/.style={column type=l, column name=Australia},
    columns/South America/.style={column type=l, column name=South America},
    columns/Africa/.style={column type=l, column name=Africa},
    columns/Europe/.style={column type=l, column name=Europe},
    columns/North America/.style={column type=l, column name=North America},
    columns/Oceania/.style={column type=l, column name=Oceania},
    every head row/.style={before row=\toprule, after row=\midrule},
    every last row/.style={after row=\bottomrule}
    ]{CSVs/spatial_kfold.csv}}
    \smallskip
\caption{{ \bfseries Number of Monitoring Stations With Positive R$^2$ Scores With A Model Trained on Some Stations in the Same Monitoring Network.} The first missing station experiment explored the model's performance when estimating a given station when no data concerning that station has been seen. The model has only seen data from a portion of the target monitoring station network, with no data from the particular monitoring station of interest. It can be seen that the relationship between continents broadly remains the same as Table \ref{tab:temporalR2Scores}. However, the number of stations is significantly reduced. Of note is that the results between air pollutants are to be expected with O$_3$ performing the best, being driven primarily by meteorological conditions, the highest spatial and temporal resolution and quality feature vector used by the model. } \label{tab:spatialKFold}
\end{table}

\subsubsection{Between Countries}
\label{sec:betweenCountries}

\begin{figure}
    \includegraphics[width=\linewidth]{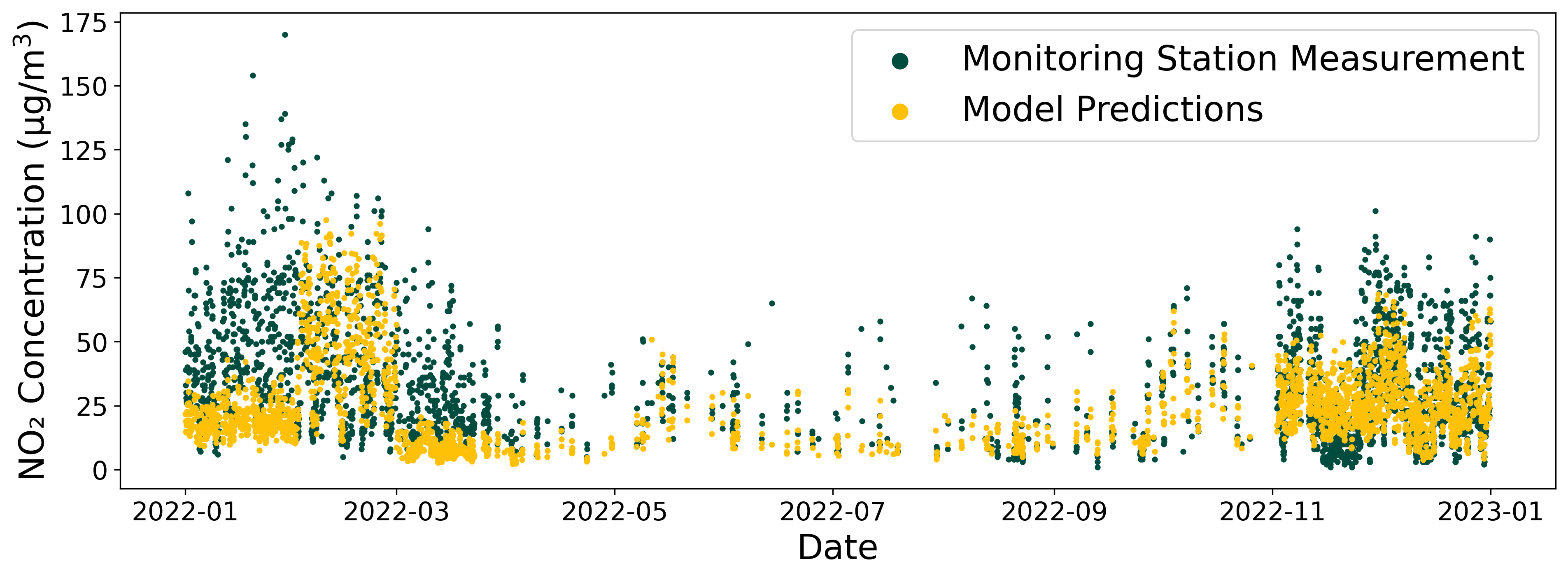}
\caption{{\bfseries Monitoring Station EEA Spain 4327. R$^2$ score -0.03, for the Between Country Experiment.} When the model has not seen any data for Spain, the estimation for EEA Spain 4327 performs poorly, with a negative R$^2$ score, representing a prediction worse than the mean of the time series.} \label{fig:monitoringStationEEASpain4327SpatialCountry}
\end{figure}

The second missing station experiment performed leave-one-out validation based on the country of the monitoring stations. In this case, the experiment aims to understand the model's performance when estimating a country's air pollution concentrations when it has not seen any data concerning the country, exploring, for example, the ability to estimate Spain's monitoring station measurements when only stations outside Spain have been used for model training. Figure \ref{fig:monitoringStationEEASpain4327SpatialCountry} shows the same monitoring station of EEA Spain 4327 as before; however, the R$^2$ has now dropped considerably to -0.03, representing a model that is estimating worse than the average of the time series. 

\begin{table}[!htb]
\resizebox{\linewidth}{!}{
\begin{filecontents*}{CSVs/spatial_country.csv}
,Pollutant,Number of Stations,Asia,Australia,South America,Africa,Europe,North America,Oceania,Antarctica
0,NO₂,356 / 2894 (12\%),4 / 295 (1\%),-,0 / 12 (0\%),0 / 55 (0\%),352 / 2527 (14\%),0 / 5 (0\%),-,-
1,O₃,1242 / 2076 (60\%),22 / 273 (8\%),-,0 / 10 (0\%),15 / 50 (30\%),1205 / 1739 (69\%),0 / 4 (0\%),-,-
2,PM₁₀,173 / 2924 (6\%),6 / 520 (1\%),0 / 53 (0\%),0 / 101 (0\%),0 / 75 (0\%),167 / 1804 (9\%),0 / 368 (0\%),0 / 3 (0\%),-
3,PM₂₅,420 / 2939 (14\%),109 / 433 (25\%),2 / 63 (3\%),0 / 114 (0\%),0 / 77 (0\%),273 / 1127 (24\%),36 / 1111 (3\%),0 / 14 (0\%),-
4,SO₂,9 / 1446 (1\%),0 / 280 (0\%),-,0 / 66 (0\%),0 / 51 (0\%),9 / 1048 (1\%),0 / 1 (0\%),-,-
\end{filecontents*}
\pgfplotstabletypeset[
    multicolumn names=l, 
    col sep=comma, 
    string type, 
    header = has colnames,
    columns={Pollutant,Number of Stations, Asia, Australia, South America, Africa, Europe, North America, Oceania},
    columns/Pollutant/.style={column type=l, column name=Pollutant Name},
    columns/Number of Stations/.style={column type=l, column name=Number of Stations},
    columns/Asia/.style={column type=l, column name=Asia},
    columns/Australia/.style={column type=l, column name=Australia},
    columns/South America/.style={column type=l, column name=South America},
    columns/Africa/.style={column type=l, column name=Africa},
    columns/Europe/.style={column type=l, column name=Europe},
    columns/North America/.style={column type=l, column name=North America},
    columns/Oceania/.style={column type=l, column name=Oceania},
    every head row/.style={before row=\toprule, after row=\midrule},
    every last row/.style={after row=\bottomrule}
    ]{CSVs/spatial_country.csv}}
    \smallskip
\caption{{ \bfseries Number of Monitoring Stations With Positive R$^2$ Scores With a Model Trained on Stations Only Outside the Target Country.} The between-country experiment repeats the same process as the within-network experiment shown in Table \ref{tab:spatialKFold} to explore the possibility of estimating given countries monitoring stations data without any in country data being seen by the model. It can be seen that the model performance drops considerably. However, there is still a positive signal for each air pollutant other than SO$_2$.} \label{tab:spatailCountry}
\end{table}

\subsubsection{Between Continents}

\begin{figure}
    \includegraphics[width=\linewidth]{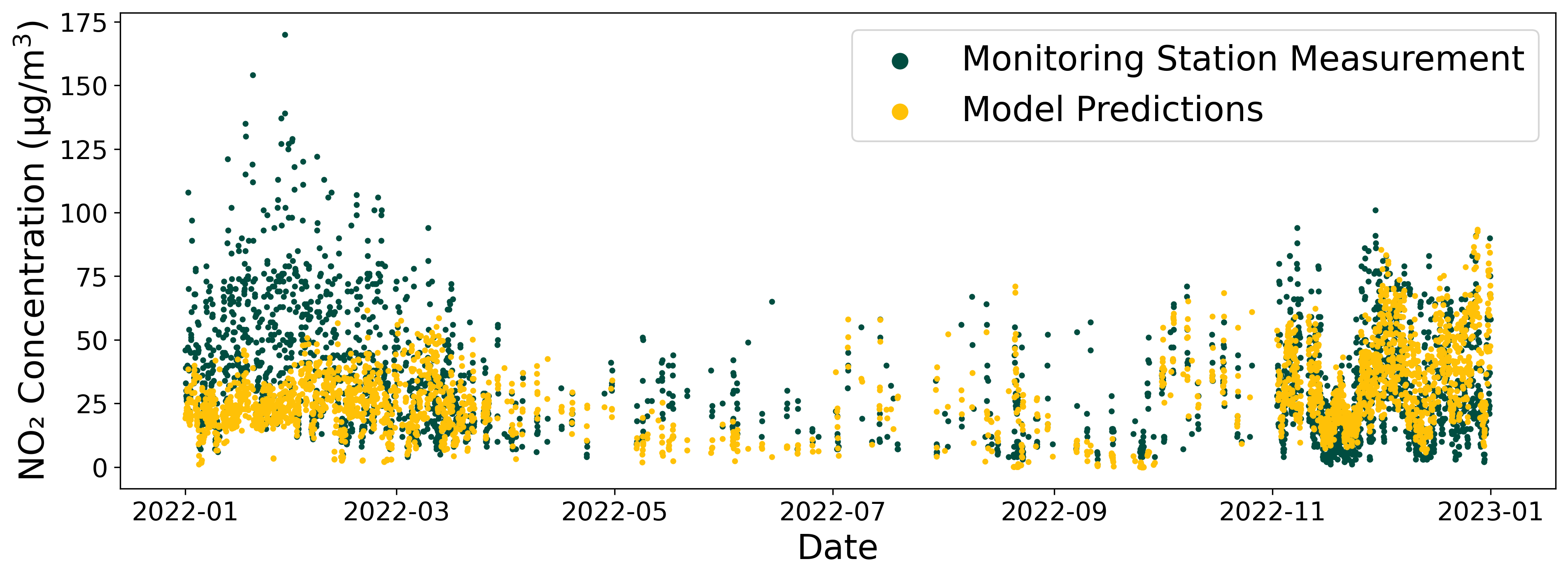}
\caption{{\bfseries Monitoring Station EEA Spain 4327. R$^2$ Score -0.07 for the Between-Continent Experiment.}  The model's performance for this station remains broadly the same, with a drop of 0.04, highlighting that for EEA Spain 4327, the drop in performance occurs from not having any data in the country. Still, data further removed does not further meaningfully reduce the performance.} \label{fig:monitoringStationEEASpain4327SpatialContinent}
\end{figure}

The final missing station experiment repeated the same process as \ref{sec:betweenCountries} but this time with continents, aiming to determine the ability to estimate Asia's monitoring stations measurements when only trained on data from North and South America, Europe, Australia and Oceania. Figure \ref{fig:monitoringStationEEASpain4327SpatialContinent} shows monitoring station EEA Spain 4327, where the R$^2$ performs marginally worse than when the between countries missing station experiment discussed before. 

\begin{table}[!htb]
\resizebox{\linewidth}{!}{
\begin{filecontents*}{CSVs/spatial_continent.csv}
,Pollutant,Number of Stations,Asia,Australia,South America,Africa,Europe,North America,Oceania,Antarctica
0,NO₂,154 / 2894 (5\%),7 / 295 (2\%),-,0 / 12 (0\%),0 / 55 (0\%),147 / 2527 (6\%),0 / 5 (0\%),-,-
1,O₃,78 / 2076 (4\%),18 / 273 (7\%),-,0 / 10 (0\%),1 / 50 (2\%),59 / 1739 (3\%),0 / 4 (0\%),-,-
2,PM₁₀,16 / 2924 (1\%),14 / 520 (3\%),0 / 53 (0\%),0 / 101 (0\%),0 / 75 (0\%),1 / 1804 (0\%),1 / 368 (0\%),0 / 3 (0\%),-
3,PM₂₅,8 / 2939 (0\%),2 / 433 (0\%),1 / 63 (2\%),0 / 114 (0\%),0 / 77 (0\%),3 / 1127 (0\%),2 / 1111 (0\%),0 / 14 (0\%),-
4,SO₂,5 / 1446 (0\%),0 / 280 (0\%),-,0 / 66 (0\%),0 / 51 (0\%),5 / 1048 (0\%),0 / 1 (0\%),-,-
\end{filecontents*}
\pgfplotstabletypeset[
    multicolumn names=l, 
    col sep=comma, 
    string type, 
    header = has colnames,
    columns={Pollutant,Number of Stations, Asia, Australia, South America, Africa, Europe, North America, Oceania},
    columns/Pollutant/.style={column type=l, column name=Pollutant Name},
    columns/Number of Stations/.style={column type=l, column name=Number of Stations},
    columns/Asia/.style={column type=l, column name=Asia},
    columns/Australia/.style={column type=l, column name=Australia},
    columns/South America/.style={column type=l, column name=South America},
    columns/Africa/.style={column type=l, column name=Africa},
    columns/Europe/.style={column type=l, column name=Europe},
    columns/North America/.style={column type=l, column name=North America},
    columns/Oceania/.style={column type=l, column name=Oceania},
    every head row/.style={before row=\toprule, after row=\midrule},
    every last row/.style={after row=\bottomrule}
    ]{CSVs/spatial_continent.csv}}
    \smallskip
\caption{{ \bfseries Number of Monitoring Stations with Positive R$^2$ Scores With a Model Trained on Stations Only Outside the Target Continent.} The between-continent experiment results show a clear drop in performance compared to the between-country. Highlighting the difference between air pollution concentrations between countries and the need for similar countries to be used to train the model to predict concentrations accurately. } \label{tab:spatialContinents}
\end{table}

\subsection{Overall Trends}

\begin{figure}[!htb]
  \begin{subfigure}{0.49\textwidth}
    \includegraphics[width=\linewidth]{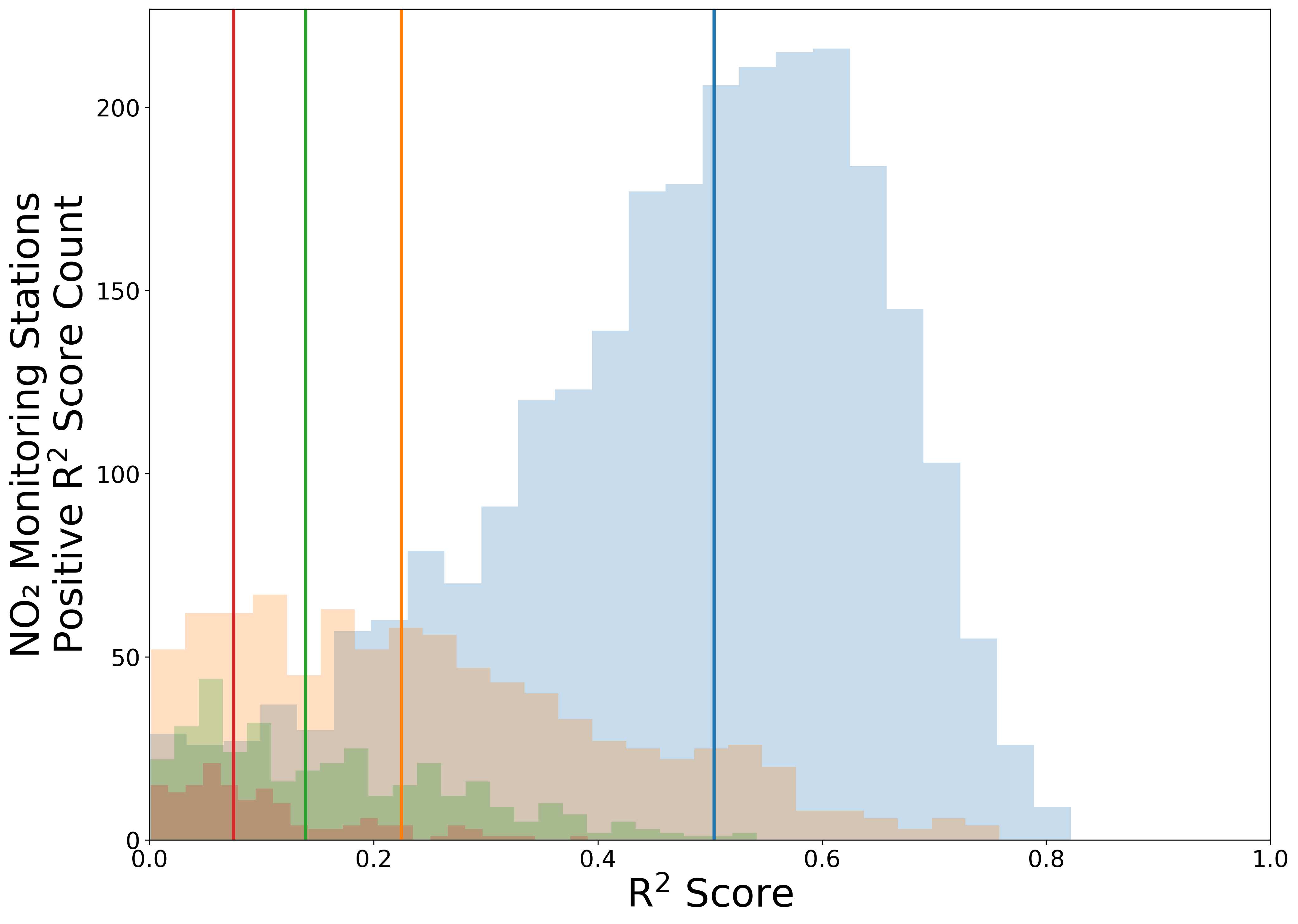}
    \caption{NO$_2$} \label{fig:temporalNO2}
  \end{subfigure}%
  \hspace*{\fill}   
   \hspace*{\fill}   
  \begin{subfigure}{0.49\textwidth}
    \includegraphics[width=\linewidth]{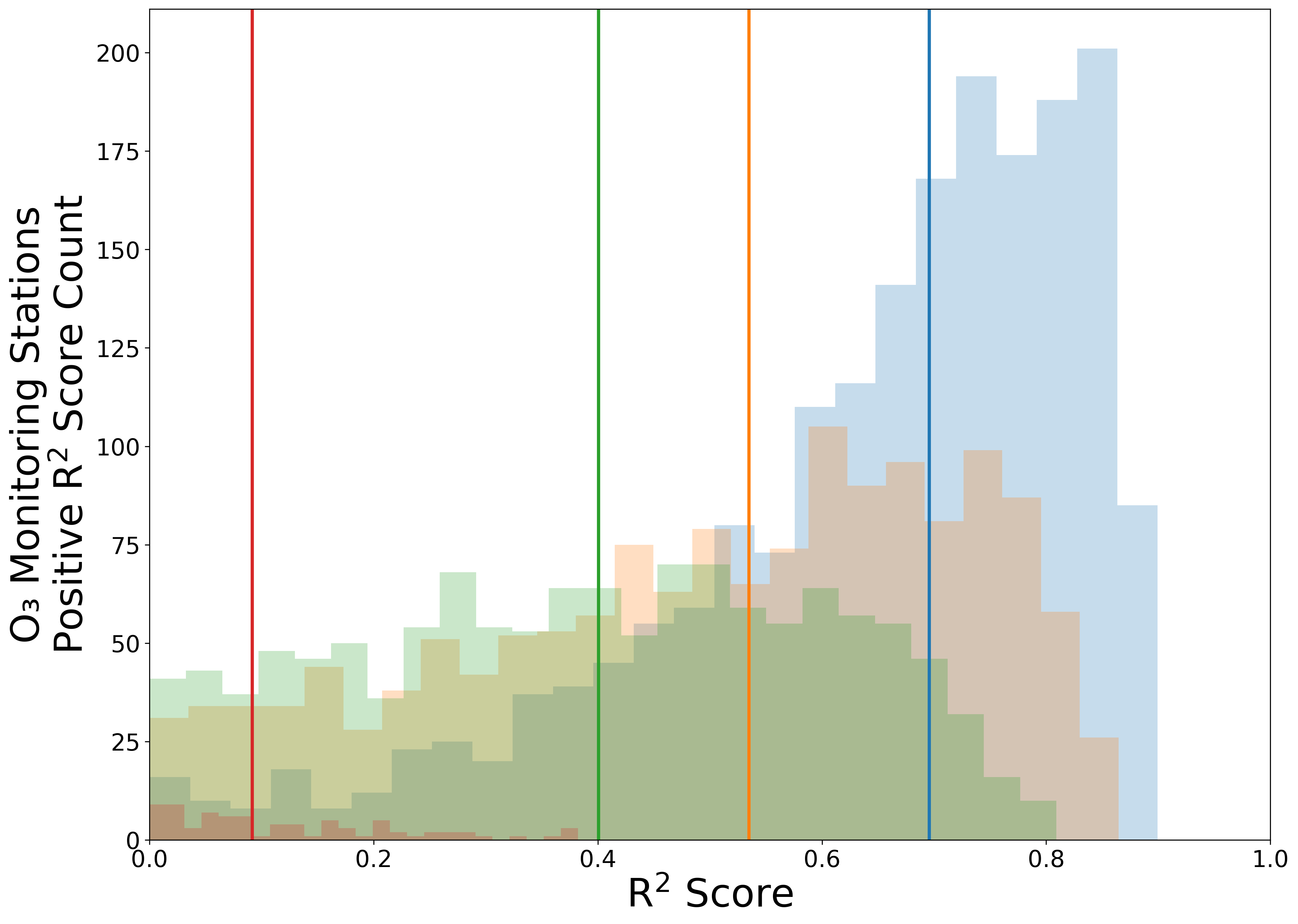}
    \caption{O$_3$} \label{fig:temporalO3}
  \end{subfigure}%
  \hspace*{\fill}   
  \\
  \begin{subfigure}{0.49\textwidth}
    \includegraphics[width=\linewidth]{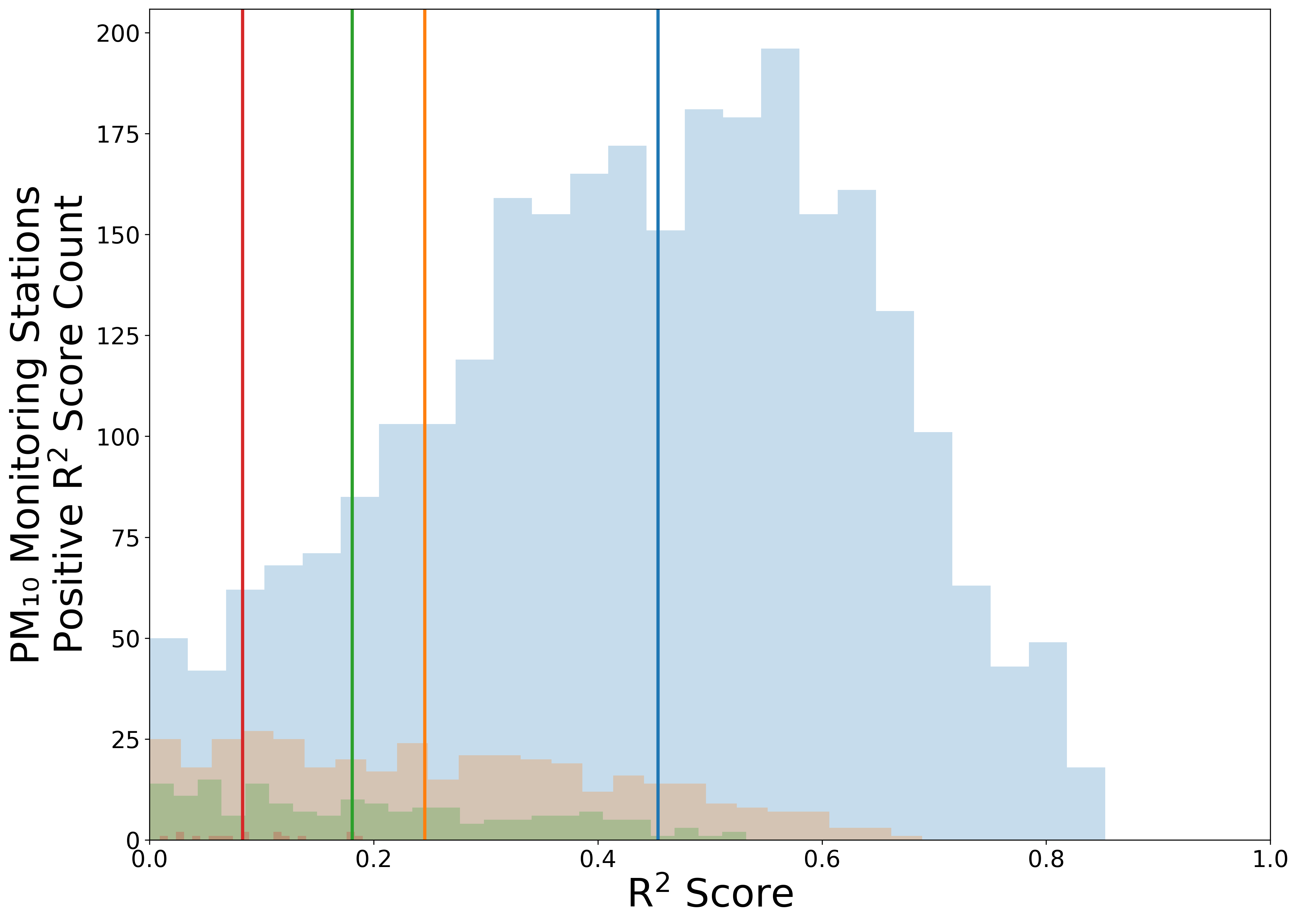}
    \caption{PM$_{10}$} \label{fig:temporalpm10}
  \end{subfigure}%
  \hspace*{\fill}   
   \hspace*{\fill}   
  \begin{subfigure}{0.49\textwidth}
    \includegraphics[width=\linewidth]{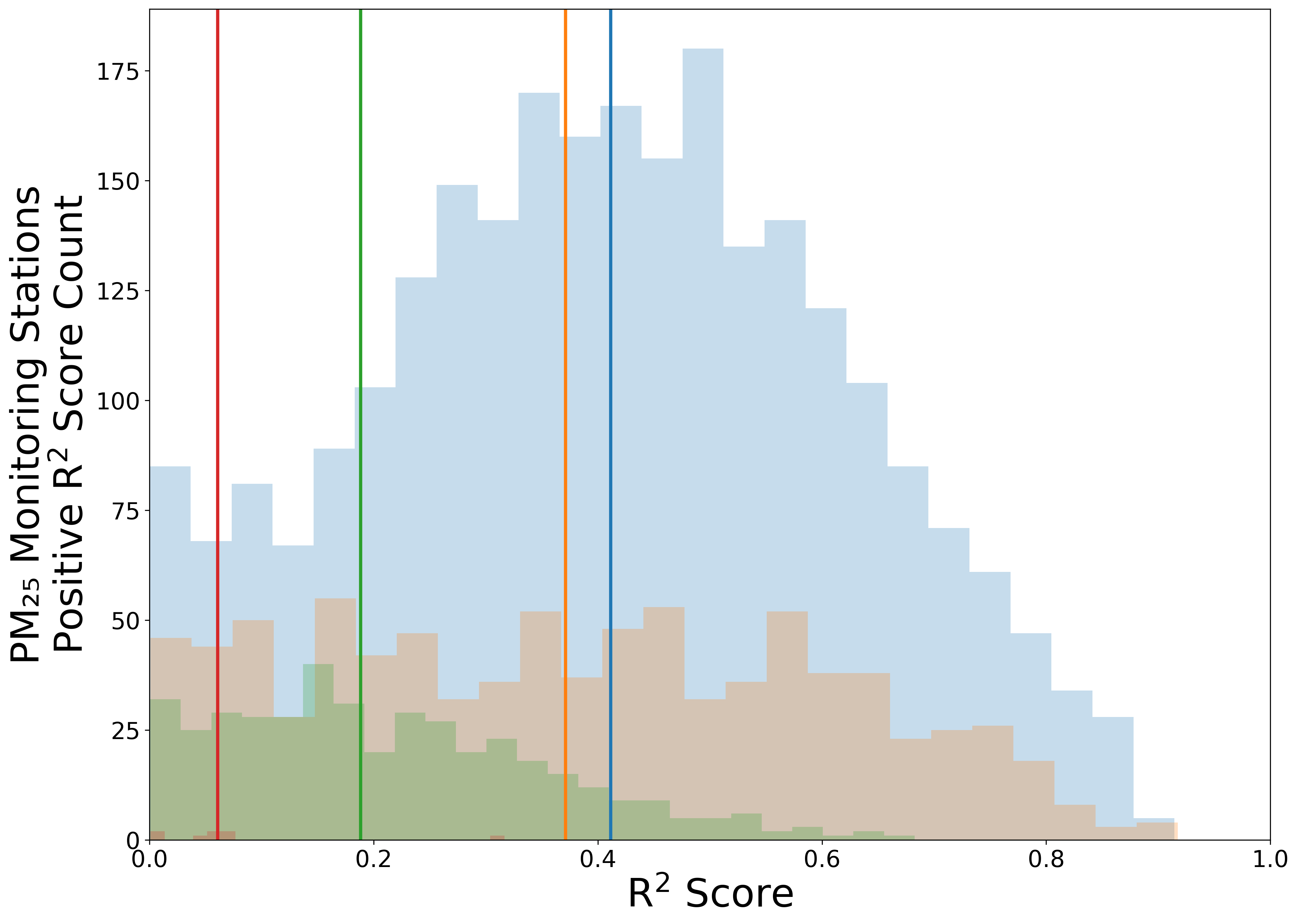}
    \caption{PM$_{2.5}$} \label{fig:temporalpm25}
  \end{subfigure}%
  \hspace*{\fill}   
  \\
  \begin{subfigure}{0.49\textwidth}
    \includegraphics[width=\linewidth]{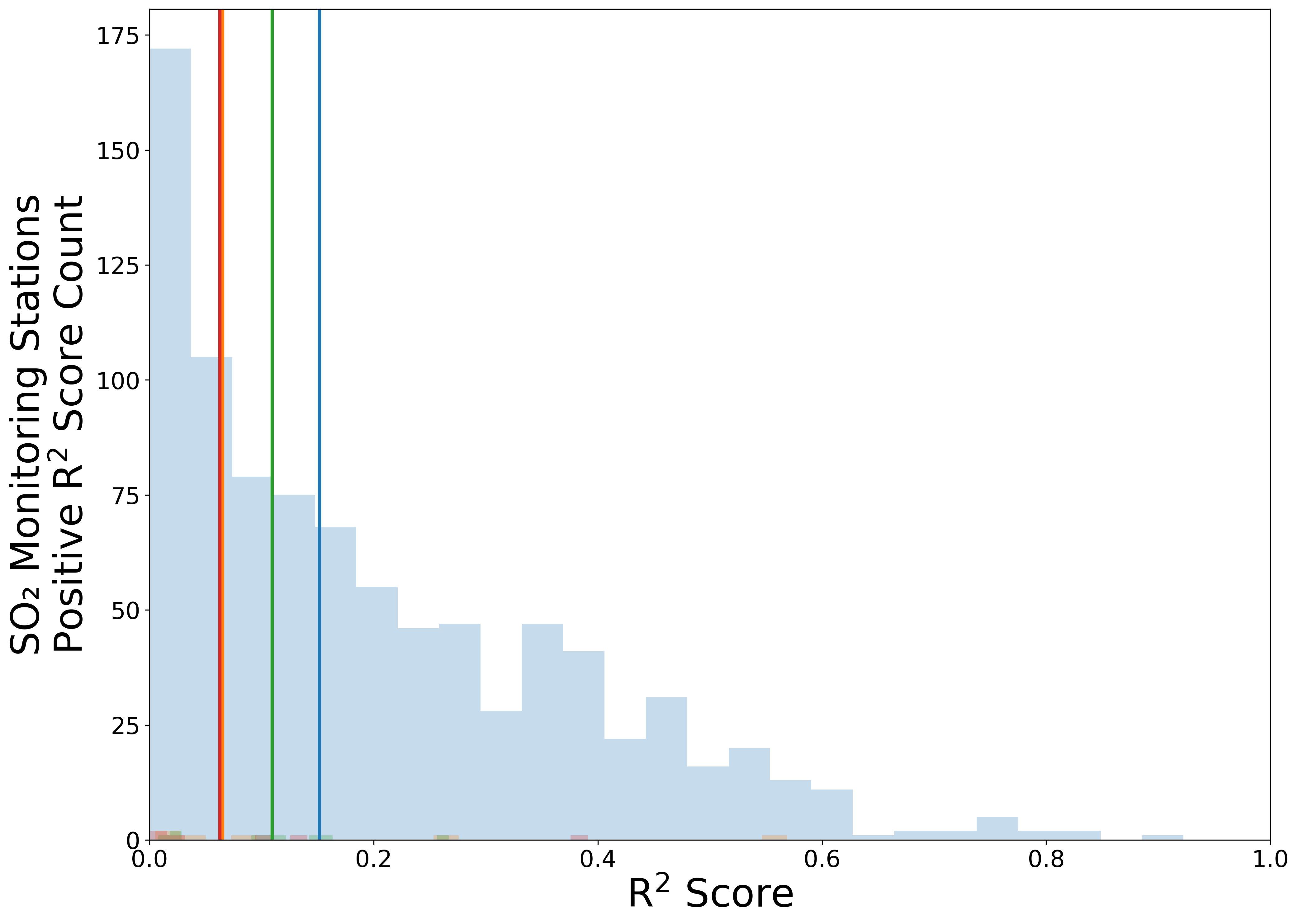}
    \caption{SO$_2$} \label{fig:temporalSO2}
  \end{subfigure}%
  \hspace*{\fill}   
   \raisebox{12mm}{
  \begin{subfigure}{0.4\textwidth}
    \includegraphics[width=\linewidth]{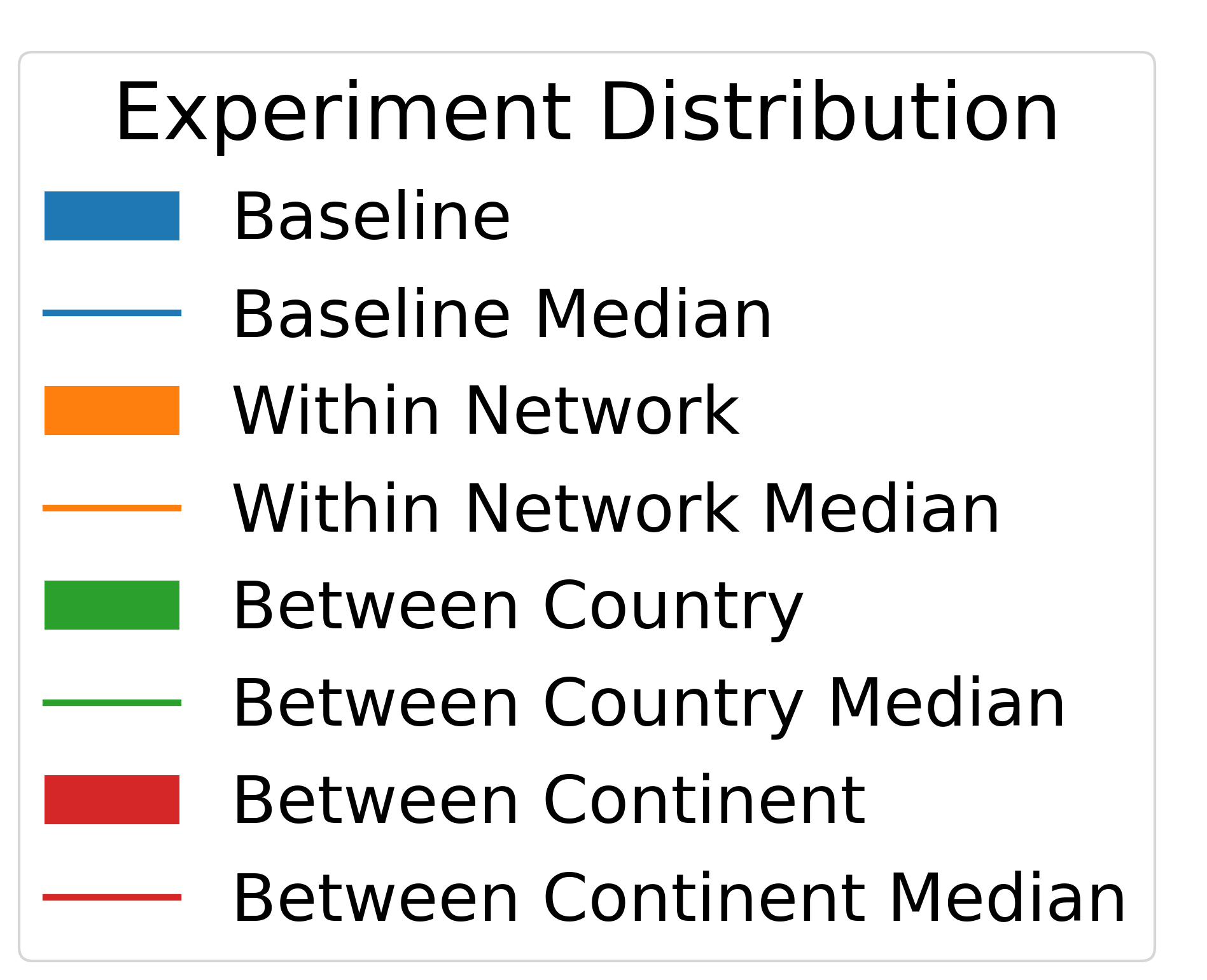}
  \end{subfigure}%
  }
  \hspace*{\fill}   
  \\
\caption{{\bfseries Positive R$^2$ Score Distribution for Each Experiment for Each Air Pollutant.} The distribution broadly follows the same trend of the reduction in magnitude and shift to the left with each experiment, except SO$_2$, likely driven by the minimal number of monitoring stations available. The distributions suggest that the model framework is robust to different air pollutants. Performance for each primarily depends on the quality of the feature vector describing the phenomena driving a given air pollutant, with O$_3$ performing the best while also having its driving phenomena, meteorological conditions, having the highest quality feature vector data, with unique data points spatially and temporally. } \label{fig:positiveR2Scores}
\end{figure}

Figure \ref{fig:positiveR2Scores} shows the overall trends of the different R$^2$ scores for all monitoring stations across each air pollutant. The trend for the overall distribution is similar to the one seen for monitoring station EEA Spain 4327 for NO$_2$. There appears to be a noticeable drop from the baseline to the within-network missing station experiment, with a considerable drop in performance during the between-country experiment, with poor performance maintained in the between-continent missing stations experiment. These initial results suggest that the current iteration of the approach works mainly for estimating air pollution in locations with similar air pollution concentrations, highlighted by the stark drop in performance when a given country's air pollution measurements are excluded from training. While there is an apparent use of the model in estimating air pollution concentration in countries with a monitoring station, understanding why the model works in this situation is paramount and is what the next section tackles.

\section{Empowering Stakeholders: Prediction Intervals}
\label{predictionIntervals}

As highlighted in Section \ref{sec:models}, the modelling approach does have merit, performing well for some monitoring stations, providing an initial signal that a robust and accurate global air pollution model driven by supervised machine learning can be achieved. However, there are locations in which the model could perform better, particularly in the context of large geographic areas where there are no monitoring stations. In some of these locations, the model has seen similar data before and performs well; in other cases, it may not. To empower stakeholders, we have also created auxiliary models to create a prediction interval for all the predictions the model has made. The prediction interval measures the model's uncertainty when predicting a location, allowing stakeholders to use or discard data points depending on their risk appetite. 

\begin{figure}
    \includegraphics[width=\linewidth]{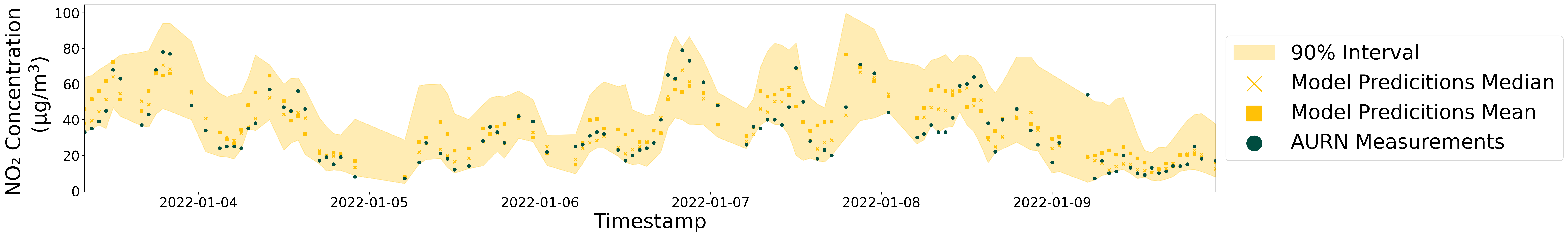}
\caption{{\bfseries EEA Spain 4327 Measurements and Model Predictions With Associated Prediction Interval.} As seen in Section \ref{sec:models}, there are some situations in which the model does not work robustly. For all predictions, a 90\% prediction interval is provided to ensure any downstream users are aware of model uncertainty for a given prediction. The interval empowers stakeholders to identify and proceed in their desired manner, depending on whether the interval is seen as too large for a given use case. } \label{fig:monitoringStationEEASpain4327PredictionInterval}
\end{figure}

Prediction intervals aim to quantify and communicate to downstream users the uncertainty associated with each prediction, helping to ensure that decision-makers understand the broader context of the prediction outside of the point estimate. The prediction interval range helps to highlight situations where the model is making predictions outside of scenarios it has seen before. The prediction interval gives an interval within which a future observation will fall with a certain probability, accounting for the variability of the regression function and the individual observation. Quantile regression is used to create the prediction interval, with three new models created for the 0.05, 0.5 and 0.95 quantiles, with the range between the 0.05 and 0.95 quantiles providing the 90\% prediction interval. Figure \ref{fig:monitoringStationEEASpain4327PredictionInterval} shows the prediction interval for the EEA Spain 4327 monitoring station for the first week of 2022, helping to provide user confidence in the estimations at different data points. 

\section{Understanding Model Performance}

\begin{figure}[!htb]
    \hspace*{\fill}   
  \begin{subfigure}{0.22\textwidth}
    \includegraphics[width=\linewidth]{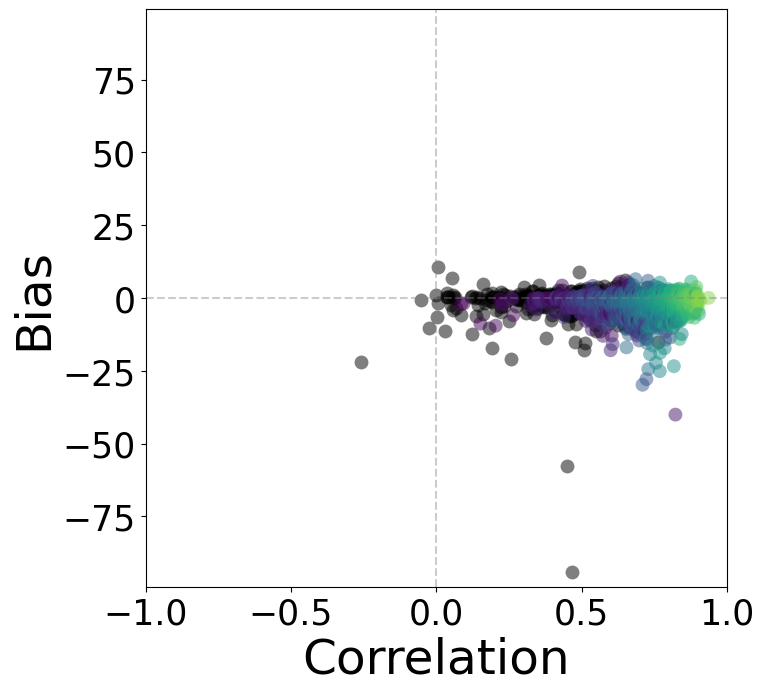}
    \caption{Baseline} 
  \end{subfigure}%
  \hspace*{\fill}   
  \begin{subfigure}{0.22\textwidth}
    \includegraphics[width=\linewidth]{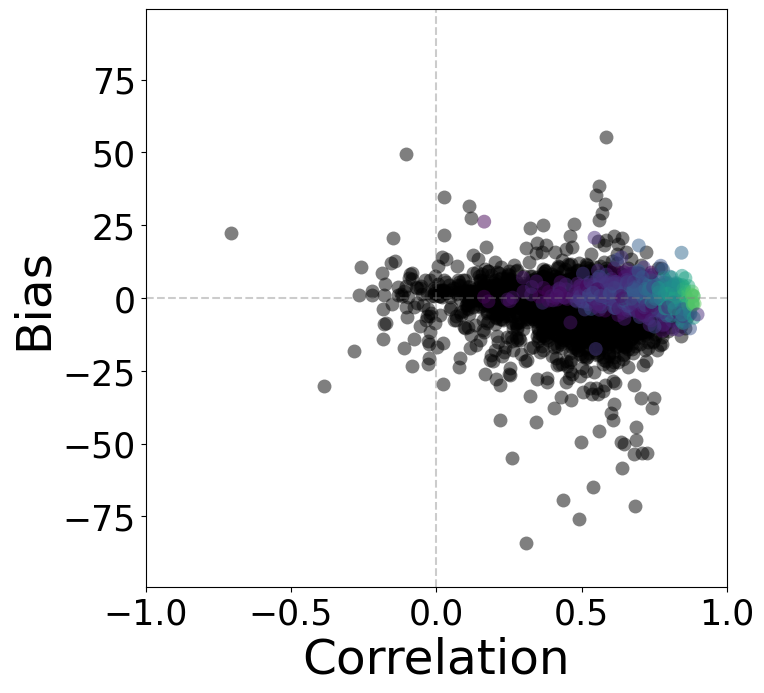}
    \caption{Within Network}
  \end{subfigure}%
  \hspace*{\fill}   
  \begin{subfigure}{0.22\textwidth}
    \includegraphics[width=\linewidth]{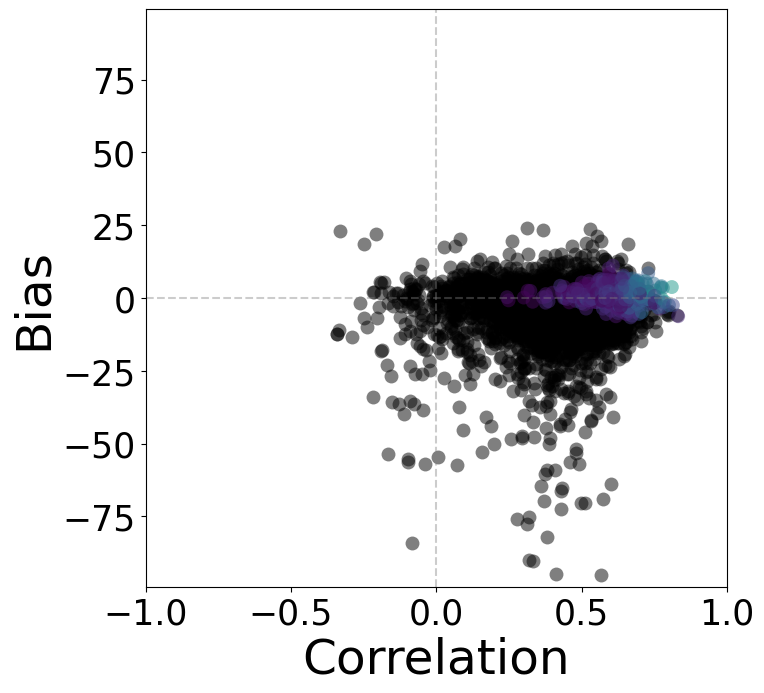}
    \caption{Between Country}
  \end{subfigure}%
  \hspace*{\fill}   
  \begin{subfigure}{0.22\textwidth}
    \includegraphics[width=\linewidth]{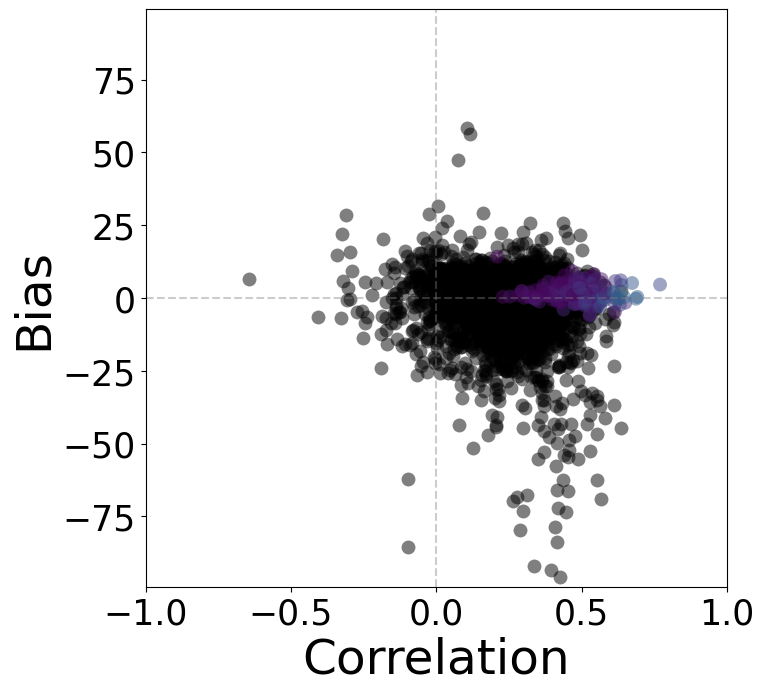}
    \caption{Between Continent}
  \end{subfigure}%
  \hspace*{\fill}   
  \raisebox{10mm}{
  \begin{subfigure}{0.08\textwidth}
    \includegraphics[width=\linewidth]{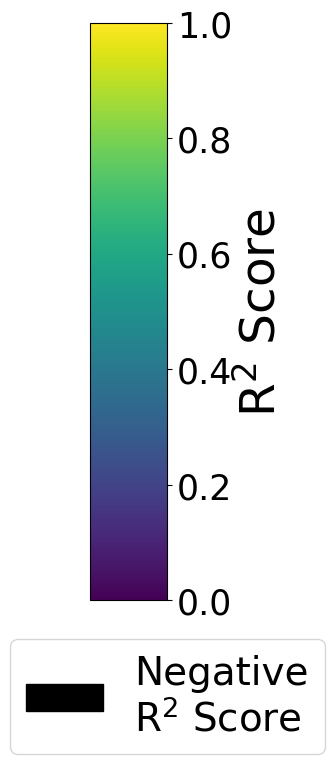}
  \end{subfigure}%
  }
  \hspace*{\fill}   
\caption{{\bfseries Bias vs Correlation for NO$_2$ for Each of the Experiments.} It can be seen that the correlation has a weaker relationship with a positive R$^2$ score than the bias does, with data points for nearly all values of correlation between 0 and 1 in Figure \ref{fig:temporalNO2} having a positive R$^2$ score, with data points deviating from a 0 bias having a negative R$^2$ score. This relationship highlights that a high-magnitude bias is driving the poor model performance for some monitoring stations. } \label{fig:correlationBiasNO2}
\end{figure}

While the R$^2$ provides a good indication of the model's performance for a given monitoring station's prediction, the bias and correlation of the prediction can provide further insight into the model's performance. Bias represents the average difference between the monitoring station measurements and the model predictions, providing a metric for the overall tendency of the modelled values to be higher or lower than the observations. Correlation quantifies the linear relationship between the monitoring station measurements and the model predictions, reflecting how well the model captures the temporal variations, with the Pearson Correlation Coefficient being used in this study. For EEA Spain 4327, the correlation and bias for the baseline are 0.9 and 0.64, respectively, with 0.89 and 0.75 for the within network experiment, and finally, 0.44 correlation and -8.8 bias for the between country missing stations experiment where the R$^2$ dropped below 0. Reanalysing the EEA Spain 4327 monitoring station with these statistics highlights that the low R$^2$ is driven primarily by bias rather than the correlation, indicating the model can capture the overall trend of the air pollution concentrations but struggles with the assignment of particular values unless air pollution measurements in a similar location to the one being estimated has been seen. The overall trends across all of the monitoring stations can be seen similarly, with Figure \ref{fig:correlationBiasNO2} showing the correlation against the bias for each monitoring station, with the points coloured by their R$^2$ score for NO$_2$. Across the monitoring stations, as the experiments progress from the baseline, it can be seen that the points at the 0 bias are the points with a positive R$^2$ score across a range of different correlations. In the within network experiment, some monitoring station predictions exhibit a large magnitude bias with a negative R$^2$ score, which is further highlighted in the between country and continent experiment. Figure \ref{S-fig:allCorrelationBias} shows the bias-correlation plots for the other air pollutants.

Table \ref{tab:biasCorrelationIQR} shows the 90\% interquartile range for the bias and correlation for NO$_2$, where it can be seen more clearly that the correlation remains constant when compared with the large changes in the bias, further supporting the idea that the increase in bias magnitude is what causes the reduction in R$^2$. Further supported when analysing the case of monitoring station EEA Spain 4327, shown in Figures \ref{fig:temporalIndividualPredicitionsEEASpain4327} \ref{fig:monitoringStationEEASpain4327SpatialKFold} \ref{fig:monitoringStationEEASpain4327SpatialCountry} \ref{fig:monitoringStationEEASpain4327SpatialContinent}, the process that appears to be followed is that the inclusion of data in a location that is to be estimated appears to ``ground'' the time series prediction, with the overall trend and correlation of the time series remaining broadly constant but the bias changing, driving changes in R$^2$ score.

\begin{table}[!htb]
\resizebox{\linewidth}{!}{
\begin{filecontents*}{CSVs/bias_correlation_iqr.csv}
,Air Pollutant,Correlation Baseline,Correlation Between Continent,Correlation Between Country,Correlation Within Network,Bias Baseline,Bias Between Continent,Bias Between Country,Bias Within Network
0,NO₂,0.33,0.38,0.49,0.52,6.1,23.55,20.82,17.0
1,O₃,0.31,0.49,0.53,0.48,6.66,35.32,27.8,23.48
2,PM₁₀,0.4,0.35,0.62,0.52,8.54,41.9,35.41,18.62
3,PM₂₅,0.45,0.4,0.57,0.58,4.66,19.57,13.09,7.61
4,SO₂,0.6,0.36,0.39,0.45,4.06,21.94,15.62,9.22
\end{filecontents*}
\pgfplotstabletypeset[
    multicolumn names=l, 
    col sep=comma, 
    string type, 
    header = has colnames,
    columns={Air Pollutant, Correlation Baseline, Correlation Within Network, Correlation Between Country, Correlation Between Continent, Bias Baseline, Bias Within Network, Bias Between Country, Bias Between Continent},
    columns/Air Pollutant/.style={column type=l, column name=\shortstack{Air\\Pollutant}},
    columns/Experiment/.style={column type=l, column name=Experiment Name},
    columns/Correlation Baseline/.style={column type={S[round-precision=2, table-format=2.2, table-number-alignment=center]}, column name=\shortstack{Correlation\\Baseline}},
    columns/Correlation Within Network/.style={column type={S[round-precision=2, table-format=2.2, table-number-alignment=center]}, column name=\shortstack{Correlation\\Within Network}},
    columns/Correlation Between Country/.style={column type={S[round-precision=2, table-format=2.2, table-number-alignment=center]}, column name=\shortstack{Correlation\\Between Country}},
    columns/Correlation Between Continent/.style={column type={S[round-precision=2, table-format=2.2, table-number-alignment=center]}, column name=\shortstack{Correlation\\Between Continent}},
    columns/Bias Baseline/.style={column type={S[round-precision=2, table-format=2.2, table-number-alignment=center]}, column name=\shortstack{Bias\\Baseline}},
    columns/Bias Within Network/.style={column type={S[round-precision=2, table-format=2.2, table-number-alignment=center]}, column name=\shortstack{Bias\\Within Network}},
    columns/Bias Between Country/.style={column type={S[round-precision=2, table-format=2.2, table-number-alignment=center]}, column name=\shortstack{Bias\\Between Country}},
    columns/Bias Between Continent/.style={column type={S[round-precision=2, table-format=2.2, table-number-alignment=center]}, column name=\shortstack{Bias\\Between Continent}},
    every head row/.style={before row=\toprule, after row=\midrule},
    every last row/.style={after row=\bottomrule}
    ]{CSVs/bias_correlation_iqr.csv}}
    \smallskip
\caption{{ \bfseries 90\% Interquartile Range (IQR) for Bias and Correlation for Each Experiment for Each Air Pollutant. } The IQR for the bias and correlation for each air pollutant follows the same trend shown in Figure \ref{fig:correlationBiasNO2} for NO$_2$. The IQR for each of the air pollutants for each of the experiments broadly remains the same, apart from a slight reduction in the last experiment, between continent when all of the correlation values shifts downwards, causing a reduction in the IQR. In contrast, the bias doubles in magnitude between the baseline and within experiments and further increases in subsequent experiments. This considerable increase is responsible for the reduction in model performance seen in Section \ref{sec:models}.} \label{tab:biasCorrelationIQR}
\end{table}

\section{Improving Model Performance}

\begin{figure}
    \includegraphics[width=\linewidth]{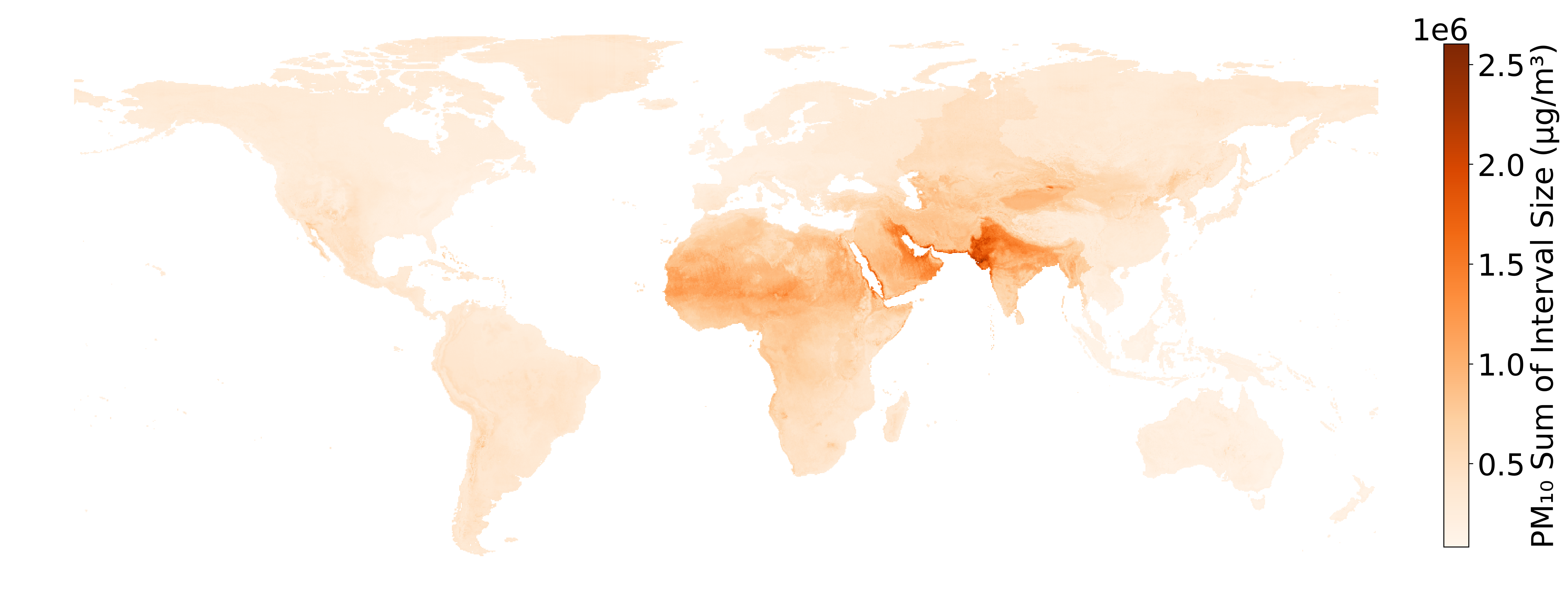}
\caption{{\bfseries Prediction Interval Size Sum for PM$_{10}$ for 2022.} The size of the prediction interval sum across 2022 highlights the locations the model is most uncertain about, providing locations spatially for future monitoring station placement. In the case of PM$_{10}$, it can be seen that Africa, the Middle East and some parts of Western Asia are where the model has the highest uncertainty. Other air pollutants are shown in Figure \ref{S-fig:allAirPollutionUncertainityMaps}.} \label{fig:predictionIntervalSize}
\end{figure}

As highlighted before, acquiring more air pollution concentration data is necessary to reduce the bias experienced when estimating locations dissimilar to previously seen locations. The size of the prediction interval discussed in Section \ref{predictionIntervals} denoted the predictions with the greatest uncertainty, highlighting locations of particular interest to the model to reduce uncertainty and the placement of future monitoring stations. Figure \ref{fig:predictionIntervalSize} shows the locations across 2022 that had the largest interval size, providing the locations that would likely benefit from additional data, either in regards to a feature vector dataset describing the phenomena driving air pollution in those locations or simply measurements of air pollution.  

\section{Research Data Output}

\begin{figure}[!htb]
  \begin{subfigure}{0.99\textwidth}
    \includegraphics[width=\linewidth]{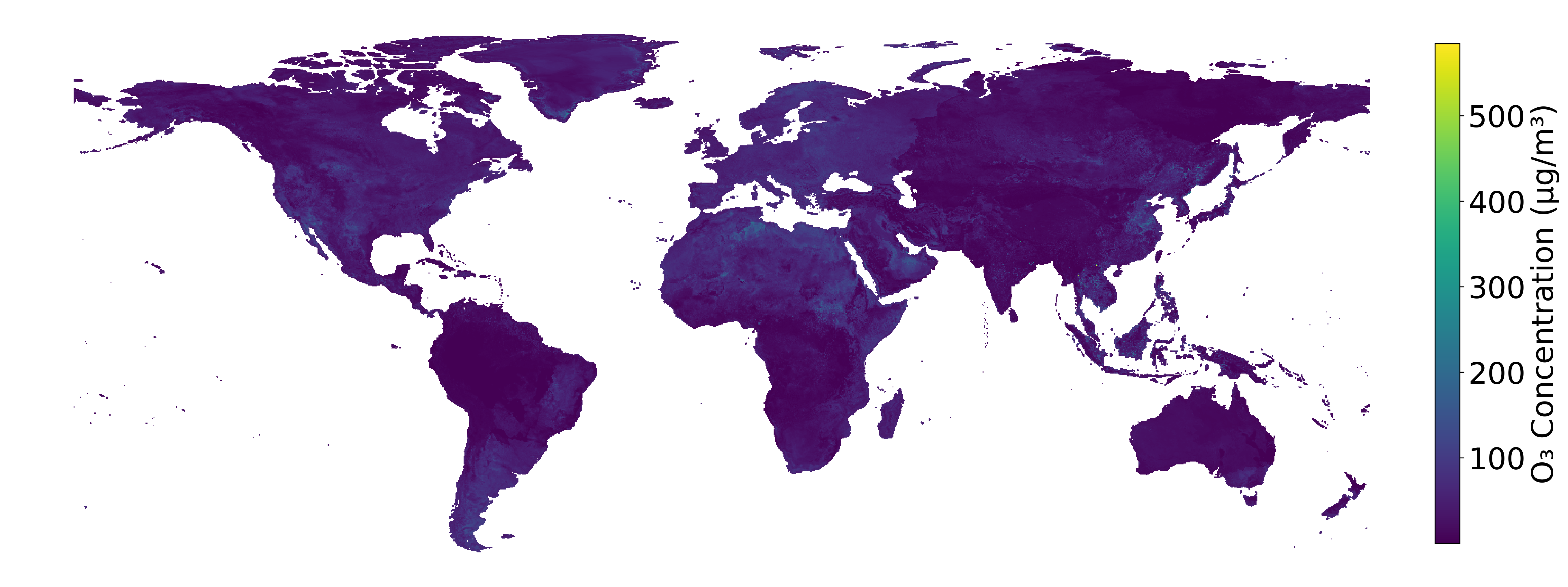}
    \caption{O$_3$ \si{\micro\gram/\meter^3} Concentrations.} \label{fig:airPollutionMapO3Concentrations}
  \end{subfigure}%
  \hspace*{\fill}   
  \\
  \hspace*{\fill}   
  \begin{subfigure}{0.99\textwidth}
    \includegraphics[width=\linewidth]{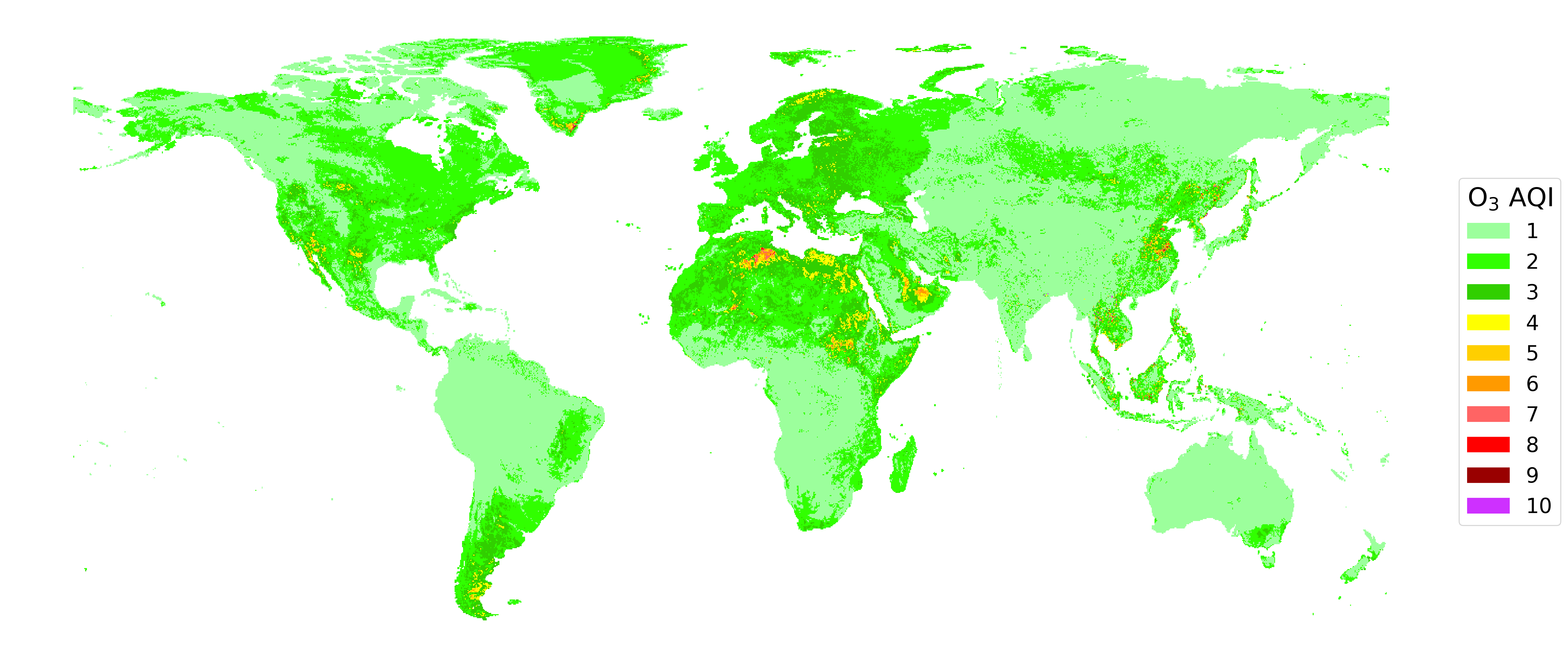}
    \caption{Derived Air Quality Index.} \label{fig:airPollutionMapO3AQI}
  \end{subfigure}%

\caption{{\bfseries Air Pollution Maps for O$_3$ at 8AM 2nd July 2022 Both in Concentrations and Air Quality Index.} The figures show the spatial resolution of the model outputs for a single hour, highlighting the locations with high concentrations of O$_3$.} \label{fig:airPollutionMapSingle}
\end{figure}

\begin{figure}
    \includegraphics[width=\linewidth]{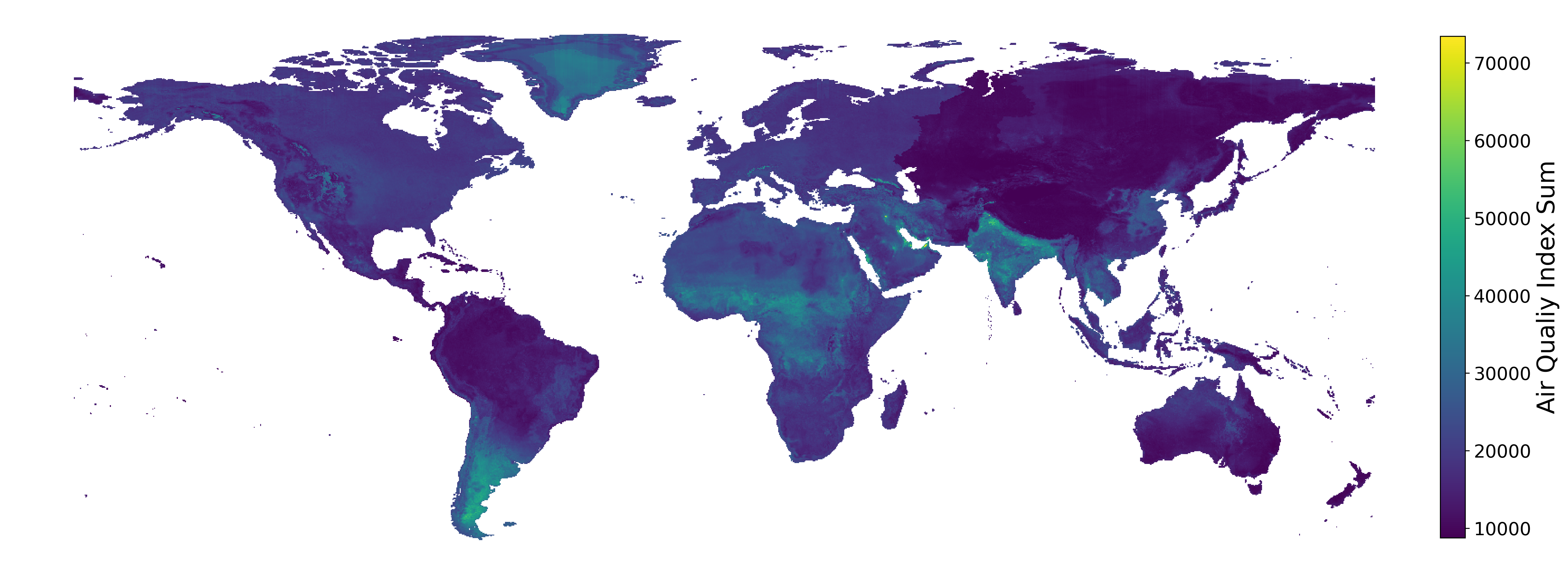}
\caption{{\bfseries Annual Air Pollution Map for Air Quality Index Summation. }Each of the individual air pollution maps for each hour shown in Figure \ref{fig:airPollutionMapSingle} can be aggregated to the temporal resolution is desired, such as the annual level, to support more strategic decision-making, rather than looking at a potential single outlier peak. } \label{fig:airPollutionMap2018AQI}
\end{figure}

As part of this work, we are providing hourly 0.25$^{\circ}$ global air pollution concentration data for NO$_2$, O$_3$, PM$_{10}$, PM$_{2.5}$, SO$_2$, supporting various downstream assessments which can now be conducted using the high temporal and spatial resolution the data-driven supervised machine learning model provides. The raw concentrations of the air pollutants, shown in Figure \ref{fig:airPollutionMapSingle}, highlight the differences in concentrations at the hourly level, helpful for scientific activities. However, the concentrations can also be used for public dissemination, such as the Air Quality Index calculated from the concentrations, shown in Figure \ref{fig:airPollutionMap2018AQI}. 

\begin{figure}
    \includegraphics[width=\linewidth]{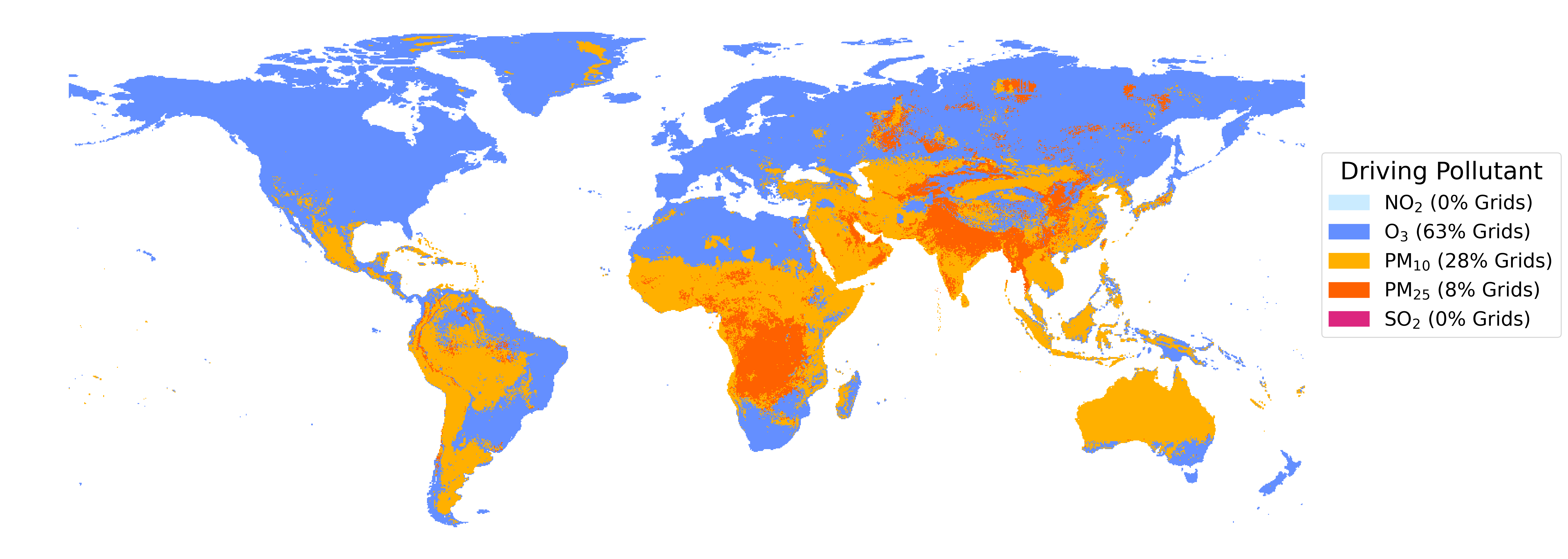}
\caption{{\bfseries Annual Air Pollution Map for Driving Air Quality Subindex.} Further to the concentrations map shown for the annual level shown in Figure \ref{fig:airPollutionMap2018AQI}, further analysis can be conducted to understand which air pollutant is driving the poor air quality in different regions, helping to design targetted interventions to the most pressing problems. } \label{fig:airPollutionMap2018DrivingSubindex}
\end{figure}

Alongside individual hours concentrations being analysed, more long-term analysis can be conducted, such as understanding across the year 2022 the air pollutant that is driving the poor air quality in a given location, shown in Figure \ref{fig:airPollutionMap2018DrivingSubindex}. Understanding the air pollutants driving poor air quality helps to provide a basis for designing interventions, with the interventions for NO$_2$ being substantially different from the interventions to reduce SO$_2$ concentrations. 

\section{Discussion}

The findings presented in Section \ref{sec:models} offer a comprehensive assessment of the model's efficacy under diverse conditions, elucidating its capacity to discern the intricate relationship between features and the target vector, while also delineating its constraints in predicting air pollution levels in unmonitored locations. The experimental outcomes reveal that the model exhibits commendable accuracy within known monitoring station networks, where it has previously encountered data. However, its predictive accuracy diminishes for countries without prior data exposure, and this decline is more pronounced for entirely new continents. This gradient in performance underscores the model's dependency on prior knowledge for optimal prediction accuracy.

While the \(R^2\) score offers a general measure of the model's performance, an examination of bias and correlation within the results elucidates that the bias is a significant factor in suboptimal model outcomes. The observed strong correlation suggests the model's proficiency in capturing the general trend of the air pollution concentration time series, yet it encounters difficulties in accurately determining the precise magnitudes. Notably, exposure to training data from the specific location in question markedly mitigates the model's bias, thereby enhancing the \(R^2\) score. This improvement underscores the indispensability of real-world air pollution measurements to refine model accuracy. Furthermore, leveraging model uncertainty to identify locations of greatest prediction uncertainty illuminates where additional data gathering would yield the most value.

This study introduces a first step towards the development of a lightweight global air pollution model, leveraging a data-driven, supervised machine learning framework. It represents the first endeavour to predict air pollution with such a high degree of spatial and temporal precision, demonstrating the scalable potential of machine learning in addressing complex environmental challenges.

Future enhancements to this research can be broadly categorized into two avenues: data refinement and model enhancement. An immediate next step involves a preprocessing stage for the input data. As illustrated in Figure \ref{fig:temporalIndividualPredicitionsCAAQM8171}, the quality of data from some monitoring stations appears dubious. Since the model currently does not distinguish between high- and low-quality data during training, any inaccuracies inherently affect model performance. By identifying and addressing anomalies and outliers, either through exclusion or special treatment, model accuracy is expected to improve. Enhancing the feature vector by incorporating additional variables, such as those related to the transportation sector, could further refine the model's performance. 

Moreover, augmenting the model to integrate regional datasets—available in certain areas like North America but not in others like Asia—would ensure the maximal utility of available data. The present model's requirement for a uniform set of feature vectors across all data points restricts the use of diverse datasets, often resulting in reliance on data with limited spatial and temporal resolution, such as the monthly emissions data employed in this study.

In conclusion, this work plays a pivotal role for a wide array of stakeholders. By providing prediction intervals alongside the predictions themselves, we equip stakeholders with the means to make informed decisions based on their specific risk tolerance, thereby facilitating a spectrum of new applications.

\textbf{Declarations}

Availability of data and materials: The air pollution concentration dataset is accessed via OpenAQ\footnote{Clickable Link: \href{https://openaq.org/}{OpenAQ Homepage}}. Sentinel 5P data is accessed via Google Earth Engine Sentinel Catalogue \footnote{Clickable Link: 
 \href{https://developers.google.com/earth-engine/datasets/catalog/sentinel-5p}{Google Earth Engine Sentinel 5p Catalogue}}. Meteorology data is accessed via ECMWF ERA5 \footnote{Clickable Link: \href{https://www.ecmwf.int/en/forecasts/dataset/ecmwf-reanalysis-v5}{ECMWF ERA5 Data Repository}}. Emissions data is accessed via the ECCAD Global Emissions Dataset \footnote{Clickable Link: \href{https://eccad.aeris-data.fr/context/}{ECCAD Homepage}}. \\
Competing interests: The authors declare that they have no competing interests. \\
Funding: Liam Berrisford is supported by a UKRI Studentship at the Center for Doctoral Training for Environmental Intelligence at the University of Exeter.\\
Author Contributions: LJB and RM jointly conceived the study's original idea. LJB performed data curation, formal analysis, investigation, methodology design, validation, visualisation, writing the original draft. RM provided oversight for the project and critically reviewed the experimental outcomes. HB contributed by providing feedback on the manuscript and suggesting changes to enhance its clarity and technical accuracy. \\
Acknowledgements: Not applicable. \\

\bibliographystyle{IEEEtran}
\bibliography{bibliography}

\begin{thebibliography}{100}
\providecommand{\url}[1]{#1}
\csname url@samestyle\endcsname
\providecommand{\newblock}{\relax}
\providecommand{\bibinfo}[2]{#2}
\providecommand{\BIBentrySTDinterwordspacing}{\spaceskip=0pt\relax}
\providecommand{\BIBentryALTinterwordstretchfactor}{4}
\providecommand{\BIBentryALTinterwordspacing}{\spaceskip=\fontdimen2\font plus
\BIBentryALTinterwordstretchfactor\fontdimen3\font minus
  \fontdimen4\font\relax}
\providecommand{\BIBforeignlanguage}[2]{{%
\expandafter\ifx\csname l@#1\endcsname\relax
\typeout{** WARNING: IEEEtran.bst: No hyphenation pattern has been}%
\typeout{** loaded for the language `#1'. Using the pattern for}%
\typeout{** the default language instead.}%
\else
\language=\csname l@#1\endcsname
\fi
#2}}
\providecommand{\BIBdecl}{\relax}
\BIBdecl

\bibitem{WHO:2021:GlobalAirPollutionStatistics}
W.~H. Organization \emph{et~al.}, \emph{WHO global air quality guidelines:
  particulate matter (PM2. 5 and PM10), ozone, nitrogen dioxide, sulfur dioxide
  and carbon monoxide}.\hskip 1em plus 0.5em minus 0.4em\relax World Health
  Organization, 2021.

\bibitem{AEATechnology:2006:PurchasingAURNCost}
{AEA Technology}, Aug 2006,
  \url{https://uk-air.defra.gov.uk/assets/documents/reports/cat06/0608141644-386_Purchasing_Guide_for_AQ_Monitoring_Equipment_Version2.pdf
  } "Accessed on: 08/2022 [Online]".

\bibitem{hoffmann:2021:WHO2021AQGs}
B.~Hoffmann, H.~Boogaard, A.~de~Nazelle, Z.~J. Andersen, M.~Abramson,
  M.~Brauer, B.~Brunekreef, F.~Forastiere, W.~Huang, H.~Kan \emph{et~al.},
  ``Who air quality guidelines 2021--aiming for healthier air for all: a joint
  statement by medical, public health, scientific societies and patient
  representative organisations,'' \emph{International journal of public
  health}, vol.~66, p. 1604465, 2021.

\bibitem{UKAIR:2023:DAQI}
\BIBentryALTinterwordspacing
{UK-AIR, DEFRA}. (2023) {Daily Air Quality Index}. Accessed on: 29/11/2023.
  [Online]. Available: \url{https://uk-air.defra.gov.uk/air-pollution/daqi}
\BIBentrySTDinterwordspacing

\bibitem{boldo:2006:PMHealthImpact}
E.~Boldo, S.~Medina, A.~Le~Tertre, F.~Hurley, H.-G. M{\"u}cke, F.~Ballester,
  I.~Aguilera, and D.~E. on~behalf of~the Apheis~group, ``Apheis: Health impact
  assessment of long-term exposure to pm 2.5 in 23 european cities,''
  \emph{European journal of epidemiology}, vol.~21, pp. 449--458, 2006.

\bibitem{ashmore:2005:OzoneVegetationImpact}
M.~Ashmore, ``Assessing the future global impacts of ozone on vegetation,''
  \emph{Plant, Cell \& Environment}, vol.~28, no.~8, pp. 949--964, 2005.

\bibitem{hajat:2015:GlobalAirPollutionResearch}
A.~Hajat, C.~Hsia, and M.~S. O’Neill, ``Socioeconomic disparities and air
  pollution exposure: a global review,'' \emph{Current environmental health
  reports}, vol.~2, pp. 440--450, 2015.

\bibitem{vieno:2016:UKAirQualitySaharanDust}
M.~Vieno, M.~R. Heal, M.~M. Twigg, I.~A. MacKenzie, C.~F. Braban, J.~Lingard,
  S.~Ritchie, R.~Beck, A.~M{\'o}ring, R.~Ots \emph{et~al.}, ``The uk
  particulate matter air pollution episode of march--april 2014: more than
  saharan dust,'' \emph{Environmental Research Letters}, vol.~11, no.~4, p.
  044004, 2016.

\bibitem{bell:2007:OzoneClimateChange}
M.~L. Bell, R.~Goldberg, C.~Hogrefe, P.~L. Kinney, K.~Knowlton, B.~Lynn,
  J.~Rosenthal, C.~Rosenzweig, and J.~A. Patz, ``Climate change, ambient ozone,
  and health in 50 us cities,'' \emph{Climatic Change}, vol.~82, pp. 61--76,
  2007.

\bibitem{goss:2020:WildfireClimateChange}
M.~Goss, D.~L. Swain, J.~T. Abatzoglou, A.~Sarhadi, C.~A. Kolden, A.~P.
  Williams, and N.~S. Diffenbaugh, ``Climate change is increasing the
  likelihood of extreme autumn wildfire conditions across california,''
  \emph{Environmental Research Letters}, vol.~15, no.~9, p. 094016, 2020.

\bibitem{knorr:2017:WildfireAirPollution}
W.~Knorr, F.~Dentener, J.-F. Lamarque, L.~Jiang, and A.~Arneth, ``Wildfire air
  pollution hazard during the 21st century,'' \emph{Atmospheric Chemistry and
  Physics}, vol.~17, no.~14, pp. 9223--9236, 2017.

\bibitem{horton:2014:StangnationEventsClimateChange}
D.~E. Horton, C.~B. Skinner, D.~Singh, and N.~S. Diffenbaugh, ``Occurrence and
  persistence of future atmospheric stagnation events,'' \emph{Nature climate
  change}, vol.~4, no.~8, pp. 698--703, 2014.

\bibitem{trenberth:2011:PrecipitationChangesClimateChange}
K.~E. Trenberth, ``Changes in precipitation with climate change,''
  \emph{Climate research}, vol.~47, no. 1-2, pp. 123--138, 2011.

\bibitem{kang:2022:LowCostAirPollutionSensor}
Y.~Kang, L.~Aye, T.~D. Ngo, and J.~Zhou, ``Performance evaluation of low-cost
  air quality sensors: A review,'' \emph{Science of The Total Environment},
  vol. 818, p. 151769, 2022.

\bibitem{DEFRA:2023:LowCostSensors}
\BIBentryALTinterwordspacing
{UK-AIR, DEFRA}. (2021) {'Low-cost' pollution sensors - understanding the
  uncertainties}. Accessed on: 29/11/2023. [Online]. Available:
  \url{https://uk-air.defra.gov.uk/research/aqeg/pollution-sensors/understanding-uncertainties.php}
\BIBentrySTDinterwordspacing

\bibitem{concas:2021:LowCostSensorCalibration}
F.~Concas, J.~Mineraud, E.~Lagerspetz, S.~Varjonen, X.~Liu, K.~Puolam{\"a}ki,
  P.~Nurmi, and S.~Tarkoma, ``Low-cost outdoor air quality monitoring and
  sensor calibration: A survey and critical analysis,'' \emph{ACM Transactions
  on Sensor Networks (TOSN)}, vol.~17, no.~2, pp. 1--44, 2021.

\bibitem{Castell:2017:LowCostSensorEUComparison}
\BIBentryALTinterwordspacing
N.~Castell, F.~R. Dauge, P.~Schneider, M.~Vogt, U.~Lerner, B.~Fishbain,
  D.~Broday, and A.~Bartonova, ``\BIBforeignlanguage{en}{Can commercial
  low-cost sensor platforms contribute to air quality monitoring and exposure
  estimates?}'' \emph{\BIBforeignlanguage{en}{Environment International}},
  vol.~99, pp. 293--302, Feb. 2017. [Online]. Available:
  \url{https://linkinghub.elsevier.com/retrieve/pii/S0160412016309989}
\BIBentrySTDinterwordspacing

\bibitem{veefkind:2012:Sentinel5PDescription}
J.~P. Veefkind, I.~Aben, K.~McMullan, H.~F{\"o}rster, J.~De~Vries, G.~Otter,
  J.~Claas, H.~Eskes, J.~De~Haan, Q.~Kleipool \emph{et~al.}, ``Tropomi on the
  esa sentinel-5 precursor: A gmes mission for global observations of the
  atmospheric composition for climate, air quality and ozone layer
  applications,'' \emph{Remote sensing of environment}, vol. 120, pp. 70--83,
  2012.

\bibitem{ESA:2023:Sentinel5POrbit}
\BIBentryALTinterwordspacing
{European Space Agency - Copernicus}. (2023) {Sentinel, Missions, Sentinel 5P -
  Orbit}. Accessed on: 29/11/2023. [Online]. Available:
  \url{https://sentinels.copernicus.eu/web/sentinel/missions/sentinel-5p/orbit}
\BIBentrySTDinterwordspacing

\bibitem{ESA:2017:ESAQualityAssurance}
\BIBentryALTinterwordspacing
------. (2017) {Sentinel-5 Precursor Calibration and Validation Plan for the
  Operational Phase}. Accessed on: 29/11/2023. [Online]. Available:
  \url{https://sentinels.copernicus.eu/documents/247904/2474724/Sentinel-5P-Calibration-and-Validation-Plan.pdf}
\BIBentrySTDinterwordspacing

\bibitem{zoogman:2017:NASATEMPORemoteSensing}
P.~Zoogman, X.~Liu, R.~Suleiman, W.~Pennington, D.~Flittner, J.~Al-Saadi,
  B.~Hilton, D.~Nicks, M.~Newchurch, J.~Carr \emph{et~al.}, ``Tropospheric
  emissions: Monitoring of pollution (tempo),'' \emph{Journal of Quantitative
  Spectroscopy and Radiative Transfer}, vol. 186, pp. 17--39, 2017.

\bibitem{eliassen:1984:LagrangianAirModels}
A.~Eliassen, ``Aspects of lagrangian air pollution modelling,'' in \emph{Air
  Pollution Modeling and Its Application III}.\hskip 1em plus 0.5em minus
  0.4em\relax Springer, 1984, pp. 3--21.

\bibitem{Byun:1984:EulerianDispersionModels}
{Daewon W. Byun, Avraham Lacser, Robert Yamartino, and Paolo Zannetti},
  ``Chapter 10 eulerian dispersion models,'' 2003.

\bibitem{zlatev:1992:EulerianModelUse}
Z.~Zlatev, J.~Christensen, and {\O}.~Hov, ``A eulerian air pollution model for
  europe with nonlinear chemistry,'' \emph{Journal of Atmospheric Chemistry},
  vol.~15, pp. 1--37, 1992.

\bibitem{Vitturi:2010:lagrangianVolcano}
M.~de'Michieli Vitturi, A.~Neri, T.~Esposti~Ongaro, S.~Lo~Savio, and E.~Boschi,
  ``Lagrangian modeling of large volcanic particles: Application to vulcanian
  explosions,'' \emph{Journal of Geophysical Research: Solid Earth}, vol. 115,
  no.~B8, 2010.

\bibitem{Henze:2007:GeosChem}
\BIBentryALTinterwordspacing
D.~K. Henze, A.~Hakami, and J.~H. Seinfeld, ``Development of the adjoint of
  geos-chem,'' \emph{Atmospheric Chemistry and Physics}, vol.~7, no.~9, pp.
  2413--2433, 2007. [Online]. Available:
  \url{https://acp.copernicus.org/articles/7/2413/2007/}
\BIBentrySTDinterwordspacing

\bibitem{hoek:2008:review}
G.~Hoek, R.~Beelen, K.~De~Hoogh, D.~Vienneau, J.~Gulliver, P.~Fischer, and
  D.~Briggs, ``A review of land-use regression models to assess spatial
  variation of outdoor air pollution,'' \emph{Atmospheric environment},
  vol.~42, no.~33, pp. 7561--7578, 2008.

\bibitem{freeman:2018:ForecastingAirPollutionConcentration}
B.~S. Freeman, G.~Taylor, B.~Gharabaghi, and J.~Th{\'e}, ``Forecasting air
  quality time series using deep learning,'' \emph{Journal of the Air \& Waste
  Management Association}, vol.~68, no.~8, pp. 866--886, 2018.

\bibitem{tao:2019:ForecastingAirPollutionConcentration2}
Q.~Tao, F.~Liu, Y.~Li, and D.~Sidorov, ``Air pollution forecasting using a deep
  learning model based on 1d convnets and bidirectional gru,'' \emph{IEEE
  access}, vol.~7, pp. 76\,690--76\,698, 2019.

\bibitem{harishkumar:2020:ForecastingAirPollutionConcentration3}
K.~Harishkumar, K.~Yogesh, I.~Gad \emph{et~al.}, ``Forecasting air pollution
  particulate matter (pm2. 5) using machine learning regression models,''
  \emph{Procedia Computer Science}, vol. 171, pp. 2057--2066, 2020.

\bibitem{van:2019:MissingLocationAirPollutionEstimationSmallArea}
S.~Van~Roode, J.~Ruiz-Aguilar, J.~Gonz{\'a}lez-Enrique, and I.~Turias, ``An
  artificial neural network ensemble approach to generate air pollution maps,''
  \emph{Environmental monitoring and assessment}, vol. 191, pp. 1--15, 2019.

\bibitem{chen:2021:MissingLocationAirPollutionMonthly}
C.-C. Chen, Y.-R. Wang, H.-Y. Yeh, T.-H. Lin, C.-S. Huang, and C.-F. Wu,
  ``Estimating monthly pm2. 5 concentrations from satellite remote sensing
  data, meteorological variables, and land use data using ensemble statistical
  modeling and a random forest approach,'' \emph{Environmental Pollution}, vol.
  291, p. 118159, 2021.

\bibitem{he:2023:AirPollutionLeaveOneOutValidationDaily}
Q.~He, T.~Ye, M.~Zhang, and Y.~Yuan, ``Enhancing the reliability of hindcast
  modeling for air pollution using history-informed machine learning and
  satellite remote sensing in china,'' \emph{Atmospheric Environment}, p.
  119994, 2023.

\bibitem{li:2020:AirPollutionLeaveOneOutValidationDaily2}
J.~Li, H.~Zhang, C.-Y. Chao, C.-H. Chien, C.-Y. Wu, C.~H. Luo, L.-J. Chen, and
  P.~Biswas, ``Integrating low-cost air quality sensor networks with fixed and
  satellite monitoring systems to study ground-level pm2. 5,''
  \emph{Atmospheric Environment}, vol. 223, p. 117293, 2020.

\bibitem{berrisford:2022:MLAnnualEnglandWales}
L.~J. Berrisford, E.~Ribeiro, and R.~Menezes, ``Estimating ambient air
  pollution using structural properties of road networks,'' \emph{arXiv
  preprint arXiv:2207.14335}, 2022.

\bibitem{berrisford:2024:MLHourlyEngland}
L.~J. Berrisford, L.~S. Neal, H.~J. Buttery, B.~R. Evans, and R.~Menezes, ``A
  framework for scalable ambient air pollution concentration estimation,''
  \emph{arXiv preprint arXiv:2401.08735}, 2024.

\bibitem{OpenAQ:2023:AboutUs}
\BIBentryALTinterwordspacing
{OpenAQ}. (2023) {About Us}. Accessed on: 29/11/2023. [Online]. Available:
  \url{https://openaq.org/about/}
\BIBentrySTDinterwordspacing

\bibitem{vallero:2014fundamentalsConvertPPBtomicrogram}
D.~A. Vallero, \emph{Fundamentals of air pollution}.\hskip 1em plus 0.5em minus
  0.4em\relax Academic press, 2014.

\bibitem{goldberg:2021:NO2DailyCycles}
D.~L. Goldberg, S.~C. Anenberg, G.~H. Kerr, A.~Mohegh, Z.~Lu, and D.~G.
  Streets, ``Tropomi no2 in the united states: A detailed look at the annual
  averages, weekly cycles, effects of temperature, and correlation with surface
  no2 concentrations,'' \emph{Earth's future}, vol.~9, no.~4, p. e2020EF001665,
  2021.

\bibitem{garland:1979:OzoneDailyCycle}
J.~Garland and R.~Derwent, ``Destruction at the ground and the diurnal cycle of
  concentration of ozone and other gases,'' \emph{Quarterly Journal of the
  Royal Meteorological Society}, vol. 105, no. 443, pp. 169--183, 1979.

\bibitem{beirle:2003:NOxWeekly}
S.~Beirle, U.~Platt, M.~Wenig, and T.~Wagner, ``Weekly cycle of no 2 by gome
  measurements: a signature of anthropogenic sources,'' \emph{Atmospheric
  Chemistry and Physics}, vol.~3, no.~6, pp. 2225--2232, 2003.

\bibitem{gietl:2009:PMWeeklyCycle}
J.~K. Gietl and O.~Klemm, ``Analysis of traffic and meteorology on airborne
  particulate matter in m{\"u}nster, northwest germany,'' \emph{Journal of the
  Air \& Waste Management Association}, vol.~59, no.~7, pp. 809--818, 2009.

\bibitem{feng:2014:WinterPMResidentialHeating}
X.~Feng, Q.~Li, Y.~Zhu, J.~Wang, H.~Liang, and R.~Xu, ``Formation and dominant
  factors of haze pollution over beijing and its peripheral areas in winter,''
  \emph{Atmospheric Pollution Research}, vol.~5, no.~3, pp. 528--538, 2014.

\bibitem{meng:2018:SO2Increases}
K.~Meng, X.~Xu, X.~Cheng, X.~Xu, X.~Qu, W.~Zhu, C.~Ma, Y.~Yang, and Y.~Zhao,
  ``Spatio-temporal variations in so2 and no2 emissions caused by heating over
  the beijing-tianjin-hebei region constrained by an adaptive nudging method
  with omi data,'' \emph{Science of the total environment}, vol. 642, pp.
  543--552, 2018.

\bibitem{Qianhui:2022:StableWinterAirPollutionIncrease}
\BIBentryALTinterwordspacing
Q.~Li, H.~Zhang, X.~Jin, X.~Cai, and Y.~Song, ``Mechanism of haze pollution in
  summer and its difference with winter in the north china plain,''
  \emph{Science of The Total Environment}, vol. 806, p. 150625, 2022. [Online].
  Available:
  \url{https://www.sciencedirect.com/science/article/pii/S004896972105703X}
\BIBentrySTDinterwordspacing

\bibitem{cichowicz:2017:OzoneWinterReduction}
R.~Cichowicz, G.~Wielgosi{\'n}ski, and W.~Fetter, ``Dispersion of atmospheric
  air pollution in summer and winter season,'' \emph{Environmental monitoring
  and assessment}, vol. 189, pp. 1--10, 2017.

\bibitem{Jurado:2021:WindSpeedDirectionAirPollutionConcentrations}
\BIBentryALTinterwordspacing
X.~Jurado, N.~Reiminger, J.~Vazquez, and C.~Wemmert, ``On the minimal wind
  directions required to assess mean annual air pollution concentration based
  on cfd results,'' \emph{Sustainable Cities and Society}, vol.~71, p. 102920,
  2021. [Online]. Available:
  \url{https://www.sciencedirect.com/science/article/pii/S2210670721002079}
\BIBentrySTDinterwordspacing

\bibitem{wallace:2010:temperatureInversionAirPollution}
J.~Wallace, D.~Corr, and P.~Kanaroglou, ``Topographic and spatial impacts of
  temperature inversions on air quality using mobile air pollution surveys,''
  \emph{Science of the total environment}, vol. 408, no.~21, pp. 5086--5096,
  2010.

\bibitem{bloomer:2009:observedOzoneAndTemperatureRelationship}
B.~J. Bloomer, J.~W. Stehr, C.~A. Piety, R.~J. Salawitch, and R.~R. Dickerson,
  ``Observed relationships of ozone air pollution with temperature and
  emissions,'' \emph{Geophysical research letters}, vol.~36, no.~9, 2009.

\bibitem{finlayson:1986atmosphericChemistryOzoneProduction}
B.~J. Finlayson-Pitts and J.~N. Pitts~Jr, ``Atmospheric chemistry. fundamentals
  and experimental techniques,'' 1986.

\bibitem{nowak:1998:VegetationTemperatureAirPollution}
D.~J. Nowak, P.~J. McHale, M.~Ibarra, D.~Crane, J.~C. Stevens, and C.~J. Luley,
  ``Modeling the effects of urban vegetation on air pollution,'' \emph{Air
  pollution modeling and its application XII}, pp. 399--407, 1998.

\bibitem{jolliet:2005:WetDeposition}
O.~Jolliet and M.~Hauschild, ``Modeling the influence of intermittent rain
  events on long-term fate and transport of organic air pollutants,''
  \emph{Environmental science \& technology}, vol.~39, no.~12, pp. 4513--4522,
  2005.

\bibitem{yuan:2017:RainfallRoadWashoff}
Q.~Yuan, H.~B. Guerra, and Y.~Kim, ``An investigation of the relationships
  between rainfall conditions and pollutant wash-off from the paved road,''
  \emph{Water}, vol.~9, no.~4, p. 232, 2017.

\bibitem{xu:2019:RainfallLeafWashOff}
X.~Xu, X.~Yu, L.~Bao, and A.~R. Desai, ``Size distribution of particulate
  matter in runoff from different leaf surfaces during controlled rainfall
  processes,'' \emph{Environmental Pollution}, vol. 255, p. 113234, 2019.

\bibitem{ning:2018:LowPressureSystemAirPollution}
G.~Ning, S.~Wang, S.~H.~L. Yim, J.~Li, Y.~Hu, Z.~Shang, J.~Wang, and J.~Wang,
  ``Impact of low-pressure systems on winter heavy air pollution in the
  northwest sichuan basin, china,'' \emph{Atmospheric Chemistry and Physics},
  vol.~18, no.~18, pp. 13\,601--13\,615, 2018.

\bibitem{vukovich:1979:HighPressureSystemAirPollution}
F.~M. Vukovich, ``A note on air quality in high pressure systems,''
  \emph{Atmospheric Environment (1967)}, vol.~13, no.~2, pp. 255--265, 1979.

\bibitem{hippler:1990:OzoneFormationPressure}
H.~Hippler, R.~Rahn, and J.~Troe, ``Temperature and pressure dependence of
  ozone formation rates in the range 1--1000 bar and 90--370 k,'' \emph{The
  Journal of chemical physics}, vol.~93, no.~9, pp. 6560--6569, 1990.

\bibitem{xiang:2019:BoundaryLayerHeightAirPollution}
Y.~Xiang, T.~Zhang, J.~Liu, L.~Lv, Y.~Dong, and Z.~Chen, ``Atmosphere boundary
  layer height and its effect on air pollutants in beijing during winter heavy
  pollution,'' \emph{Atmospheric Research}, vol. 215, pp. 305--316, 2019.

\bibitem{davies:2007:BoundaryLayerHeightAirPollution}
F.~Davies, D.~Middleton, and K.~Bozier, ``Urban air pollution modelling and
  measurements of boundary layer height,'' \emph{Atmospheric environment},
  vol.~41, no.~19, pp. 4040--4049, 2007.

\bibitem{xiang:2019:BoundaryLayerHeightAirPollution2}
Y.~Xiang, T.~Zhang, J.~Liu, L.~Lv, Y.~Dong, and Z.~Chen, ``Atmosphere boundary
  layer height and its effect on air pollutants in beijing during winter heavy
  pollution,'' \emph{Atmospheric Research}, vol. 215, pp. 305--316, 2019.

\bibitem{hersbach:2016:era5}
H.~Hersbach, ``The era5 atmospheric reanalysis.'' in \emph{AGU fall meeting
  abstracts}, vol. 2016, 2016, pp. NG33D--01.

\bibitem{soulie:2023:EmissionsDatasetAnthrogenic}
A.~Soulie, C.~Granier, S.~Darras, N.~Zilbermann, T.~Doumbia, M.~Guevara, J.-P.
  Jalkanen, S.~Keita, C.~Liousse, M.~Crippa, D.~Guizzardi, R.~Hoesly, and S.~J.
  Smith, ``Global anthropogenic emissions (cams-glob-ant) for the copernicus
  atmosphere monitoring service simulations of air quality forecasts and
  reanalyses,'' \emph{Earth Syst. Sci. Data}, 2023.

\bibitem{granier:2019:EmissionsDatasetAnthrogenicCopernicus}
C.~Granier, S.~Darras, H.~Denier van~der Gon, J.~Doubalova, N.~Elguindi,
  B.~Galle, M.~Gauss, M.~Guevara, J.-P. Jalkanen, J.~Kuenen, C.~Liousse,
  B.~Quack, D.~Simpson, and K.~Sindelarova, ``The copernicus atmosphere
  monitoring service global and regional emissions (april 2019 version),'' 4
  2019, report, April 2019 version.

\bibitem{ECCAD:2023:AboutUs}
\BIBentryALTinterwordspacing
{ECCAD}. (2023) {About Us}. Accessed on: 29/11/2023. [Online]. Available:
  \url{https://eccad.aeris-data.fr/context/}
\BIBentrySTDinterwordspacing

\bibitem{adebiyi:2022:EmissionsPetroleumRefining}
F.~M. Adebiyi, ``Air quality and management in petroleum refining industry: A
  review,'' \emph{Environmental Chemistry and Ecotoxicology}, vol.~4, pp.
  89--96, 2022.

\bibitem{corbett:1997:ShipEmissions}
J.~J. Corbett and P.~Fischbeck, ``Emissions from ships,'' \emph{Science}, vol.
  278, no. 5339, pp. 823--824, 1997.

\bibitem{tao:2013:ShipHighSulfur}
L.~Tao, D.~Fairley, M.~J. Kleeman, and R.~A. Harley, ``Effects of switching to
  lower sulfur marine fuel oil on air quality in the san francisco bay area,''
  \emph{Environmental science \& technology}, vol.~47, no.~18, pp.
  10\,171--10\,178, 2013.

\bibitem{zisi:2021:EUEmissionsImpact}
V.~Zisi, H.~N. Psaraftis, and T.~Zis, ``The impact of the 2020 global sulfur
  cap on maritime co2 emissions,'' \emph{Maritime Business Review}, vol.~6,
  no.~4, pp. 339--357, 2021.

\bibitem{sotoodeh:2022:FugitiveEmissionsDefinition}
K.~Sotoodeh, \emph{Case studies of material corrosion prevention for oil and
  gas valves}.\hskip 1em plus 0.5em minus 0.4em\relax Gulf Professional
  Publishing, 2022.

\bibitem{santacatalina:2010:FugitiveEmissionsImpactPM}
M.~Santacatalina, C.~Reche, M.~Minguill{\'o}n, A.~Escrig, V.~Sanfelix,
  A.~Carratal{\'a}, J.~Nicol{\'a}s, E.~Yubero, J.~Crespo, A.~Alastuey
  \emph{et~al.}, ``Impact of fugitive emissions in ambient pm levels and
  composition: A case study in southeast spain,'' \emph{Science of The Total
  Environment}, vol. 408, no.~21, pp. 4999--5009, 2010.

\bibitem{qian:2020:ChinaCoalSO2Emissions}
Y.~Qian, L.~Scherer, A.~Tukker, and P.~Behrens, ``China's potential so2
  emissions from coal by 2050,'' \emph{Energy Policy}, vol. 147, p. 111856,
  2020.

\bibitem{querol:1995:CoalSO2Content}
X.~Querol, J.~Fern{\'a}ndez-Turiel, and A.~L{\'o}pez-Soler, ``Trace elements in
  coal and their behaviour during combustion in a large power station,''
  \emph{Fuel}, vol.~74, no.~3, pp. 331--343, 1995.

\bibitem{shi:2021:OilSulfurContent}
Q.~Shi and J.~Wu, ``Review on sulfur compounds in petroleum and its products:
  State-of-the-art and perspectives,'' \emph{Energy \& Fuels}, vol.~35, no.~18,
  pp. 14\,445--14\,461, 2021.

\bibitem{chaaban:2004:NaturalGasSulfurEmissions}
F.~Chaaban, T.~Mezher, and M.~Ouwayjan, ``Options for emissions reduction from
  power plants: an economic evaluation,'' \emph{International journal of
  electrical power \& energy systems}, vol.~26, no.~1, pp. 57--63, 2004.

\bibitem{zhang:2007:BiomassHouseholdAirPollutionSulfur}
J.~Zhang and K.~R. Smith, ``Household air pollution from coal and biomass fuels
  in china: measurements, health impacts, and interventions,''
  \emph{Environmental health perspectives}, vol. 115, no.~6, pp. 848--855,
  2007.

\bibitem{fuller:2014:TrainEmissions}
G.~Fuller, T.~Baker, A.~Tremper, D.~Green, A.~Font, M.~Priestman, D.~Carslaw,
  D.~Dajnak, and S.~Beevers, ``Air pollution emissions from diesel trains in
  london,'' \emph{King’s College London}, 2014.

\bibitem{barrett:2010:AeroplanesEmissions}
S.~R. Barrett, R.~E. Britter, and I.~A. Waitz, ``Global mortality attributable
  to aircraft cruise emissions,'' \emph{Environmental science \& technology},
  vol.~44, no.~19, pp. 7736--7742, 2010.

\bibitem{gonzalez:2016:AgriculturalEquipmentEmissions}
M.~Gonzalez-de Soto, L.~Emmi, C.~Benavides, I.~Garcia, and P.~Gonzalez-de
  Santos, ``Reducing air pollution with hybrid-powered robotic tractors for
  precision agriculture,'' \emph{Biosystems Engineering}, vol. 143, pp. 79--94,
  2016.

\bibitem{wang:2023:constructionEmissions}
C.~Wang, W.~Duan, S.~Cheng, and K.~Jiang, ``Emission inventory and air quality
  impact of non-road construction equipment in different emission stages,''
  \emph{Science of The Total Environment}, p. 167416, 2023.

\bibitem{barati:2015:MiningEquipmentEmissions}
K.~Barati and X.~Shen, ``Modeling emissions of construction and mining
  equipment by tracking field operations,'' in \emph{ISARC. Proceedings of the
  International Symposium on Automation and Robotics in Construction},
  vol.~32.\hskip 1em plus 0.5em minus 0.4em\relax IAARC Publications, 2015,
  p.~1.

\bibitem{watkins1991:RoadPollutionAirPollution}
L.~Watkins, ``Air pollution from road vehicles,'' 1991.

\bibitem{yan:2011:RoadVechilesPMEmissions}
F.~Yan, E.~Winijkul, S.~Jung, T.~C. Bond, and D.~G. Streets, ``Global emission
  projections of particulate matter (pm): I. exhaust emissions from on-road
  vehicles,'' \emph{Atmospheric Environment}, vol.~45, no.~28, pp. 4830--4844,
  2011.

\bibitem{archer:2016:ResidentialCookingImpact}
S.~Archer-Nicholls, E.~Carter, R.~Kumar, Q.~Xiao, Y.~Liu, J.~Frostad, M.~H.
  Forouzanfar, A.~Cohen, M.~Brauer, J.~Baumgartner \emph{et~al.}, ``The
  regional impacts of cooking and heating emissions on ambient air quality and
  disease burden in china,'' \emph{Environmental science \& technology},
  vol.~50, no.~17, pp. 9416--9423, 2016.

\bibitem{LIANG:2020:VOCEmissionsInventoryIndustry}
\BIBentryALTinterwordspacing
X.~Liang, X.~Sun, J.~Xu, and D.~Ye, ``Improved emissions inventory and vocs
  speciation for industrial ofp estimation in china,'' \emph{Science of The
  Total Environment}, vol. 745, p. 140838, 2020. [Online]. Available:
  \url{https://www.sciencedirect.com/science/article/pii/S004896972034362X}
\BIBentrySTDinterwordspacing

\bibitem{ehrlich:2007:PMChemicalIndustryEmissions}
C.~Ehrlich, G.~Noll, W.-D. Kalkoff, G.~Baumbach, and A.~Dreiseidler, ``Pm10,
  pm2. 5 and pm1. 0—emissions from industrial plants—results from
  measurement programmes in germany,'' \emph{Atmospheric Environment}, vol.~41,
  no.~29, pp. 6236--6254, 2007.

\bibitem{guo:2022:SO2NOXChemical}
Y.~Guo, L.~Zhu, X.~Wang, X.~Qiu, W.~Qian, and L.~Wang, ``Assessing
  environmental impact of nox and so2 emissions in textiles production with
  chemical footprint,'' \emph{Science of The Total Environment}, vol. 831, p.
  154961, 2022.

\bibitem{meo:2004:DustCementEmisions}
S.~A. Meo, ``Health hazards of cement dust.'' \emph{Saudi medical journal},
  vol.~25, no.~9, pp. 1153--1159, 2004.

\bibitem{cabanes:2020:PlasticProductionVOC}
A.~Cabanes, F.~J. Vald{\'e}s, and A.~Fullana, ``A review on vocs from recycled
  plastics,'' \emph{Sustainable materials and technologies}, vol.~25, p.
  e00179, 2020.

\bibitem{yuan:2010:SolventSourceUse}
B.~Yuan, M.~Shao, S.~Lu, and B.~Wang, ``Source profiles of volatile organic
  compounds associated with solvent use in beijing, china,'' \emph{Atmospheric
  Environment}, vol.~44, no.~15, pp. 1919--1926, 2010.

\bibitem{lal:2008:AgriculturalWasteBurning}
M.~Lal, ``An over view to agriculture! waste burning,'' \emph{Indian Journal of
  Air Pollution Control Vol. VIII No. IMarch}, pp. 48--50, 2008.

\bibitem{kumar:2013:AgriculturalPollution}
P.~Kumar and L.~Joshi, ``Pollution caused by agricultural waste burning and
  possible alternate uses of crop stubble: a case study of punjab,''
  \emph{Knowledge systems of societies for adaptation and mitigation of impacts
  of climate change}, pp. 367--385, 2013.

\bibitem{huber:2008:LandfillBiogenicEmissions}
M.~Huber-Humer, J.~Gebert, and H.~Hilger, ``Biotic systems to mitigate landfill
  methane emissions,'' \emph{Waste Management \& Research}, vol.~26, no.~1, pp.
  33--46, 2008.

\bibitem{nair:2019:LandfillVOCEmissions}
A.~T. Nair, J.~Senthilnathan, and S.~S. Nagendra, ``Emerging perspectives on
  voc emissions from landfill sites: Impact on tropospheric chemistry and local
  air quality,'' \emph{Process safety and environmental protection}, vol. 121,
  pp. 143--154, 2019.

\bibitem{rani:2008:WasteIncineration}
D.~A. Rani, A.~Boccaccini, D.~Deegan, and C.~R. Cheeseman, ``Air pollution
  control residues from waste incineration: current uk situation and assessment
  of alternative technologies,'' \emph{Waste Management}, vol.~28, no.~11, pp.
  2279--2292, 2008.

\bibitem{law:2012:WasteWaterTreatmentEmissions}
Y.~Law, L.~Ye, Y.~Pan, and Z.~Yuan, ``Nitrous oxide emissions from wastewater
  treatment processes,'' \emph{Philosophical Transactions of the Royal Society
  B: Biological Sciences}, vol. 367, no. 1593, pp. 1265--1277, 2012.

\bibitem{sindelarova:2022:BIogenicEmissionsOne}
K.~Sindelarova, J.~Markova, D.~Simpson, P.~Huszar, J.~Karlicky, S.~Darras, and
  C.~Granier, ``High-resolution biogenic global emission inventory for the time
  period 2000--2019 for air quality modelling,'' \emph{Earth Syst. Sci. Data},
  vol.~14, pp. 251--270, 2022.

\bibitem{sindelarova:2014:BIogenicEmissionsTwo}
K.~Sindelarova, C.~Granier, I.~Bouarar, A.~Guenther, S.~Tilmes, T.~Stavrakou,
  J.-F. M{\"u}ller, U.~Kuhn, P.~Stefani, and W.~Knorr, ``Global dataset of
  biogenic voc emissions calculated by the megan model over the last 30
  years,'' \emph{Atmospheric Chemistry \& Physics}, vol.~14, pp. 9317--9341,
  2014, reviewer link to CAMS-GLOB-BIO snapshot dataset for ESSD special issue
  on surface emissions.

\bibitem{granier:2019:BiogenicEmissionsThree}
C.~Granier, S.~Darras, H.~Denier van~der Gon, J.~Doubalova, N.~Elguindi,
  B.~Galle, M.~Gauss, M.~Guevara, J.-P. Jalkanen, J.~Kuenen, C.~Liousse,
  B.~Quack, D.~Simpson, and K.~Sindelarova, ``The copernicus atmosphere
  monitoring service global and regional emissions (april 2019 version),'' 4
  2019, report, April 2019 version.

\bibitem{hudman:2008:biogenicCOEmissions}
R.~C. Hudman, L.~T. Murray, D.~J. Jacob, D.~Millet, S.~Turquety, S.~Wu,
  D.~Blake, A.~Goldstein, J.~Holloway, and G.~W. Sachse, ``Biogenic versus
  anthropogenic sources of co in the united states,'' \emph{Geophysical
  Research Letters}, vol.~35, no.~4, 2008.

\bibitem{ke:2017:lightgbmDefinition}
G.~Ke, Q.~Meng, T.~Finley, T.~Wang, W.~Chen, W.~Ma, Q.~Ye, and T.-Y. Liu,
  ``Lightgbm: A highly efficient gradient boosting decision tree,''
  \emph{Advances in neural information processing systems}, vol.~30, 2017.

\bibitem{van:2018:AirPollutionOutlier}
V.~Van~Zoest, A.~Stein, and G.~Hoek, ``Outlier detection in urban air quality
  sensor networks,'' \emph{Water, Air, \& Soil Pollution}, vol. 229, pp. 1--13,
  2018.

\bibitem{rollo:2023:AirPollutionAnaomaly}
F.~Rollo, C.~Bachechi, and L.~Po, ``Anomaly detection and repairing for
  improving air quality monitoring,'' \emph{Sensors}, vol.~23, no.~2, p. 640,
  2023.

\bibitem{hodson:2022:RMSEMAE}
T.~O. Hodson, ``Root-mean-square error (rmse) or mean absolute error (mae):
  When to use them or not,'' \emph{Geoscientific Model Development}, vol.~15,
  no.~14, pp. 5481--5487, 2022.

\bibitem{yang:2022:nonlinearAirPollution}
L.~Yang, J.~Yang, M.~Liu, X.~Sun, T.~Li, Y.~Guo, K.~Hu, M.~L. Bell, Q.~Cheng,
  H.~Kan \emph{et~al.}, ``Nonlinear effect of air pollution on adult pneumonia
  hospital visits in the coastal city of qingdao, china: A time-series
  analysis,'' \emph{Environmental Research}, vol. 209, p. 112754, 2022.

\bibitem{zhao:2019:nonlinearAirPollution2}
B.~Zhao, S.~Wang, D.~Ding, W.~Wu, X.~Chang, J.~Wang, J.~Xing, C.~Jang, J.~S.
  Fu, Y.~Zhu \emph{et~al.}, ``Nonlinear relationships between air pollutant
  emissions and pm2. 5-related health impacts in the beijing-tianjin-hebei
  region,'' \emph{Science of the Total Environment}, vol. 661, pp. 375--385,
  2019.

\bibitem{hoerl:1970:RidgeRegression}
A.~E. Hoerl and R.~W. Kennard, ``Ridge regression: Biased estimation for
  nonorthogonal problems,'' \emph{Technometrics}, vol.~12, no.~1, pp. 55--67,
  1970.

\bibitem{LightGBM:2023:Parameters}
\BIBentryALTinterwordspacing
{Microsoft}. (2023) {LightGBM - Parameters}. Accessed on: 29/11/2023. [Online].
  Available: \url{https://lightgbm.readthedocs.io/en/latest/Parameters.html}
\BIBentrySTDinterwordspacing

\bibitem{Github:2017:ParameterGridSuggestions}
\BIBentryALTinterwordspacing
{Salil Mishra}. (2023) {Hyper parameter optimization - suggested parameter grid
  · ISSUE \#695 · Microsoft/LIGHTGBM}. Accessed on: 29/11/2023. [Online].
  Available: \url{https://github.com/microsoft/LightGBM/issues/695}
\BIBentrySTDinterwordspacing

\bibitem{Ng:2020:CrossValidation}
\BIBentryALTinterwordspacing
A.~Ng. (2020) {Lecture 8 - data splits, Models and; cross-validation | Stanford
  CS229: Machine learning (Autumn 2018)}. Accessed on: 29/11/2023. [Online].
  Available: \url{https://www.youtube.com/watch?v=rjbkWSTjHzM}
\BIBentrySTDinterwordspacing

\bibitem{scikit:2023:R2Score}
\BIBentryALTinterwordspacing
{sklearn}. (2023) {sklearn metric r2 score}. Accessed on: 29/11/2023. [Online].
  Available:
  \url{https://scikit-learn.org/stable/modules/generated/sklearn.metrics.r2_score.html}
\BIBentrySTDinterwordspacing

\bibitem{CPCB:2023:AirPollutionData}
\BIBentryALTinterwordspacing
{Central Pollution Control Board of India}. (2023) {Scientific and Technical
  Activity : (Information Technology) - Integrated Transmission of Real-Time
  Information from Continuous Ambient Air Quality Monitoring Stations (CAAQMS)
  to CPCB Server}. Accessed on: 29/11/2023. [Online]. Available:
  \url{https://cpcb.nic.in/uploads/it_technical_activity.pdf}
\BIBentrySTDinterwordspacing

\end{thebibliography}

\end{document}


\maketitle

\begin{figure}
  \hspace*{\fill}   
  \begin{subfigure}{0.49\textwidth}
    \includegraphics[width=\linewidth]{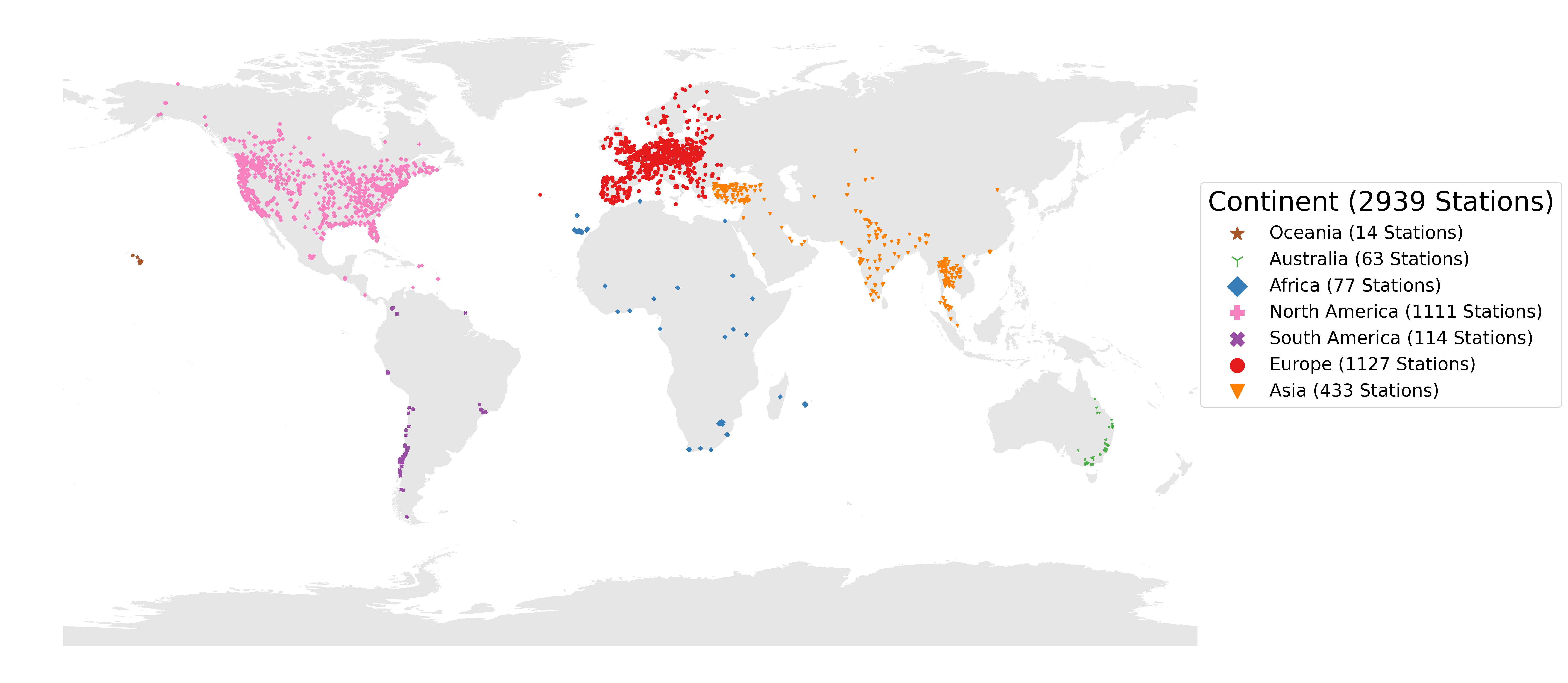}
    \caption{PM$_{2.5}$} \label{fig:openAQMSPM25}
  \end{subfigure}%
  \hspace*{\fill}   
  \begin{subfigure}{0.49\textwidth}
    \includegraphics[width=\linewidth]{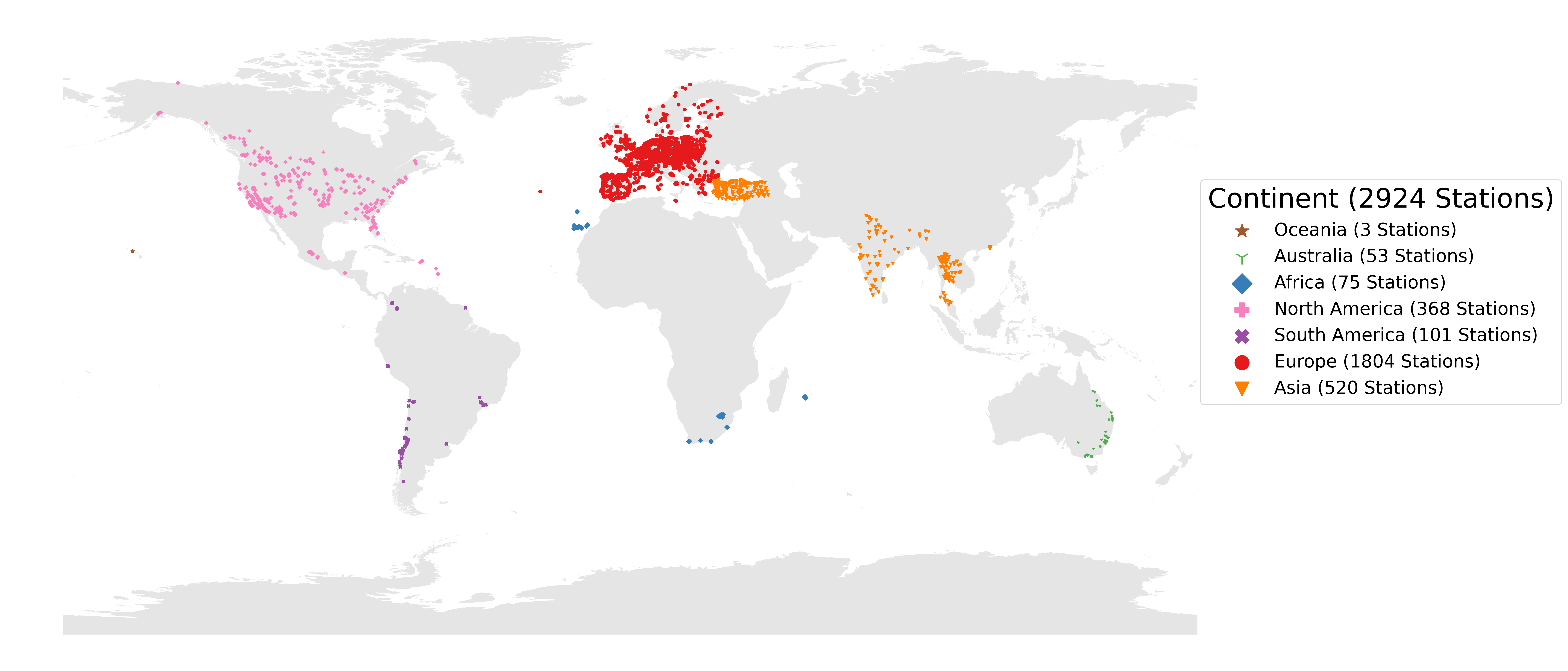}
    \caption{PM$_{10}$} \label{fig:openAQMSPM10}
  \end{subfigure}%
  \hspace*{\fill}   
  \\
  \hspace*{\fill}   
  \begin{subfigure}{0.49\textwidth}
    \includegraphics[width=\linewidth]{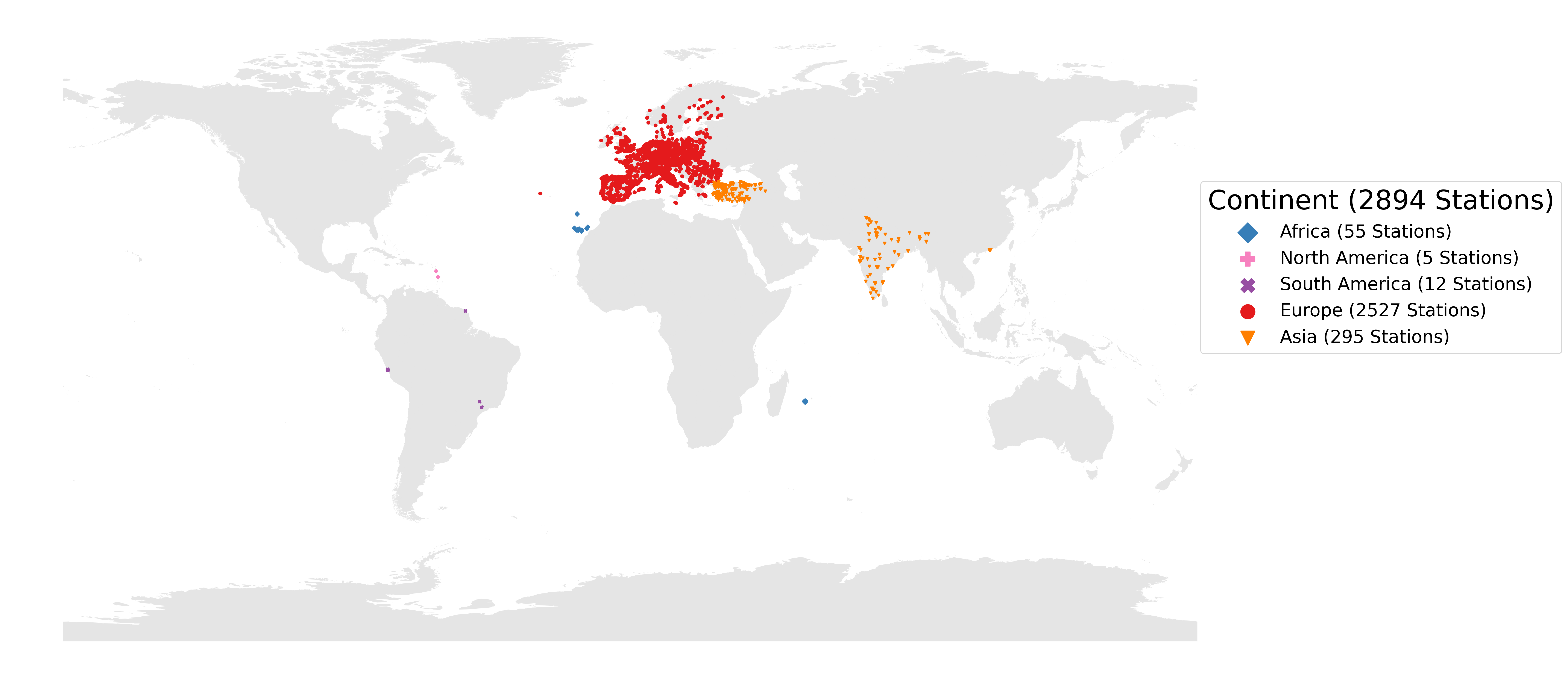}
    \caption{NO$_2$} \label{fig:openAQMSNO2}
  \end{subfigure}
  \hspace*{\fill}   
  \begin{subfigure}{0.49\textwidth}
    \includegraphics[width=\linewidth]{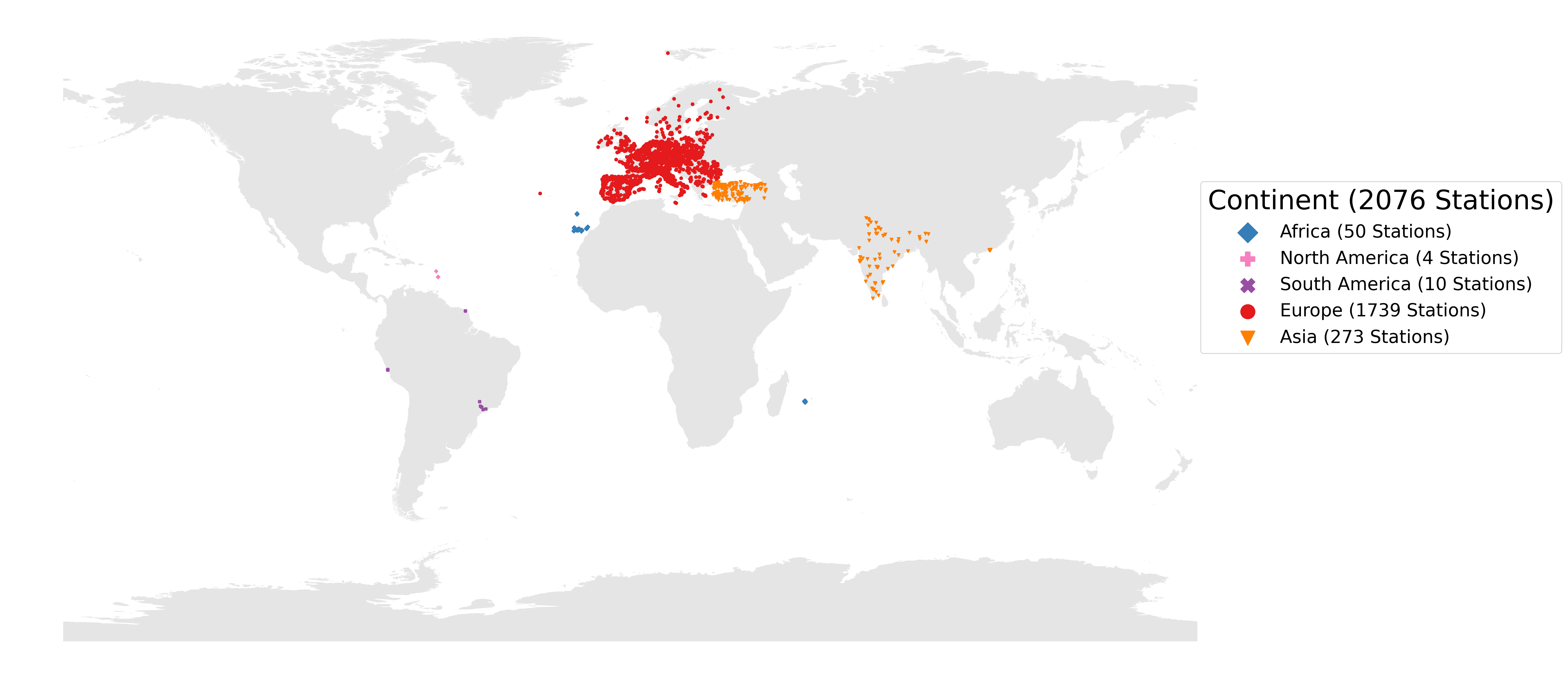}
    \caption{O$_3$} \label{fig:openAQMS03}
  \end{subfigure}
  \hspace*{\fill}   
  \\
  \hspace*{\fill}   
  \begin{subfigure}{0.49\textwidth}
    \includegraphics[width=\linewidth]{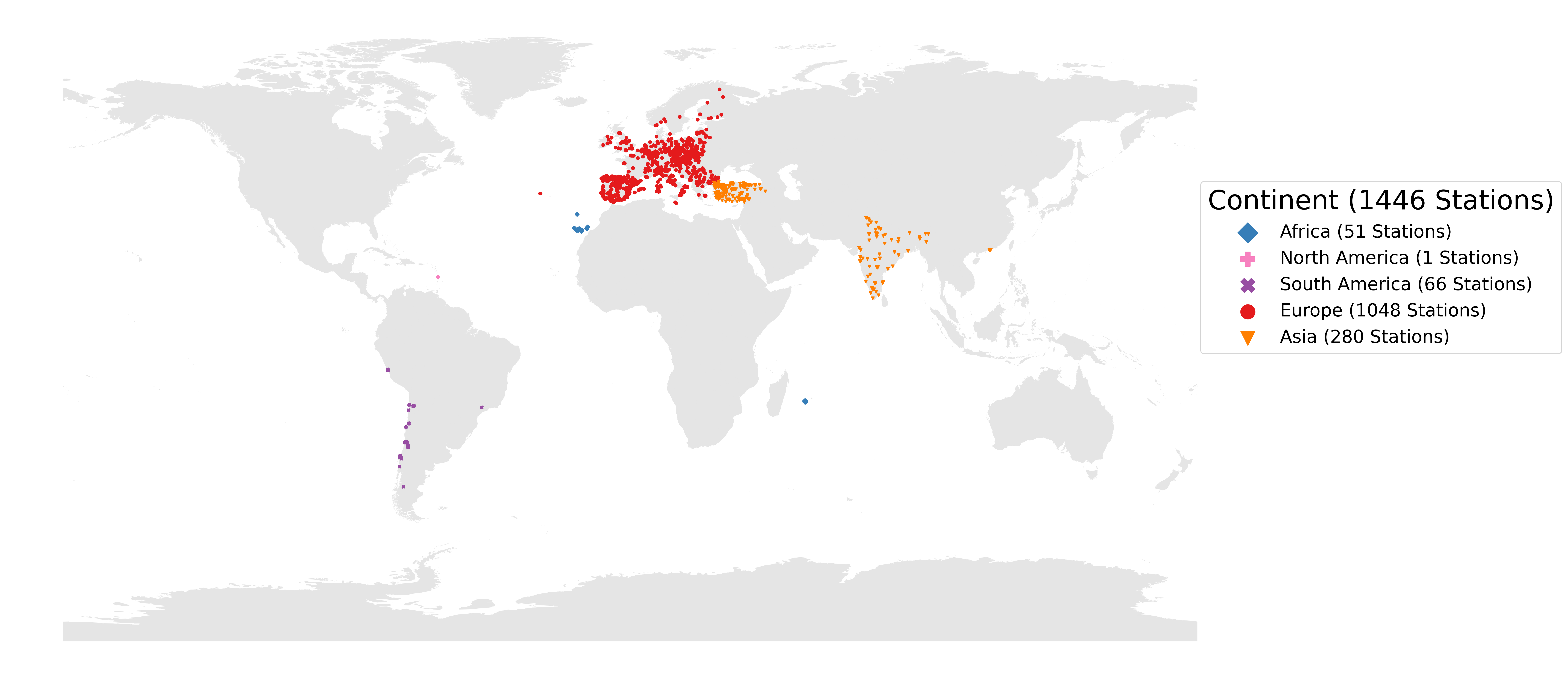}
    \caption{SO$_2$} \label{fig:openAQMSSO2}
  \end{subfigure}
  \hspace*{\fill}   
\caption{{\bfseries Spatial Distribution of Monitoring Station Locations Within the OpenAQ Dataset for 2022 for Individual Air Pollutants} The impact of removing monitoring stations that measure in ppb on spatial distribution can be seen for NO$_2$, O$_3$ and SO$_2$ with a distinct lack of monitoring stations in North America.} \label{fig:allOpenAQMonitoringStationLocations}
\end{figure}

\begin{figure}[!htb]
    \hspace*{\fill}   
  \begin{subfigure}{0.23\textwidth}
    \includegraphics[width=\linewidth]{Figures/ModelPerformance_Bias_Correlation/correlation_bias_plot_no2_Temporal.png}
    \caption{NO$_2$ Baseline}
  \end{subfigure}%
  \hspace*{\fill}   
  \begin{subfigure}{0.23\textwidth}
    \includegraphics[width=\linewidth]{Figures/ModelPerformance_Bias_Correlation/correlation_bias_plot_no2_KFold.png}
    \caption{NO$_2$ Within Network}
  \end{subfigure}%
  \hspace*{\fill}   
  \begin{subfigure}{0.23\textwidth}
    \includegraphics[width=\linewidth]{Figures/ModelPerformance_Bias_Correlation/correlation_bias_plot_no2_Country.png}
    \caption{NO$_2$ Between Country}
  \end{subfigure}%
  \hspace*{\fill}   
  \begin{subfigure}{0.23\textwidth}
    \includegraphics[width=\linewidth]{Figures/ModelPerformance_Bias_Correlation/correlation_bias_plot_no2_Continent.png}
    \caption{NO$_2$ Between Continent}
  \end{subfigure}%
  \hspace*{\fill}   
  \raisebox{10mm}{
  \begin{subfigure}{0.08\textwidth}
    \includegraphics[width=\linewidth]{Figures/ModelPerformance_Bias_Correlation/correlation_bias_plot_legend.png}
  \end{subfigure}%
  }
  \hspace*{\fill}   
  \\
   \hspace*{\fill}   
  \begin{subfigure}{0.23\textwidth}
    \includegraphics[width=\linewidth]{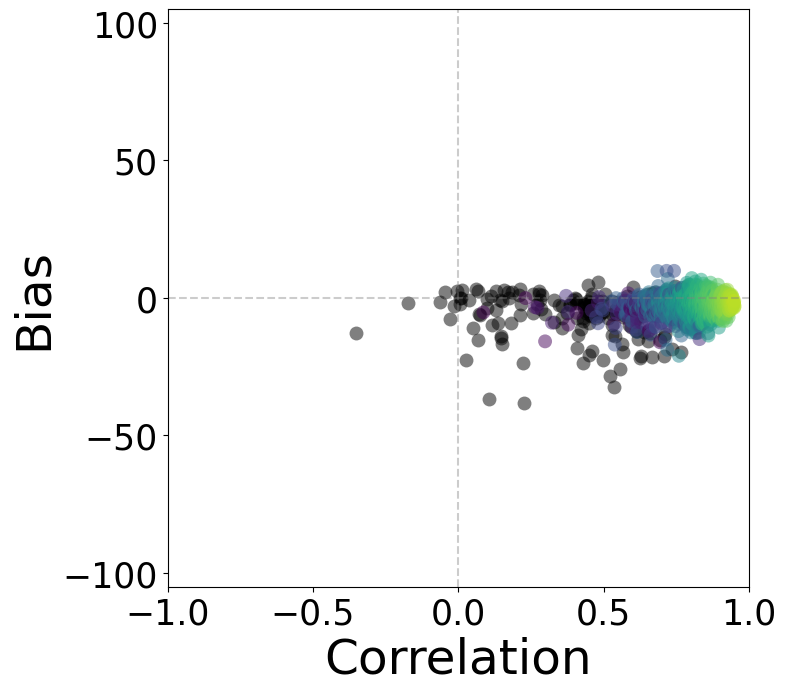}
    \caption{O$_3$ Baseline}
  \end{subfigure}%
  \hspace*{\fill}   
  \begin{subfigure}{0.23\textwidth}
    \includegraphics[width=\linewidth]{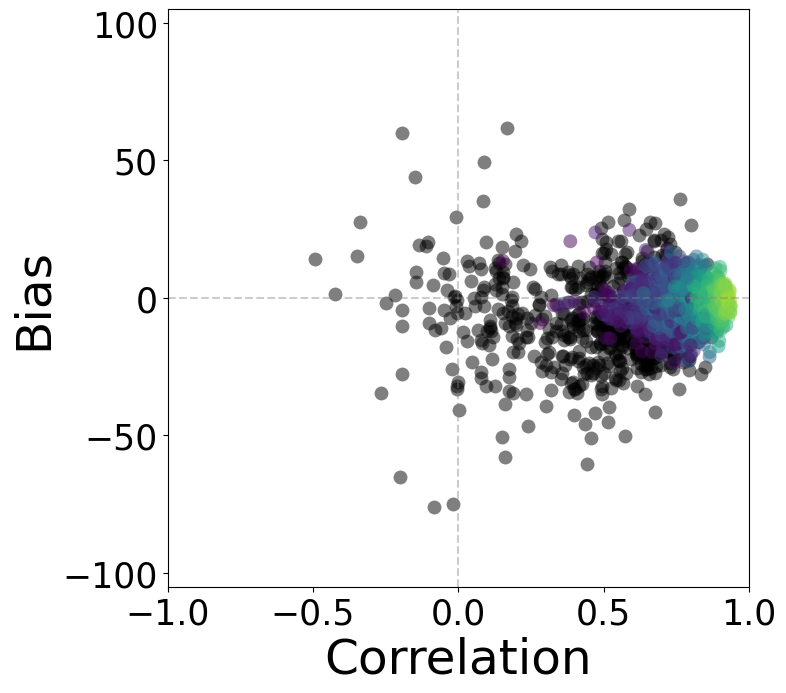}
    \caption{O$_3$ Within Network}
  \end{subfigure}%
  \hspace*{\fill}   
  \begin{subfigure}{0.23\textwidth}
    \includegraphics[width=\linewidth]{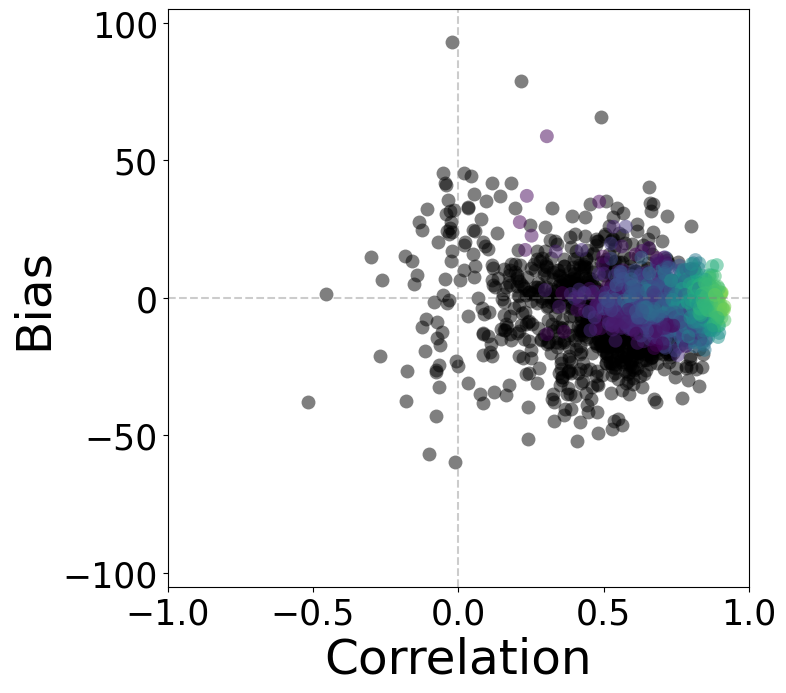}
    \caption{O$_3$ Between Country}
  \end{subfigure}%
  \hspace*{\fill}   
  \begin{subfigure}{0.23\textwidth}
    \includegraphics[width=\linewidth]{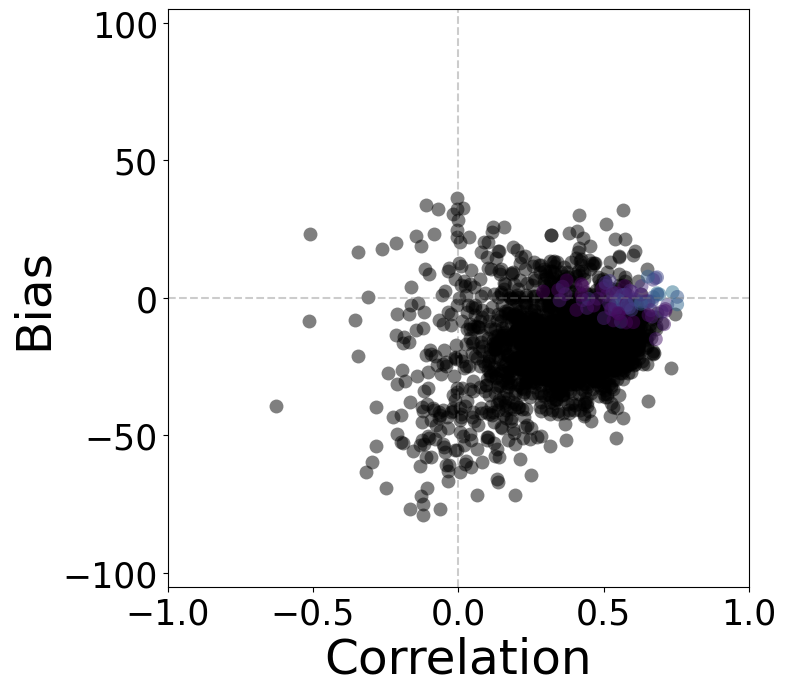}
    \caption{O$_3$ Between Continent}
  \end{subfigure}%
  \hspace*{\fill}   
  \raisebox{10mm}{
  \begin{subfigure}{0.08\textwidth}
    \includegraphics[width=\linewidth]{Figures/ModelPerformance_Bias_Correlation/correlation_bias_plot_legend.png}
  \end{subfigure}%
  }
  \hspace*{\fill}   
  \\
   \hspace*{\fill}   
  \begin{subfigure}{0.23\textwidth}
    \includegraphics[width=\linewidth]{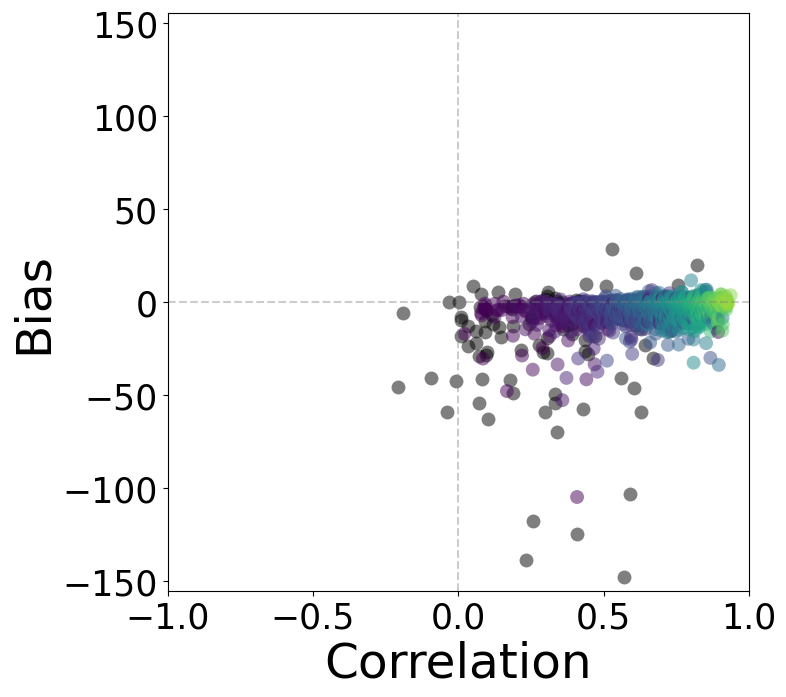}
    \caption{PM$_{10}$ Baseline}
  \end{subfigure}%
  \hspace*{\fill}   
  \begin{subfigure}{0.23\textwidth}
    \includegraphics[width=\linewidth]{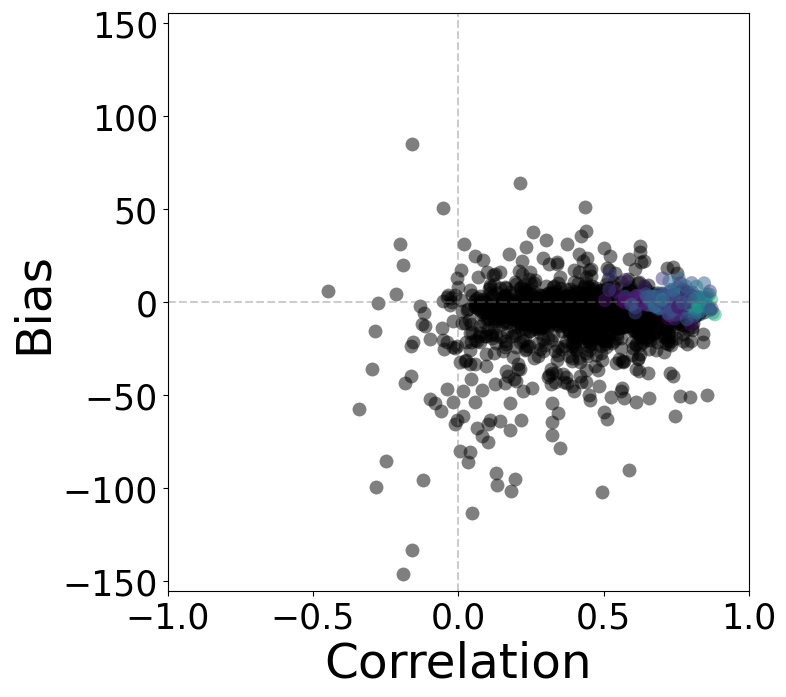}
    \caption{PM$_{10}$ Within Network}
  \end{subfigure}%
  \hspace*{\fill}   
  \begin{subfigure}{0.23\textwidth}
    \includegraphics[width=\linewidth]{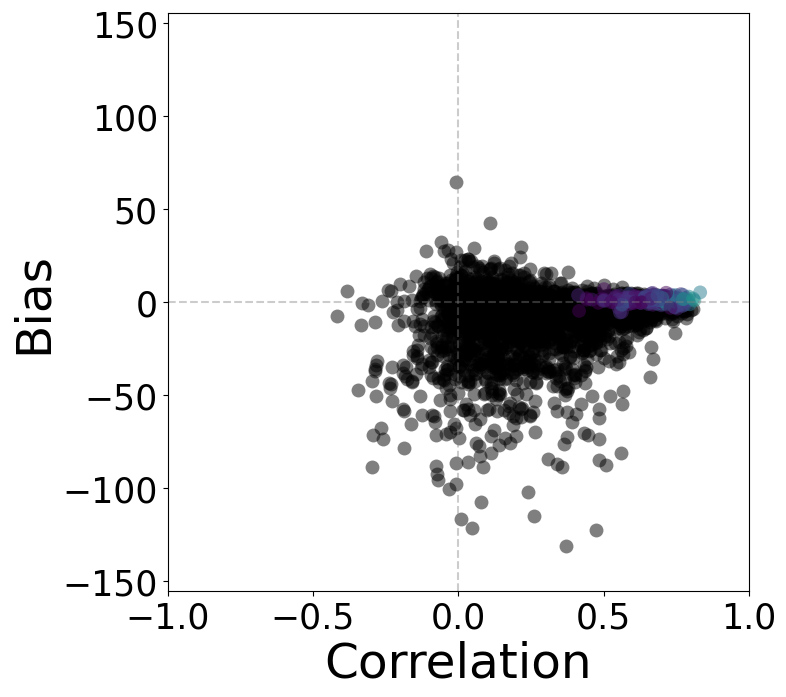}
    \caption{PM$_{10}$ Between Country}
  \end{subfigure}%
  \hspace*{\fill}   
  \begin{subfigure}{0.23\textwidth}
    \includegraphics[width=\linewidth]{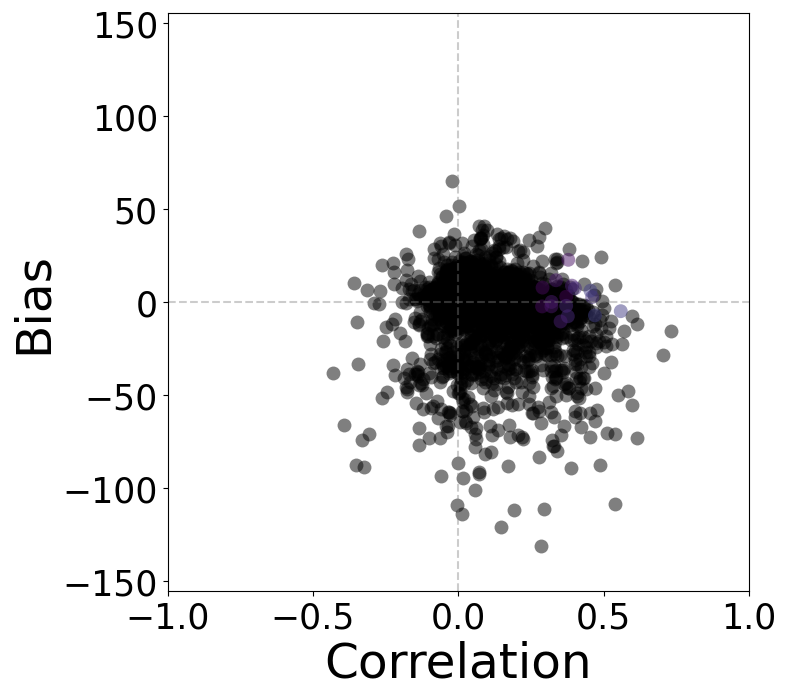}
    \caption{PM$_{10}$ Between Continent}
  \end{subfigure}%
  \hspace*{\fill}   
  \raisebox{10mm}{
  \begin{subfigure}{0.08\textwidth}
    \includegraphics[width=\linewidth]{Figures/ModelPerformance_Bias_Correlation/correlation_bias_plot_legend.png}
  \end{subfigure}%
  }
  \hspace*{\fill}   
  \\
   \hspace*{\fill}   
  \begin{subfigure}{0.23\textwidth}
    \includegraphics[width=\linewidth]{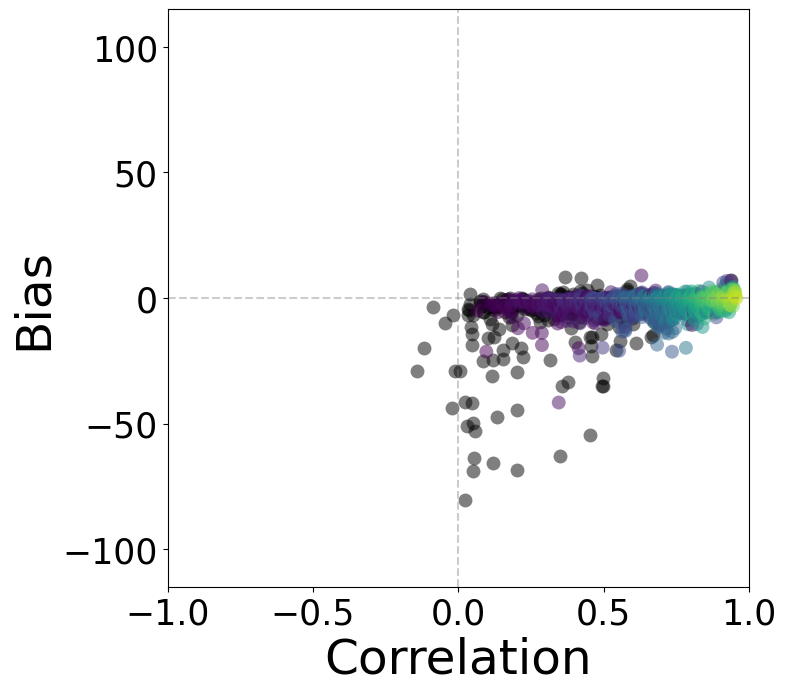}
    \caption{PM$_{2.5}$ Baseline}
  \end{subfigure}%
  \hspace*{\fill}   
  \begin{subfigure}{0.23\textwidth}
    \includegraphics[width=\linewidth]{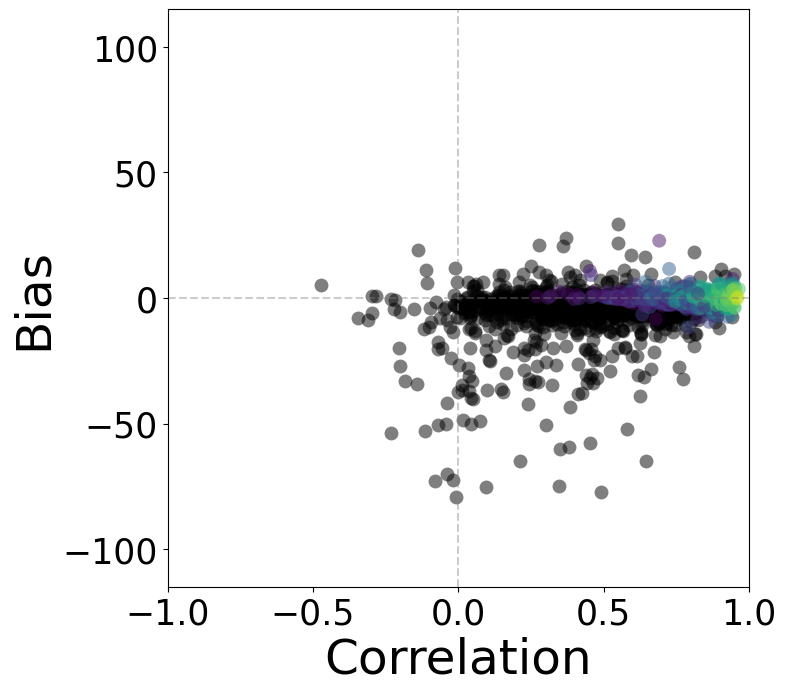}
    \caption{PM$_{2.5}$ Within Network}
  \end{subfigure}%
  \hspace*{\fill}   
  \begin{subfigure}{0.23\textwidth}
    \includegraphics[width=\linewidth]{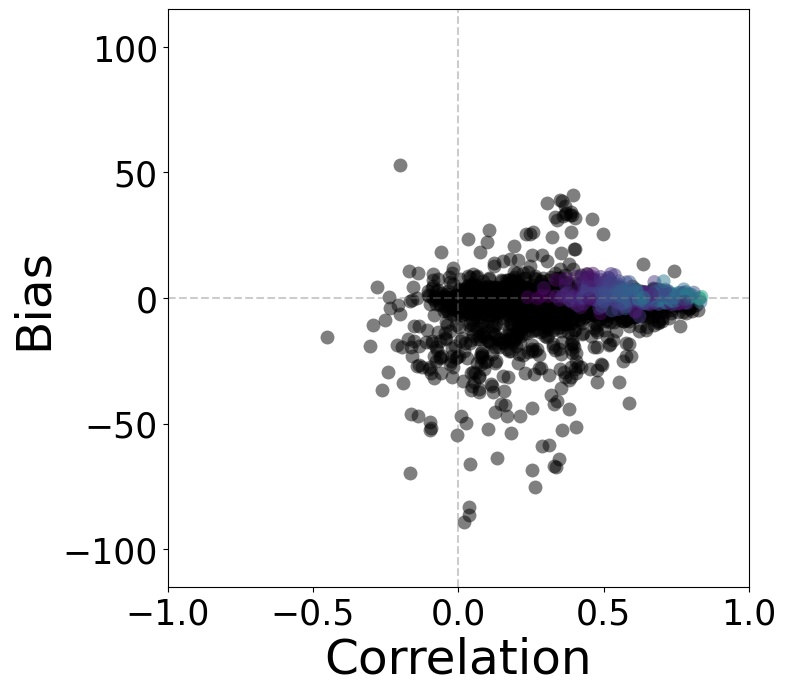}
    \caption{PM$_{2.5}$ Between Country}
  \end{subfigure}%
  \hspace*{\fill}   
  \begin{subfigure}{0.23\textwidth}
    \includegraphics[width=\linewidth]{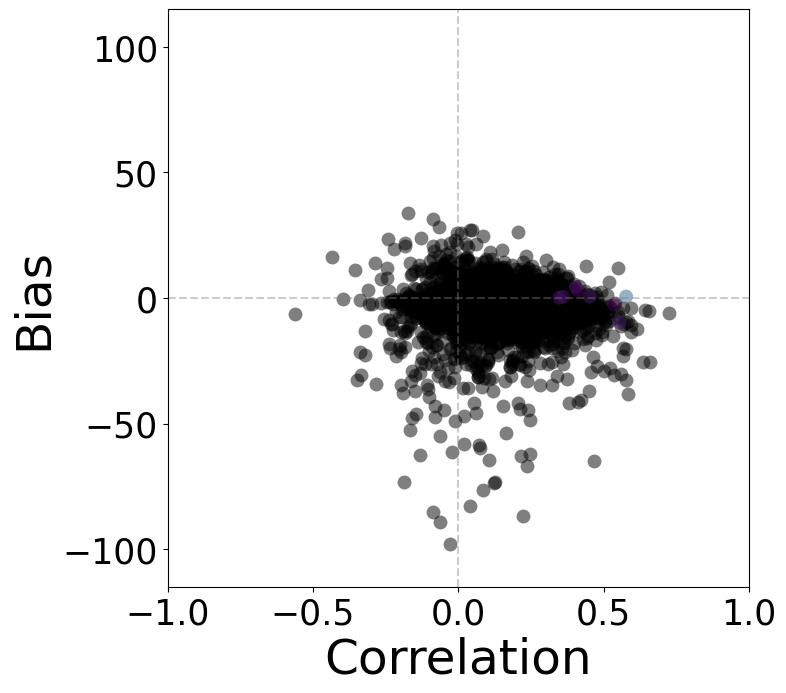}
    \caption{PM$_{2.5}$ Between Continent}
  \end{subfigure}%
  \hspace*{\fill}   
  \raisebox{10mm}{
  \begin{subfigure}{0.08\textwidth}
    \includegraphics[width=\linewidth]{Figures/ModelPerformance_Bias_Correlation/correlation_bias_plot_legend.png}
  \end{subfigure}%
  }
  \hspace*{\fill}   
  \\
   \hspace*{\fill}   
  \begin{subfigure}{0.23\textwidth}
    \includegraphics[width=\linewidth]{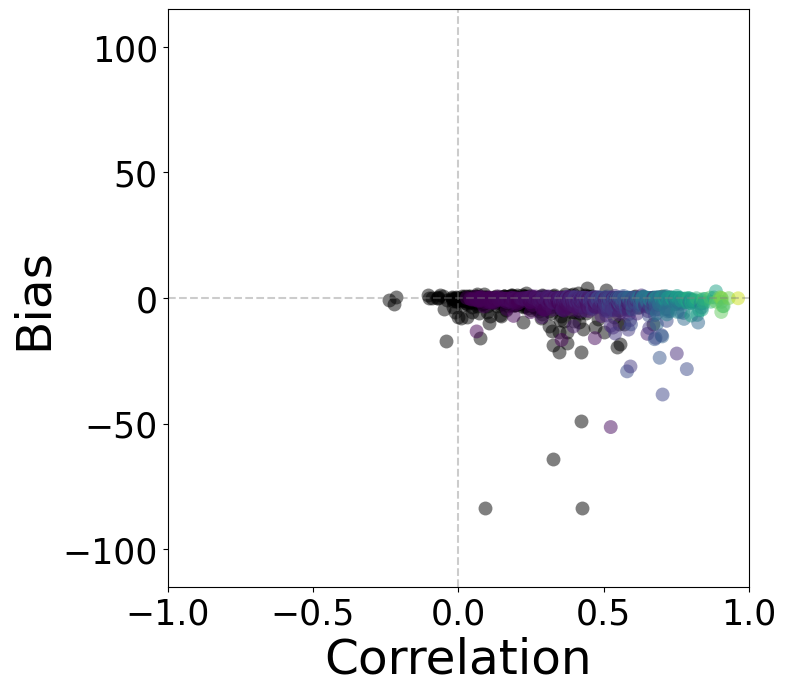}
    \caption{SO$_2$ Baseline} 
  \end{subfigure}%
  \hspace*{\fill}   
  \begin{subfigure}{0.23\textwidth}
    \includegraphics[width=\linewidth]{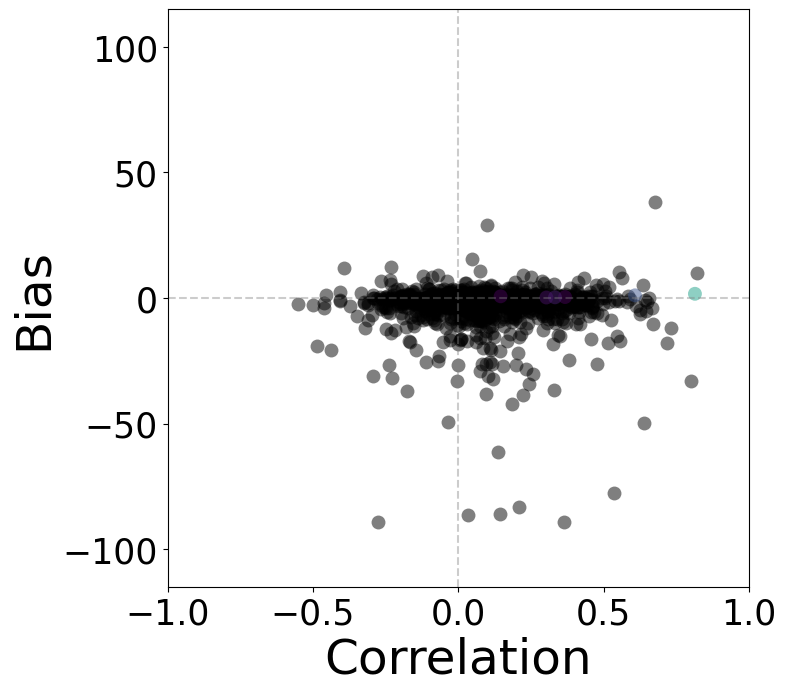}
    \caption{SO$_2$ Within Network} 
  \end{subfigure}%
  \hspace*{\fill}   
  \begin{subfigure}{0.23\textwidth}
    \includegraphics[width=\linewidth]{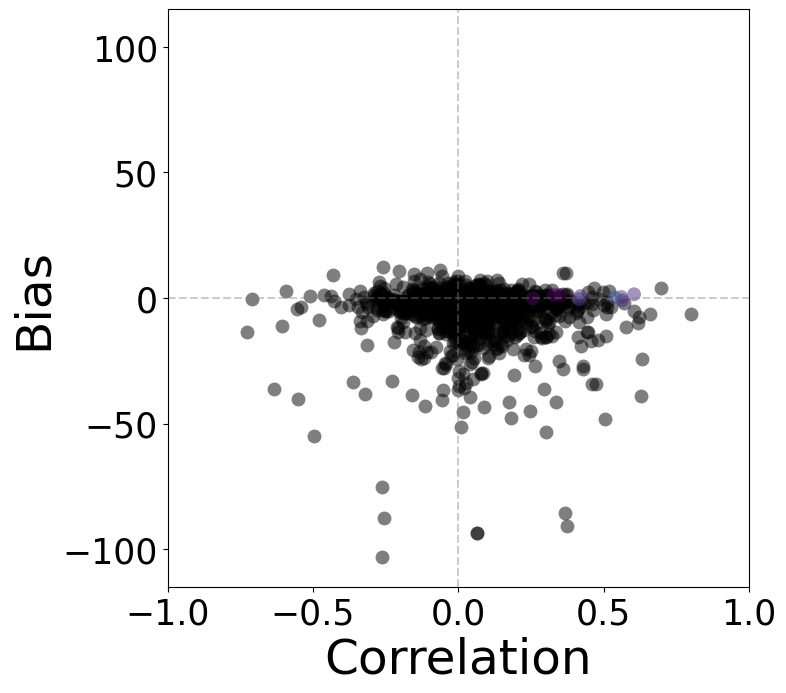}
    \caption{SO$_2$ Between Country} 
  \end{subfigure}%
  \hspace*{\fill}   
  \begin{subfigure}{0.23\textwidth}
    \includegraphics[width=\linewidth]{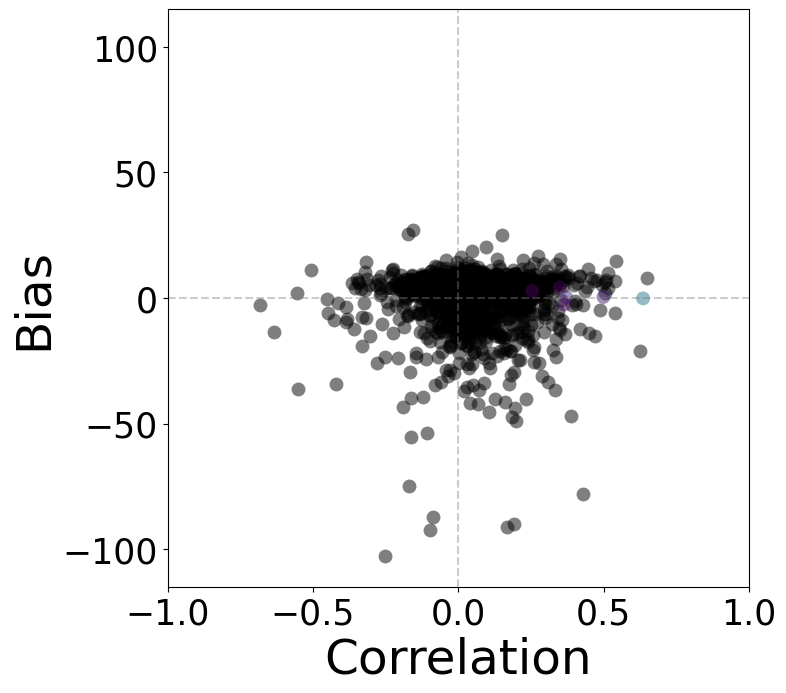}
    \caption{SO$_2$ Between Continent} 
  \end{subfigure}%
  \hspace*{\fill}   
  \raisebox{10mm}{
  \begin{subfigure}{0.08\textwidth}
    \includegraphics[width=\linewidth]{Figures/ModelPerformance_Bias_Correlation/correlation_bias_plot_legend.png}
  \end{subfigure}%
  }
  \hspace*{\fill}   
\caption{{\bfseries Bias vs Correlation for All Air Pollutants for Each Experiment.} Experiments across all air pollutants share the common trend of the correlation having a weaker relationship with a positive R$^2$ than the bias does. The situation of a high magnitude bias driving the model's poor performance can be seen across all air pollutants and experiments conducted. } \label{fig:allCorrelationBias}
\end{figure}

\begin{figure}
\hspace*{\fill}   
  \begin{subfigure}{0.49\textwidth}
    \includegraphics[width=\linewidth]{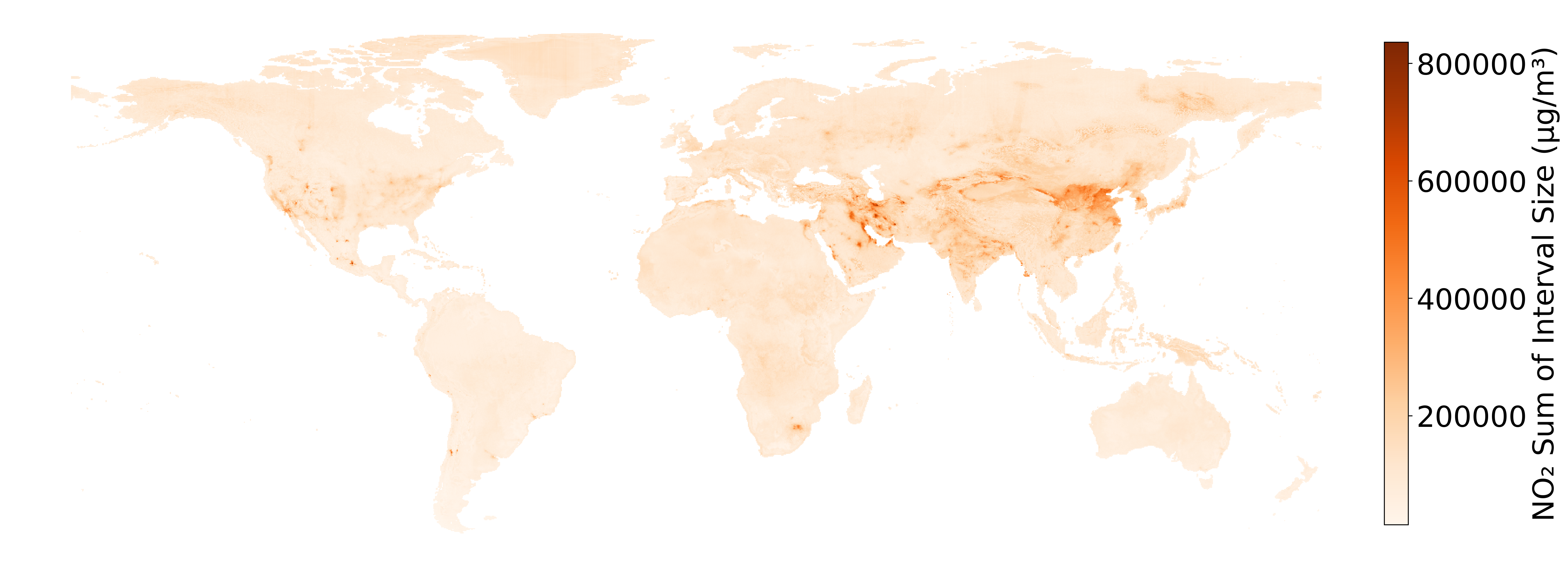}
    \caption{NO$_2$}
  \end{subfigure}%
 \hspace*{\fill}   
  \begin{subfigure}{0.49\textwidth}
    \includegraphics[width=\linewidth]{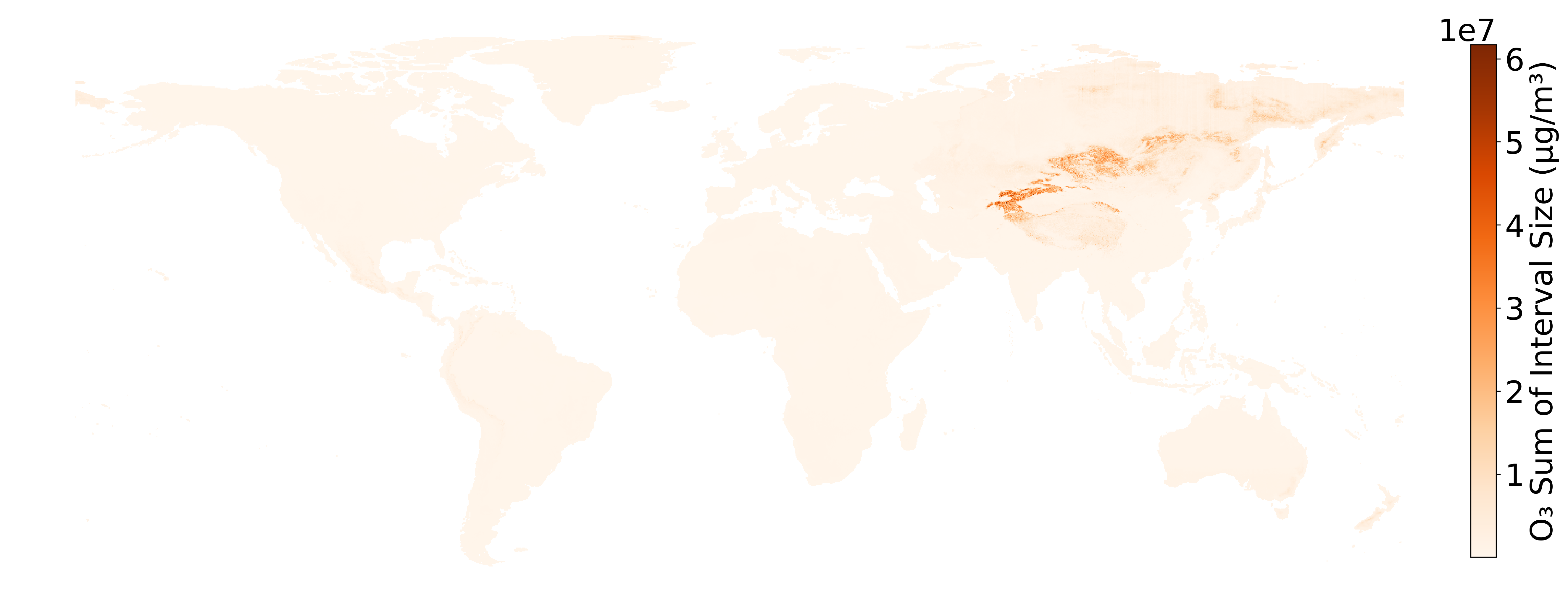}
    \caption{O$_3$}
  \end{subfigure}%
  \hspace*{\fill}   
  \\
  \hspace*{\fill}   
  \begin{subfigure}{0.49\textwidth}
    \includegraphics[width=\linewidth]{Figures/UncertainityPredictionInterval/uncertainity_interval_sum_pm10.png}
    \caption{PM$_{10}$}
  \end{subfigure}%
  \hspace*{\fill}   
  \begin{subfigure}{0.49\textwidth}
    \includegraphics[width=\linewidth]{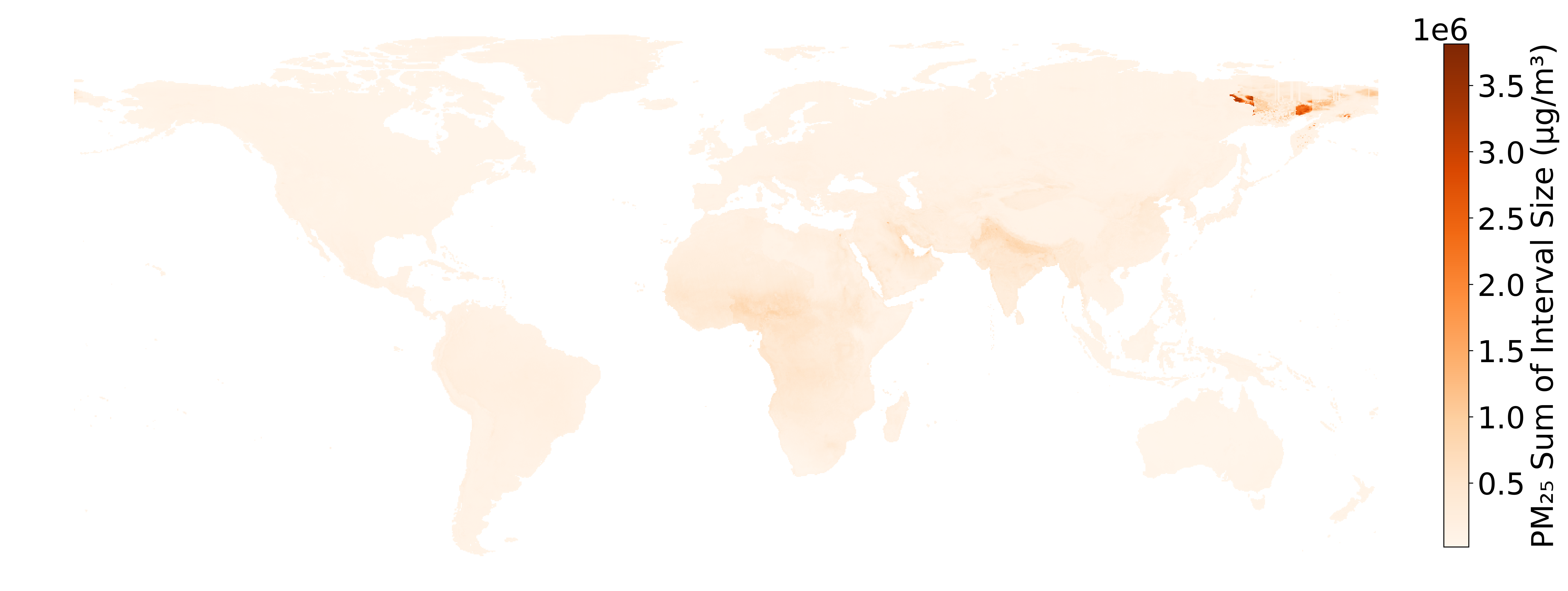}
    \caption{PM$_{2.5}$}
  \end{subfigure}%
  \hspace*{\fill}   
  \\
  \hspace*{\fill}   
  \begin{subfigure}{0.49\textwidth}
    \includegraphics[width=\linewidth]{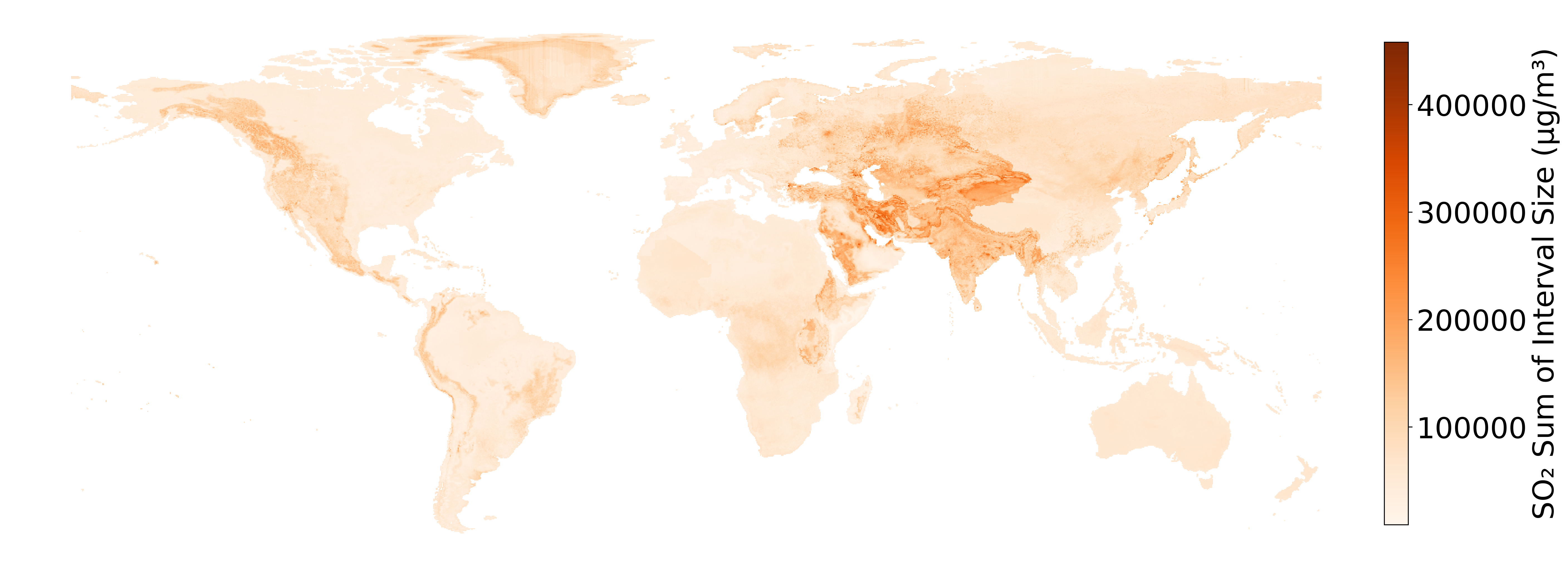}
    \caption{SO$_2$}
  \end{subfigure}%
  \hspace*{\fill}   
\caption{{\bfseries Prediction Interval Size Sum for All Air Pollutants for 2022.}} \label{fig:allAirPollutionUncertainityMaps}
\end{figure}